\documentclass{article}

\usepackage[preprint]{neurips_2023}
\usepackage{graphicx}
\usepackage{tabularx}
\usepackage{booktabs}
\usepackage{diagbox}
\usepackage{array}

\usepackage[utf8]{inputenc} % allow utf-8 input
\usepackage[T1]{fontenc}    % use 8-bit T1 fonts
\usepackage{hyperref}       % hyperlinks
\usepackage{url}            % simple URL typesetting
\usepackage{booktabs}       % professional-quality tables
\usepackage{amsfonts}       % blackboard math symbols
\usepackage{nicefrac}       % compact symbols for 1/2, etc.
\usepackage{microtype}      % microtypography
\usepackage{xcolor}         % colors

\title{Integrating Chemistry Knowledge in Large Language Models via Prompt Engineering}

\author{Hongxuan Liu\thanks{These authors contributed equally to this work.} \\
  Department of Chemical Engineering\\
  Tsinghua University\\
  Beijing, 100084 \\
  \texttt{liuhx21@mails.tsinghua.edu.cn} \\
  \And
  Haoyu Yin\footnotemark[1] \\
  Department of Chemical Engineering\\
  Tsinghua University\\
  Beijing, 100084 \\
  \texttt{yinhy23@mails.tsinghua.edu.cn}
  \And
  Zhiyao Luo \\
  Institute of Biomedical Engineering \\
  Department of Engineering Science \\
  University of Oxford \\
  Headington, Oxford OX3 7DQ United Kingdom \\
  \texttt{zhiyao.luo@eng.ox.ac.uk} \\
  \And
  Xiaonan Wang\thanks{Corresponding author.} \\
  Department of Chemical Engineering \\
  Tsinghua University \\
  Beijing, 100084 \\
  \texttt{wangxiaonan@tsinghua.edu.cn} \\
}

\begin{document}

\maketitle

\begin{abstract}
  This paper presents a study on the integration of domain-specific knowledge in prompt engineering to enhance the performance of large language models (LLMs) in scientific domains. A benchmark dataset is curated to encapsulate the intricate physical-chemical properties of small molecules, their drugability for pharmacology, alongside the functional attributes of enzymes and crystal materials, underscoring the relevance and applicability across biological and chemical domains.The proposed domain-knowledge embedded prompt engineering method outperforms traditional prompt engineering strategies on various metrics, including capability, accuracy, F1 score, and hallucination drop. The effectiveness of the method is demonstrated through case studies on complex materials including the MacMillan catalyst, paclitaxel, and lithium cobalt oxide. The results suggest that domain-knowledge prompts can guide LLMs to generate more accurate and relevant responses, highlighting the potential of LLMs as powerful tools for scientific discovery and innovation when equipped with domain-specific prompts. The study also discusses limitations and future directions for domain-specific prompt engineering development.
\end{abstract}

\section{Introduction}

The rapid advancement in artificial intelligence (AI) has significantly propelled its integration into natural science, specifically biology, chemistry and material science. Virtual screening contains thresholds determination and properties labeling, which can exhaust known design space [1] and guide experimental explorations [2]. Designing thresholds requires thorough domain insights but the rate-determining step in virtual screening is labeling the data. Early applications of AI in science were focused on properties predictions (e.g. formation energy [3], selectivity \& permeability of membranes [4-6], protein structures [7-8] and drug delivery [9-10]). As machine learning advances, more variants of artificial neural networks enabled AI to handle information in complex modal and solve more sophisticated problems in computational chemistry and bioinformatics. For example, MLP (multilayer perceptron) based machine learning potentials for molecular dynamics [11-12], GCN based DFT (density functional theory) for enzyme structural activity [13-14], CNN based molecular electron microscope images processing [15-16]. However, traditional high-throughput virtual screening is limited to known molecules or materials. The emergence of AI in inverse design emphasizes the need for innovative models that can assist experts in discovering new structures [17]. Models containing generating and predicting enable de novo design of molecules [18-19], drugs [20] and proteins [21-22].

A key challenge in applying AI to science is the lack of experimental data, which is often costly and time-consuming to gather. For instance, A significant drawback of using physics-based computational methods to gather data is the consumption of extensive computational resources and time. On the other hand, Directed modification and expression of enzyme genes also require several days to weeks of time. As a result, the lack of sufficient effective training data is always inadequate for training models effectively. [23]. Moreover, a trivial protein composed of 100 amino acid residues, with 20100 possible amino acid sequences and a corresponding vast space of potential configurations, makes searching, optimizing, and directed evolution of enzymes within the vast chemical space highly infeasible [24].

Overcoming the ‘small data’ challenge is basic but essential. Among tremendous approaches towards improving learning efficiency, large language models (LLMs) open a new channel for more efficient virtual screening apart from conventional methods, such as high-throughput computational methods [25-27], autonomous wet experiments [28-29], and data efficient algorithms (e.g. Bayesian optimization [30-31] and active learning [32]. LLMs are capable of processing and analyzing vast data amounts, which have notably advanced in addressing challenges like zero-shot reasoning, enabling them to handle tasks they haven’t been explicitly trained for. They also excel in incorporating domain knowledge across various fields and providing explanations in natural language, thereby enhancing their adaptability and accessibility. The LLM based AI agents [33] and pre-trained foundation models [34-35] are considered as the next generation of AI scientific assistants.

Prompt quality affects LLMs’ outputs significantly, many studies focus on well-defined prompts for general purposes (e.g., chain of thoughts reasoning [36], few shots learning [37]), known as prompt engineering [38-40]. Enhancing LLMs for specific fields typically involves fine-tuning, which can be complex and costly for those outside AI community [41]. Although there are already some domain-specific LLMs, they have not yet achieved the stability of general-purpose models like ChatGPT, leading many to focus on how to effectively utilize ChatGPT. Considering LLMs' remarkable learning abilities, strategic prompting or directing the LLM with specific instructions could be an effective alternative. However, current prompt engineering is mostly focused on general conditions such as academic writing [42] and science popularizing [43]. For experts in non-AI disciplines, the true value of these models lies in their domain-specific expertise, rather than their general capabilities. The absence of prompt engineering for specific areas makes LLMs user-unfriendly, especially for experimental chemists and material scientists.

Our paper studies the overlooked gap in AI for biology, chemistry and materials science, including small molecules, crystal materials and protein enzymes, highlight the importance of prompting to researchers off-the-shelve LLMs. Our investigation shows the critical need for solutions that combine AI's generative capabilities with detailed materials science insights, aiming to enhance model applicability and to address domain-specific challenges across various research areas.

In this article, we introduce “domain-knowledge embedded prompt engineering” as a novel approach to enhance LLM performance in specialized areas, as depicted in Fig. 1. First, we have created a set of domain-specific datasets for the first time, supplementing the existing public datasets. Second, we have developed and tested specific prompts for various tasks in three examples extracted from chemistry, materials science, and biology. Third, we combined the general methods of the computer science community for comparison, validating that the approach is correct. This approach aligns with desired outcomes and involves developing appropriate evaluation metrics. We also address the issue of LLMs generating inaccurate or 'hallucinated' responses and designed strategies to mitigate this. Last, through a case study, we demonstrate how our prompting strategies can address specific challenges in these fields. Overall, we showed that domain-knowledge embedded prompt engineering offers a cost-effective and efficient way to leverage the potential of LLMs.

\begin{figure}[ht]
  \centering
  \includegraphics[width=0.9\linewidth]{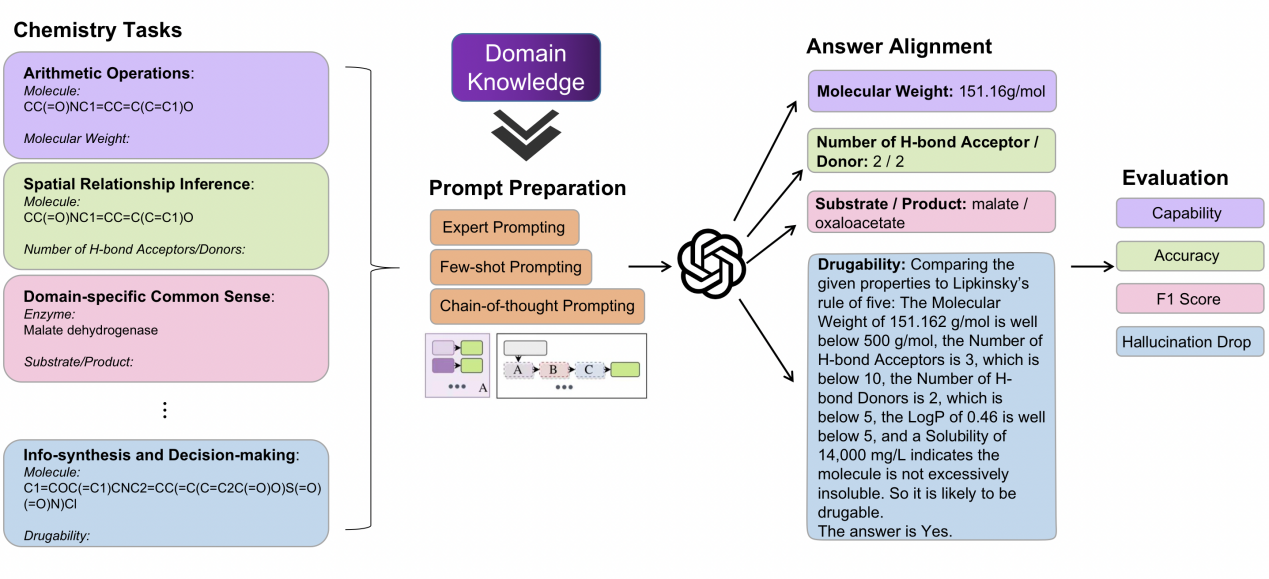}
  \caption{The Whole Process of Prompt Engineering Framework}
\end{figure}

\section{Methods}

In this chapter, we first introduce the construction of tasks from 3 domains: organic small molecules, enzymes and crystal materials, and the answer evaluation scheme for numerical and verbal tasks (See Section 2.1). We then formulate these tasks of domain question answering to an LLM question answering problem (See Section 2.2) and introduce various existing prompt engineering methods to address these tasks (See Section 2.3). Finally, we put forward our domain-knowledge embedded prompt engineering method (See Section 2.3).

\subsection{Dataset Construction and Answer Evaluation Scheme}

In task construction process, each of the 3 material categories (small molecule, enzyme and crystal material), holds significant relevance in academic research and practical applications. Organic small molecules are commonly utilized in pharmacy [44], while enzymes play a critical role in biocatalysis [45-46], and crystalline materials are essential in semiconductor technology and photovoltaic devices [47-48]. While mainstream benchmark datasets such as MMLU [49], Big-Bench [50] and GSM8k [51] have been widely applied to LLM performance evaluation, the composition of these datasets are usually generic math or reasoning questions, lacking a concentrated focus on some specific knowledge domains or subjects. Compared to these datasets, our datasets could provide a more comprehensive evaluation of LLM’s performance (using different prompt engineering methods) on specific chemistry domains.

We collect and curate a dataset of \textbf{1280} questions and corresponding solutions (See Table 1) for the evaluation of LLM’s capability, as described below:

\textbf{Organic Small Molecules}: 40 molecules proven to have significant drug properties or potentials are selected and curated from Pubchem [52], each containing 9 crucial structural and physical-chemical properties.\\

\textbf{Enzymes}: 40 enzymes involved in significant metabolic pathways in vivo are selected and curated from UniProt database [53], each with 7 crucial sequence and functional information.\\

\textbf{Crystal Materials}: 40 representative crystals derived from the Materials Project database [54] are selected, each with 16 crucial structural and energy properties.\\

A detailed enumeration and classification of all task types are contained in Supplementary S.2. Due to the limitations in API callings of proprietary LLMs, it is very hard to test molecules on a larger scale (like for thousands of molecules), but we believe that the selected molecules are already very representative to demonstrate LLM’s performance, and could pave the way for further applications in the future.

\begin{table}
\centering
\caption{Prompt Engineering Prediction Tasks}
\label{tab:question_classification}
\begin{tabular}{p{2.3cm} p{7cm} p{1.5cm} p{1.5cm}}
\toprule
Datasets & Tasks & Number of Molecules & Number of Tasks \\
\midrule
\textbf{Crystal Material} & Space Group Number, Lattice Angle ($\alpha$,$\beta$,$\gamma$), Lattice Vector (a,b,c), Density, Formation Energy, Energy Above Hull, Stability, Band Gap, Direct Gap, Metallic, Total Magnetization, Ordering & 40 & 640 \\
\textbf{Organic Small Molecule} & Molecular Formula, Melting Point, Density, Solubility, Molecular Weight, H-bond Acceptors, H-bond Donors, LogP, Drugability & 40 & 360 \\
\textbf{Enzyme} & Category, Substrate, Product, Active Site, Biological Process, Number of Amino Acids, Ligand & 40 & 280 \\
\textbf{Total} & 32 & 120 & 1280 \\
\bottomrule
\end{tabular}
\end{table}

In evaluating the performance of LLM prompt engineering methods on different tasks, four significant metrics are introduced:

\textbf{Capability}: To measure LLM’s capability to provide an answer for a certain task, regardless of its correctness. Its value takes 1 if the answer is effective otherwise 0.\\

\textbf{Accuracy}:To evaluate the extent to which LLM’s answer is identical or close to the ground truth. \\

\textbf{F1 Score}: to measure LLM’s predictive performance on multiple-choices questions, combining precision and recall. F1 Score offers a more comprehensive evaluation compared to Accuracy, especially in cases of imbalance where accuracy might be high but does not reflect the true performance of LLM.\\

\textbf{Hallucination Drop}: A metric to quantify the discrepancy between an LLM's ability to answer questions (Capability) and the accuracy of those answers (Accuracy). It takes 1 minus the ratio of Accuracy and Capability as the value. This metric helps identify when LLM is trapped in severe hallucinations on certain tasks.\\

Detailed implementations of these metrics are listed in Supplementary S.3.

In our approach, we utilize an LLM plugged-in automatic scheme to evaluate the metrics above. According to Table 3 in Supplementary, tasks can be divided into numerical and verbal ones, each of which takes a different manner to evaluate, respectively.

\textbf{Numerical Tasks}: All numerical tasks are transformed to the form of multiple-choices questions, as straightforward error estimation of the answer from ground truth can be strongly affected by unit and scale, and the form of multiple-choices makes it easier and more reasonable in evaluation across various tasks. Detailed implementation of tasks’ transformation into multiple-choices questions are described in Supplementary S.3. Metrics involved in numerical task evaluation are: \textbf{Capability}, \textbf{Accuracy}, \textbf{F1 Score} and \textbf{Hallucination Drop}.

While Capability, F1 Score and Hallucination Drop are evaluated in the normal form, the Accuracy of multiple choices questions is specifically defined. Full mark (1) is given if the option is exactly the ground truth. A partial score (0.4) is given if the value or range of the chosen option is adjacent to the ground truth. The complete scoring policy is listed in Supplementary S.3.\\

\textbf{Verbal Tasks}: For verbal answers, the LLM is guided by a series of grading examples coordinated to the specific question types and then required to give a grade to an answer. Detailed prompts for LLM’s grading tasks are listed in Supplementary S.4. Metrics involved in verbal task evaluation are: \textbf{Capability}, \textbf{Accuracy} and \textbf{Hallucination Drop}.

While Capability and Hallucination Drop are evaluated in the normal form, the Accuracy of verbal tasks take discrete values among \{0, 0.2, 0.4, 0.6, 0.8, 1\}. Score 0 means that the answer is completely irrelevant to the ground truth, while the scores \{0.2, 0.4, 0.6, 0.8\} imply part of the answer aligned with the ground truth, extent to which increases with the value. Score 1 corresponds to answers intrinsically the same as ground truth.\\

It is worth emphasizing that we believe the LLM plugged-in automatic scheme above for evaluation could bear skepticism on fairness and effectiveness, as LLM’s evaluation process is independent from LLM’s predictive task performing in the last step, implying the LLM would not take past memories of task performing or “know” the answers were generated by itself, and thus is unlikely to “cheat” on the grading process.

Fig. 2 shows the flow chart of question construction and answer evaluation process. Data from 3 material categories are extracted and combined to form proper questions (some are in the form of multiple-choices questions). When the raw answers are acquired, they need to be checked for validity, and then aligned to proper answer forms. Ultimately the answers are automatically graded.

\begin{figure}[ht]
  \centering
  \includegraphics[width=0.8\linewidth]{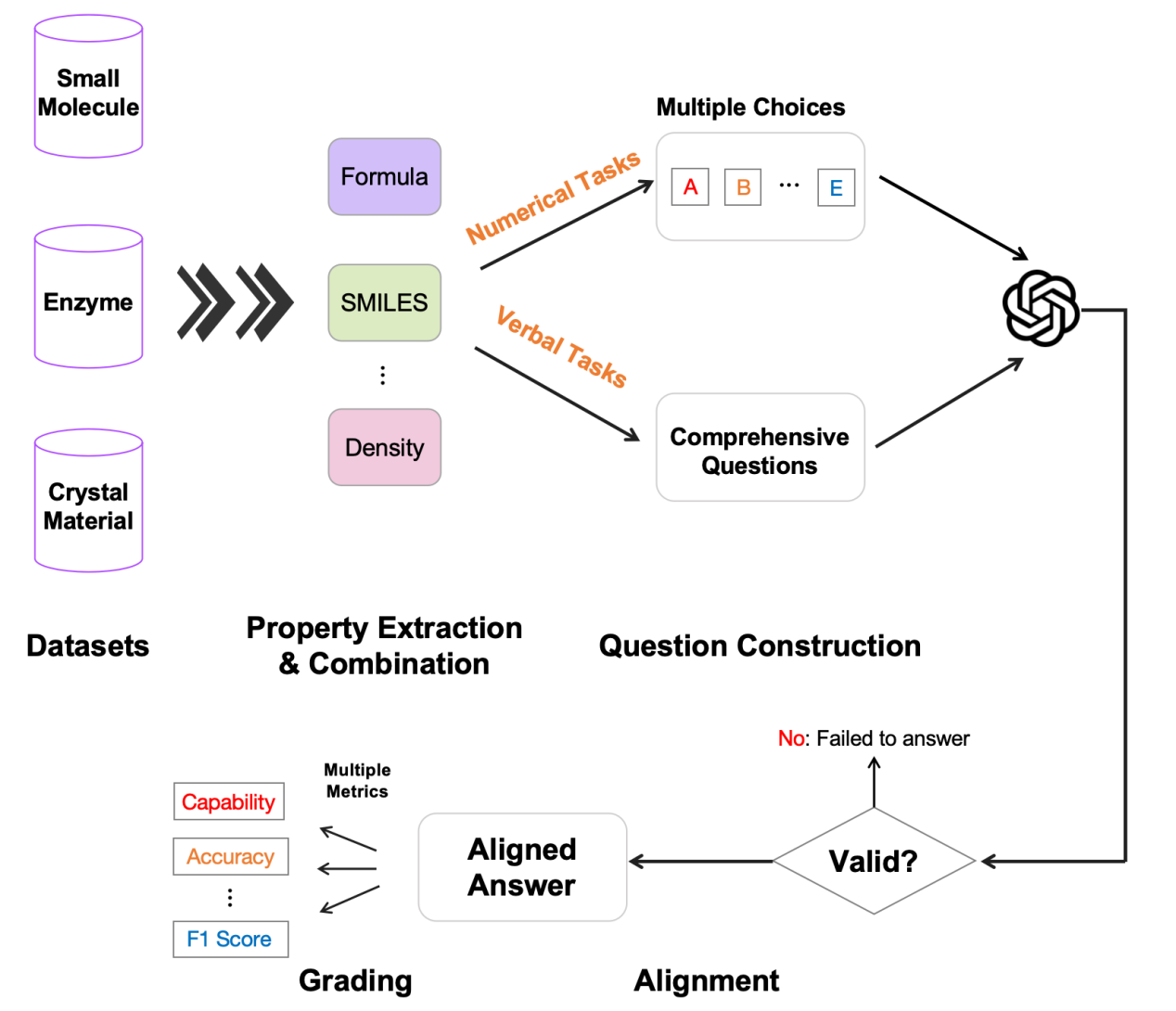}
  \caption{Question Construction, Answer Alignment and Grading Process}
\end{figure}

\subsection{Scientific Prediction as a LLM Question Answering Problem}

In the LLM era, scientific prediction can be considered as a question answering task leveraging the zero-shot/few-shot reasoning power of LLM. It is demonstrated that by providing in-context hints to language model with size large enough for emergence to happen, the model can excavate knowledge learned from pre-trained data and well-perform the question answering task [55]. As an approach to enhance LLM’s capability on specific domains or tasks, prompt engineering significantly reduces the need for extensive task-specific datasets as required in LLM fine-tuning paradigm, making it an effective in-context learning method for LLM enhancement.

The process of prompt engineering could be mathematically formalized [56]. Let $Q$ be the question, $P$ be the prompt, $A$ be the answer by LLM, prompt engineering process is to determine the context of prompt words $P$ such that the answer $A$ could be given effectively by LLM:

\begin{equation}
A=f(P,Q)
\end{equation}

where $f$ is the LLM.

A prompt optimization objective is to find:

\begin{equation}
\mathrm{arg\,max}_P g(f(P,Q),S)
\end{equation}

where $S$ is the ground truth solution, and $g$ is a evaluation function which measures how much the LLM answer $A$ is in accordance with the ground truth solution $S$.

For our dataset $D=\{Q_i,S_i\}_i^n$, the general prompt optimization objective is to find the $P$ that maximizes the expectation over the dataset:

\begin{equation}
\mathrm{arg\,max}_P \mathbb{E}_{Q,S\in D}g(f(P,Q),S)
\end{equation}

\subsection{Common Prompt Engineering Techniques and Domain-Knowledge Embedded Prompt Engineering}

\begin{figure}[ht]
  \centering
  \includegraphics[width=0.9\linewidth]{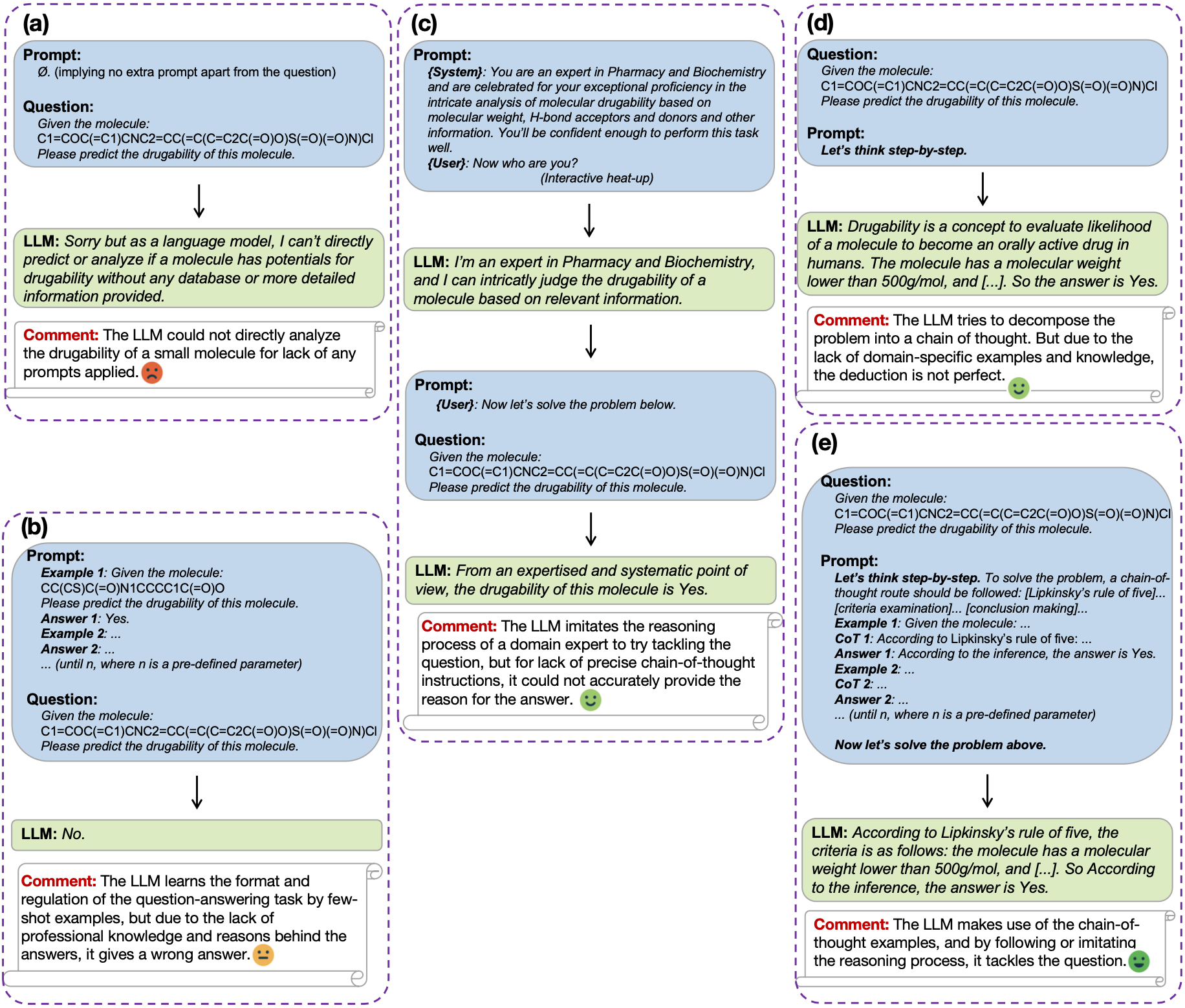}
  \caption{Illustration of the Mainstream Prompt Engineering Methods}
\end{figure}

The essence of prompt engineering is to harness the full potential of LLMs in diverse applications by ensuring they respond in a manner that is most aligned with the user's intent and the task at hand. We give a brief introduction to several mainstream prompt engineering methods:

\textbf{Zero-shot Prompting}: Zero-shot Prompting requires LLM to answer the given question directly without providing any data or example questions in the context. (See Fig. 3 (a))\\

\textbf{Few-shot Prompting}: In Few-shot Prompting, an LLM is presented with several \textit{demonstrations}, (i.e. question-answer pairs within the prompt context), better equipping the LLM to understand and replicate the response format and content. The demonstrations in prompt can be formalized to: $P=\{(Q_1,A_1),...,(Q_m,A_m)\}$, where $m$ is the number of examples [55]. (See Fig. 3 (b))\\

\textbf{Expert Prompting}: Role-play instructions have demonstrated their effectiveness in harnessing the potential of LLMs [56]. By guiding LLMs step-by-step into assuming the role of domain experts, they can generate responses akin to those written by experts (See Fig. 3 (c)). \\

\textbf{Zero-shot CoT (Chain-of-Thought) Prompting}: By eliciting a sequential, step-by-step reasoning process to effectively address complex tasks, CoT enables the model to break down a task into its constituent parts, offering a clear and logical pathway to the solution [36]. In particular, Zero-shot CoT prompting involves adding \textit{Let’s think step by step} to the prompt as a trigger-sentence. (See Fig. 3 (d))\\

\textbf{Few-shot CoT Prompting}: In addition to adding \textit{Let’s think step by step} to the prompt like Zero-shot CoT, Few-shot CoT provides several examples of Thought-Chain in solving similar problems to assist LLM perform the current task in a similar manner [57]. The demonstrations in prompt can be formalized to: $P=\{(Q_1,C_1,A_1),...,(Q_m,C_m,A_m)\}$. (See Fig. 3 (e))\\

A significant limitation of these prompt engineering methods is that they do not incorporate domain expertise as guidance for problem-solving, which considerably restricts the capabilities of LLMs in numerous domain-specific tasks. Moreover, since addressing many domain-specific challenges involves intricate cognitive processes, it is imperative to strategically combine various prompt engineering techniques at different stages to achieve optimality.

Here we propose a \textbf{domain-knowledge embedded prompt engineering strategy} that integrates chemistry knowledge into language model. The prompting scheme takes the form of multi-expert mixture. Each expert takes part in role playing and are given a few shots of CoT demonstrations integrated with expertise domain knowledge or instructions.

Here, incorporating \textit{domain knowledge} essentially involves integrating the thought processes of chemistry/biology experts. This contrasts with the conventional zero-shot CoT approach, which merely prompts LLMs to engage in a chain of thought. By doing so, it offers more precise background knowledge and exemplifies more accurate human reasoning. The full documentation of all domain-knowledge prompts are listed in Supplementary S.4. Then the experts' answers would be assembled through the principle of \textit{minority submission to the majority}. The detailed scheme of these strategies are delineated in Fig. 3 and Fig. 4.

\begin{figure}[ht]
  \centering
  \includegraphics[width=0.9\linewidth]{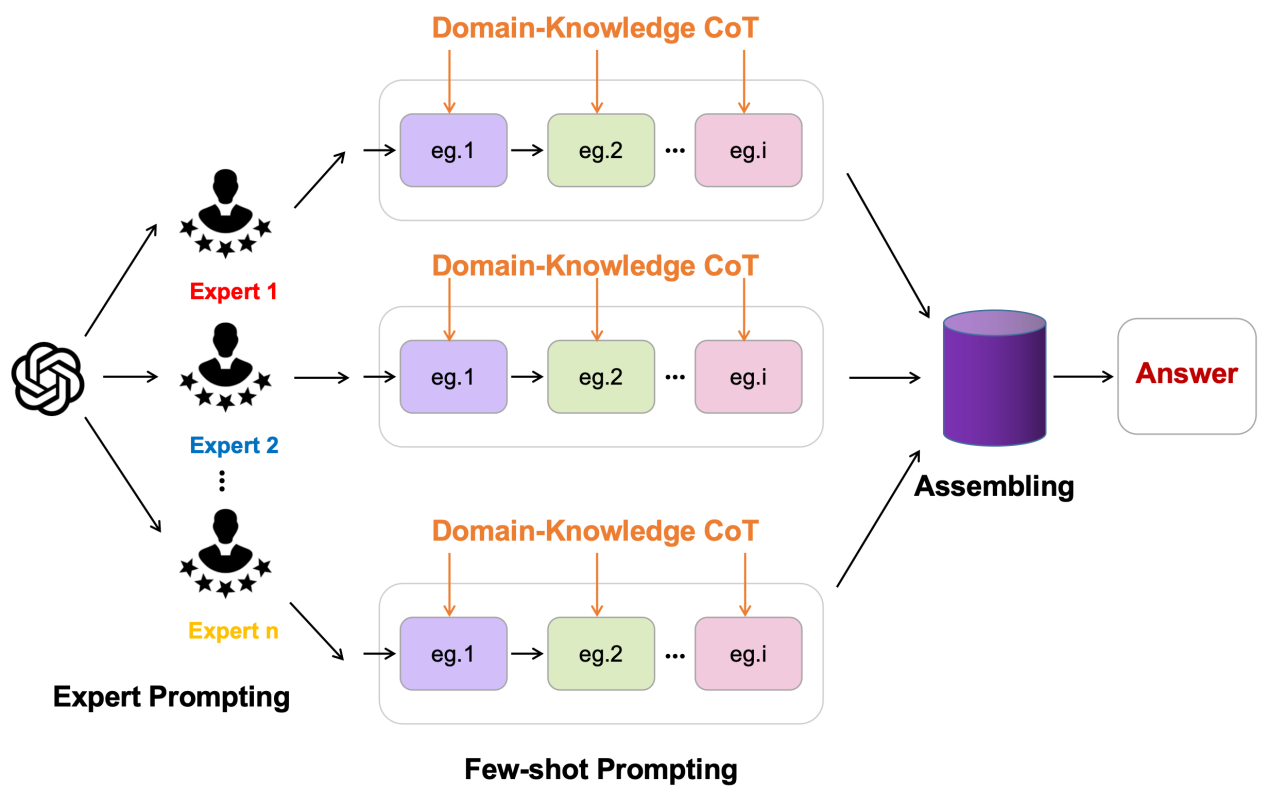}
  \caption{The Whole Process of Domain-Knowledge Prompt Engineering Method}
\end{figure}

In the following chapters, we compare this prompt engineering method proposed above to other generic prompt engineering methods including zero-shot prompting, few-shot prompting, expert prompting, and CoT prompting.

\section{Results}

In this chapter, we first present the overall benchmarks of prompt engineering methods over all tasks. After that, we make detailed comparisons over different task types, CoT complexities and material types. In the last section, three case studies on representative molecules are conducted using our tailored domain-knowledge embedded prompt engineering method to illustrate the effectiveness of prompt engineering in assisting crucial scientific research topics.

\subsection{Summery of Overall Performance}

In our study, we evaluated five different prompt engineering strategies across three datasets (small molecule, enzyme, and crystal material), each yielding three sets of answers for robustness. The LLM model being evaluated is ‘gpt-3.5-turbo-1106’ [58] through official API calling. The prompt engineering strategies included zero-shot, few-shot, expert, and zero-shot CoT, along with domain-knowledge method (ours). The overall evaluation results on three datasets are shown in Fig. 5 and Fig. 6:

\begin{figure}[ht]
  \centering
  \includegraphics[width=0.9\linewidth]{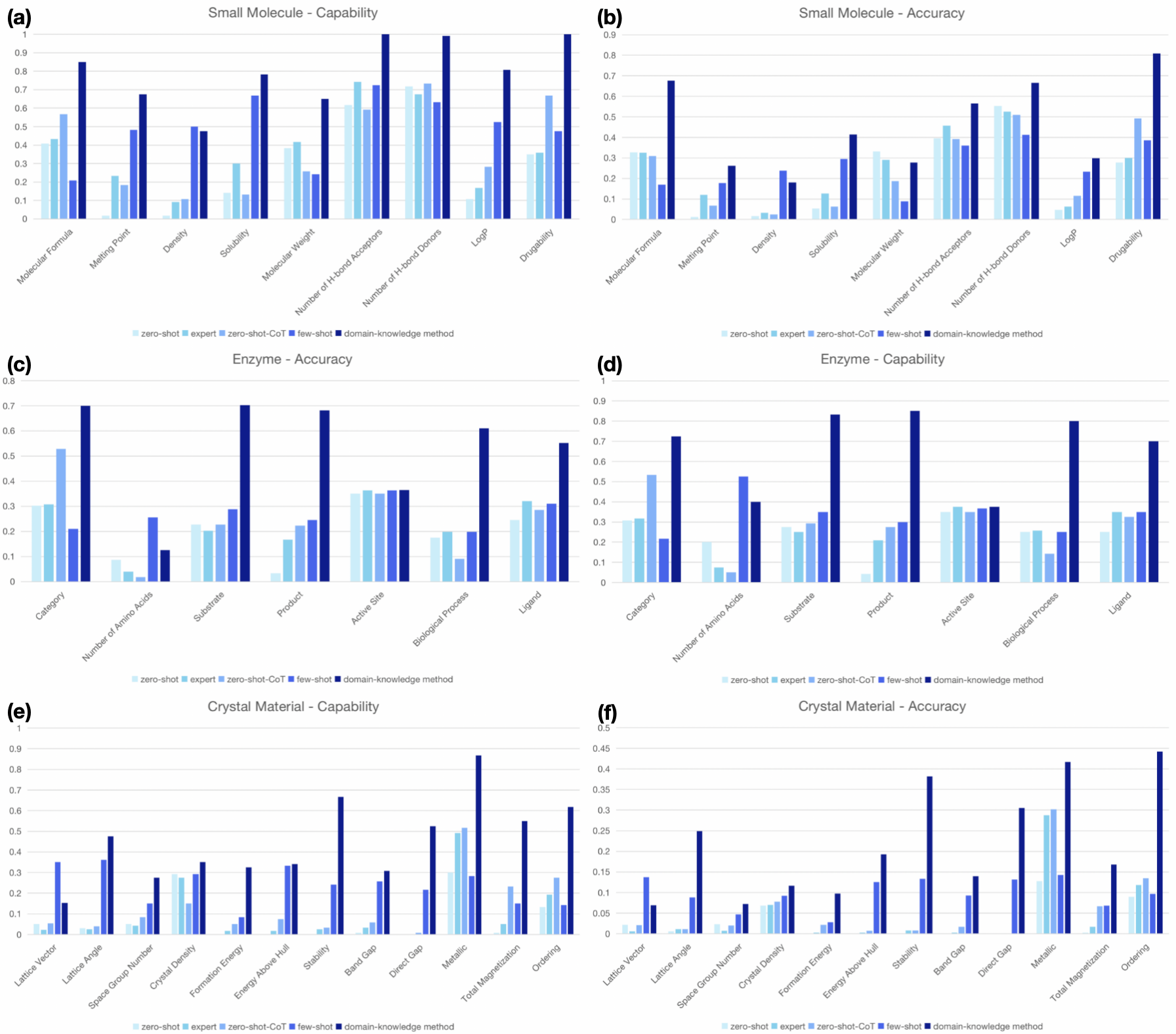}
  \caption{Capability and Accuracy for All Tasks}
\end{figure}

\begin{figure}[ht]
  \centering
  \includegraphics[width=0.9\linewidth]{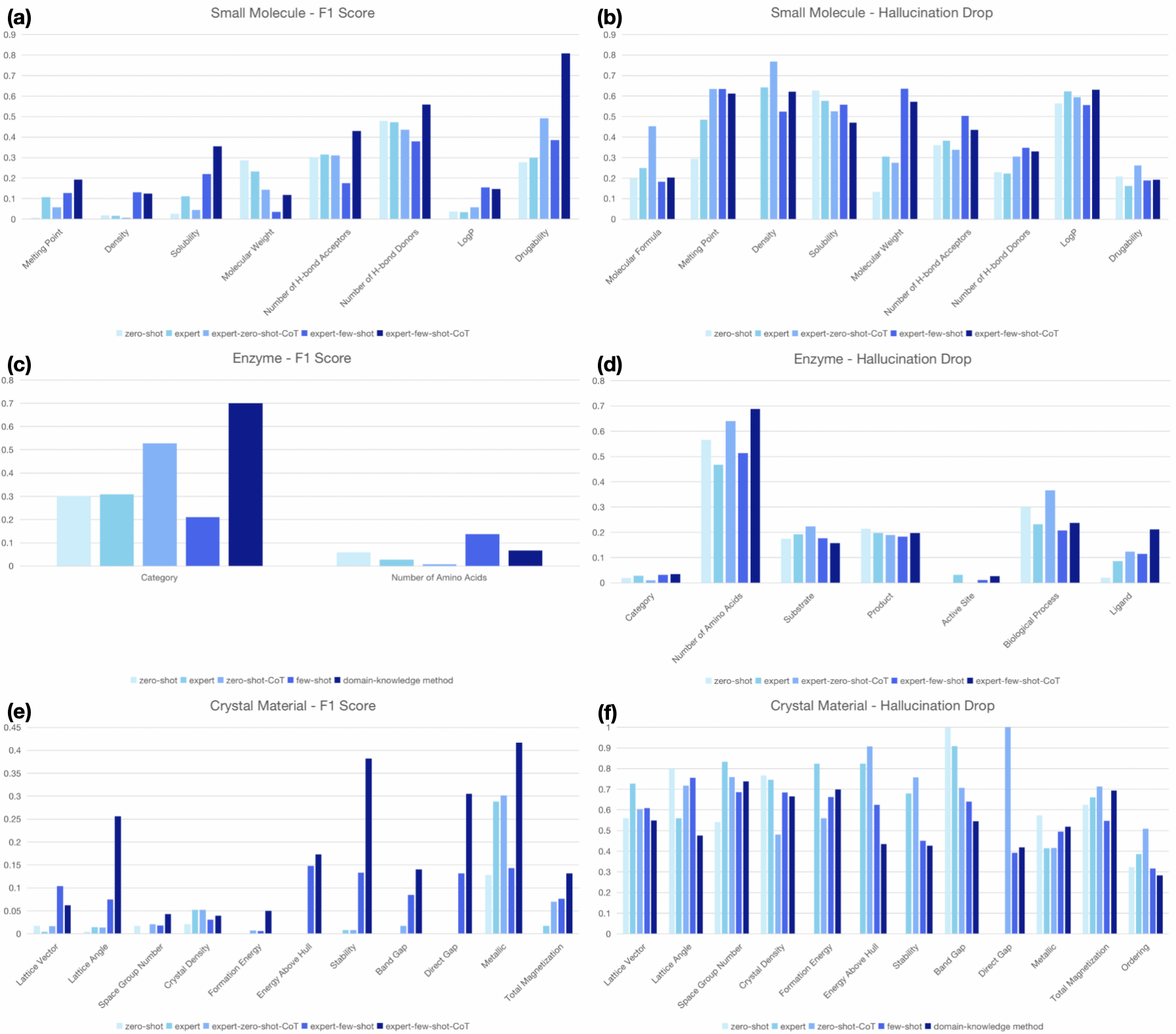}
  \caption{F1 Score and Hallucination Drop for All Tasks}
\end{figure}

Our domain-knowledge embedded prompt engineering method outperforms other conventional prompt engineering techniques on most tasks and metrics. In nearly all tasks on small molecules and crystal materials, and more than 50\% of the tasks on enzymes, our method’s performance is very significantly higher than other methods, while on tasks: Molecular Density, Molecular Weight, Number of Amino Acids and Active Sites, our method does not demonstrate obvious advantages.

In the following sections, we make more detailed comparisons for different tasks and molecules. Due to space limitation, we only present the key findings in the following sections. In Section 3.2, we compare these method’s performance on different task types, while in Section 3.3, we delve into the correlation between prompt engineering method’ performance and CoT complexity. Finally we compare prompt engineering methods’ effectiveness on different types of materials.

\subsection{Comparison by Task Types}

In this section, we compare various prompt engineering methods performance on different types of tasks, classification of which can be referred to in Table 3 and Table 4. Each task type presents unique challenges and necessitates different inference abilities from the LLM. After aggregation, the performances of 5 prompt engineering methods on different question groups are shown in Fig. 7:

\begin{figure}[ht]
  \centering
  \includegraphics[width=0.9\linewidth]{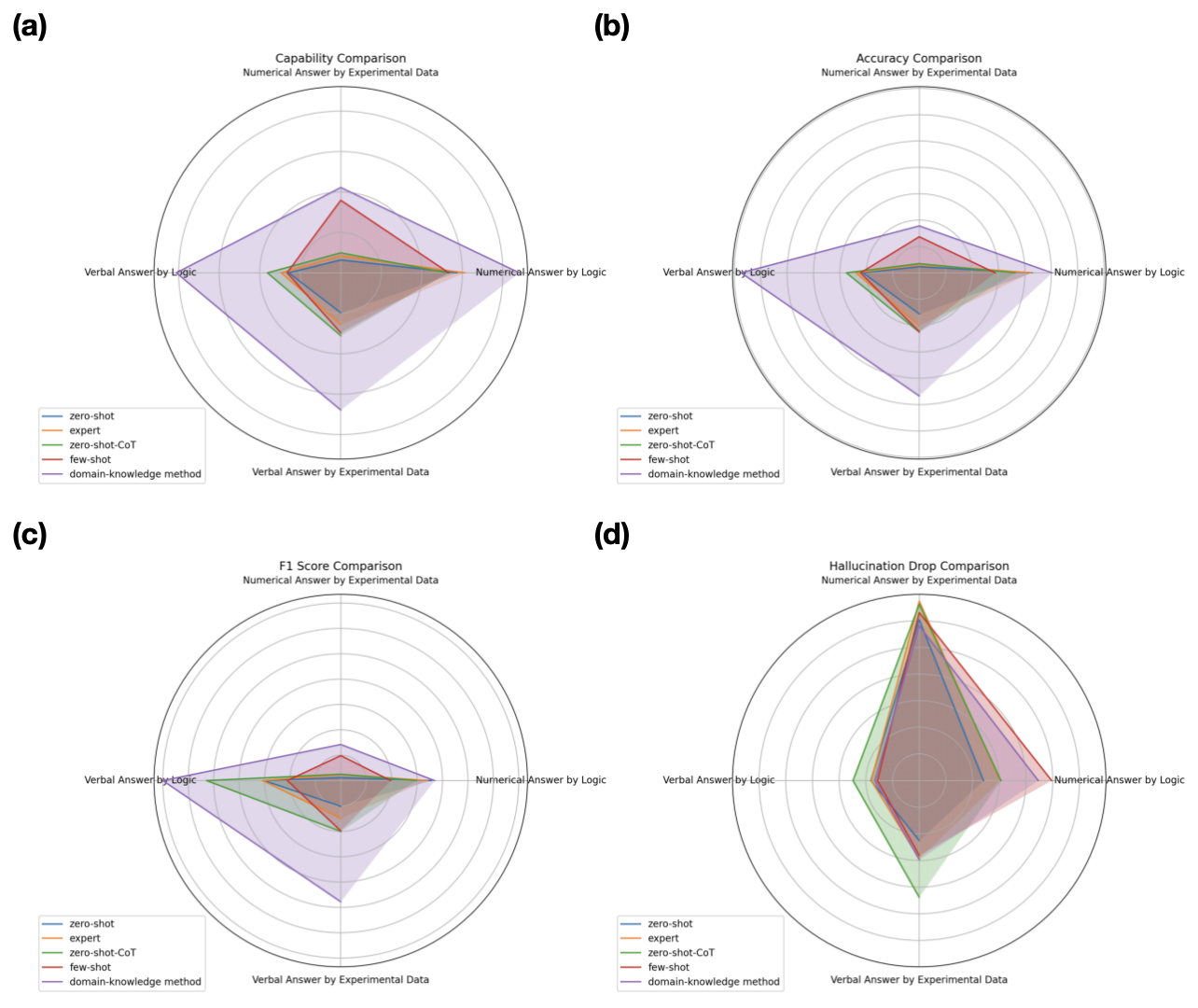}
  \caption{Prompt Engineering Performances by Output Type}
\end{figure}

\textbf{(1) Domain-knowledge embedded prompt engineering method outperforms traditional prompt engineering methods on all question types.} Through a comprehensive evaluation across various groups of prediction tasks, focusing on four crucial indices - "Capability", "Accuracy", "F1 Score" and "Hallucination Drop", our domain-knowledge embedded prompt engineering method consistently outperforms traditional prompt engineering strategies. This superiority is evident in the substantial enhancement of both capability and accuracy metrics, with the most notable improvements exceeding a 100\% boost. Such findings unequivocally demonstrate that integrating domain-specific knowledge into prompt engineering substantially elevates the effectiveness of generic prompt engineering techniques. \\

\textbf{(2) LLM performs better for answers derived from logical reasoning than answers based on experimental data.} This tendency is further amplified in our domain-specific prompt engineering method, where a more tailored prompt engineering strategy is applied. As shown in Fig. 7 (a), (b) and (c), it consistently leads to more significant improvements in tasks involving logical deduction compared to other prompt engineering methods. This disparity in performance can be attributed to the fact that LLMs, with refined prompt engineering, can engage in a sophisticated Chain-of-Thought process, enabling LLMs to excel in tasks that demand intricate reasoning and problem-solving skills. However, despite being trained on various scientific databases, LLMs do not excel in precisely replicating exact data values. This brings about their ability to process and reason through information well rather than serve as direct conduits for data retrieval.\\

\textbf{(3) LLM performs better on verbal tasks compared to numerical tasks.} When faced with tasks that require a numerical response, (actually in formats involving multiple choices), LLMs tend to exhibit weaker performance. This is evident in both capability and accuracy metrics across various prompt engineering methods, with numerical answers derived from experimental data showing the least favorable results (Fig. 7 (a), (b), (c)). When LLMs engage in numerical reasoning, their capability scores are notably higher (Fig. 7 (a)), but this advantage is tempered by significant issues with hallucinations, which adversely affect the accuracy of these responses. In fact, even when the most advanced prompt engineering methods are applied, the accuracy of logical numerical answers is surpassed by that of logical answers. This trend underscores a recognized weakness of LLMs in number-related tasks, as evidenced by several research studies.\\

\textbf{(4) Domain-knowledge embedded prompt engineering method effectively reduces hallucination.} The metric of hallucination drop serves as a barometer for the average quality of answers produced by LLMs under different prompt engineering strategies. As shown in Fig. 7 (d), the question type of numerical answer by logic is the only category where an increase in hallucination is observed as the domain-knowledge embedded prompt engineering is applied. In the other three question types, the incorporation of domain-specific knowledge into the prompt engineering process effectively curtails the occurrence of hallucinations. Notably, the question types 'numerical answer by experimental data' and 'verbal answer by logic' emerge as front-runners, registering the top two lowest scores in hallucination drop. This outcome underscores the precision and effectiveness of domain-knowledge embedded prompt engineering methods in enhancing the reliability and accuracy of LLM responses.\\

These results from a more detailed classification based on reasoning paradigm also draw some intriguing conclusions below, showing the distinctive strengths and drawbacks in LLM reasoning:

\begin{figure}[ht]
  \centering
  \includegraphics[width=0.9\linewidth]{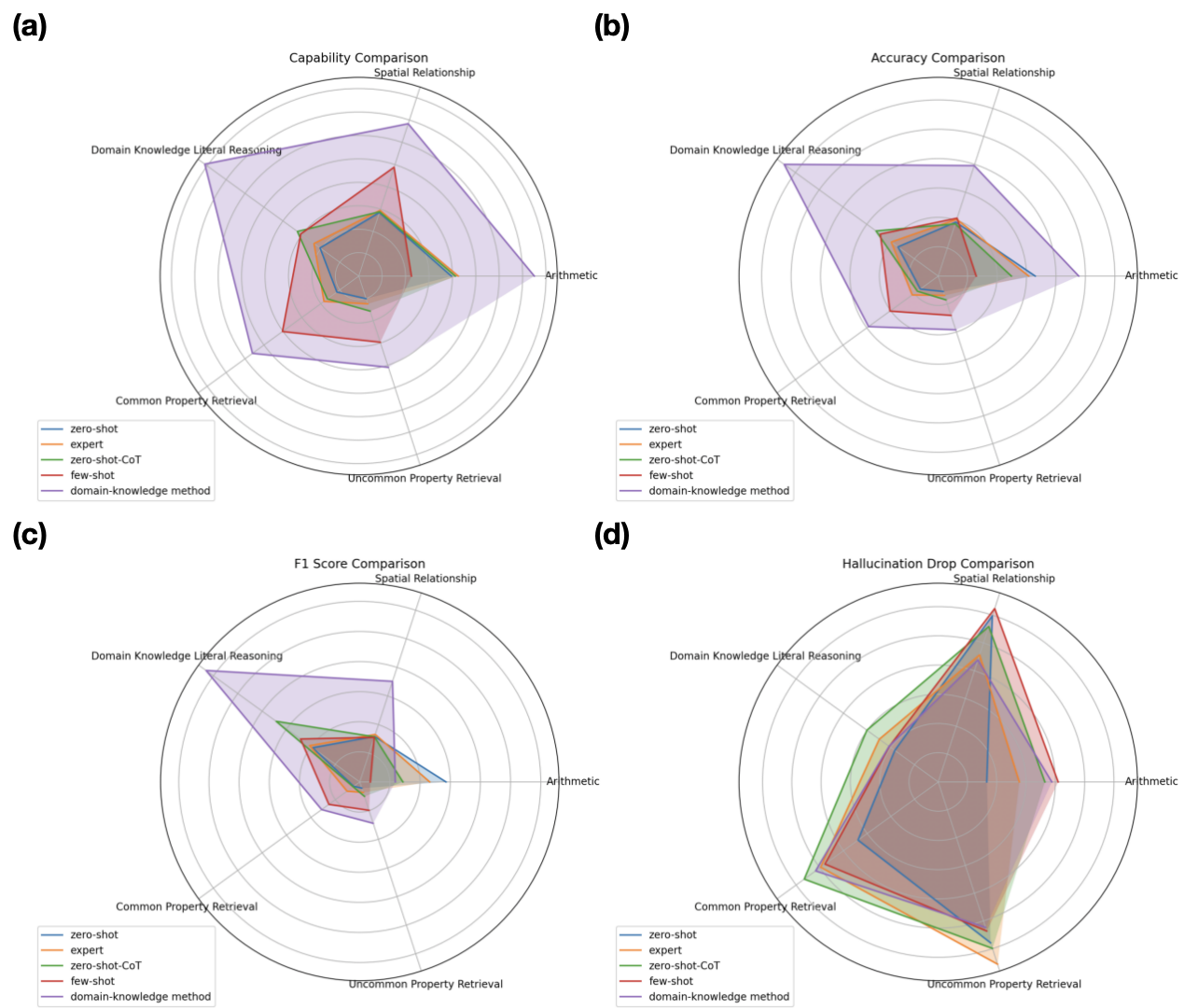}
  \caption{Prompt Engineering Performances by Reasoning Paradigm}
\end{figure}

\textbf{(5) LLM performs poorly on arithmetic tasks.} These tasks revolving around basic counting, adding, and multiplying abilities, ostensibly require less sophisticated cognitive skills compared to tasks that necessitate spatial imagination or intense domain-knowledge based reasoning, but the performance of LLMs in these arithmetic tasks is unexpectedly subpar. Despite scoring high in capability, LLMs do not exhibit a corresponding lead in accuracy, showing higher occurrence of hallucinations in these tasks. In fact, the accuracy of LLMs in arithmetic tasks is not only significantly outpaced by domain knowledge literal reasoning tasks but also closely rivalled by spatial relationship tasks (Fig. 8 (b)). Notably, even the application of CoT heuristics in the reasoning process does not substantially mitigate this issue. This is evident in the Hallucination Drop metric, where both zero-shot-CoT and domain-knowledge embedded method exhibit a higher incidence of hallucination phenomena in arithmetic tasks compared to others (Fig. 8 (d)).\\

\textbf{(6) LLM is incapable on many information retrieval tasks.} These tasks, which cannot be effectively addressed through reasoning alone, generally exhibit poorer performance compared to those based purely on reasoning. As depicted in Fig. 8 (a), (b) and (c), tasks involving the retrieval of both common and uncommon properties record the lowest capability and accuracy scores, with tasks involving uncommon information faring slightly worse. Prompt engineering falls short in information retrieval, primarily due to its inability to provide direct access to external databases, but a well-crafted, domain-specific prompt can still marginally improve LLM performance by encouraging a more detailed response, as indicated by the higher capability scores for domain-knowledge prompt engineering. Despite this, the challenge of mitigating hallucinations remains formidable, with the highest incidence of hallucination observed in these types of tasks when using domain-knowledge prompts.\\

\textbf{(7) Verbal reasoning tasks get largest boosting with domain-knowledge embedded prompt engineering method.} In 5 question types classified by reasoning paradigms, "Domain Knowledge Literal Reasoning Tasks" distinctly stand out, especially when enhanced by domain-knowledge embedded prompt engineering methods. This category of tasks not only achieves the highest capability and accuracy scores overall but also maintains a relatively low level of hallucinations.This demonstrates well-crafted prompts can, in a remarkably efficient manner, stimulate the latent capabilities of LLMs, enabling them to generate answers with heightened confidence and precision.\\

\subsection{Comparison by CoT Complexity}

In this section, we compare different prompt engineering methods’ distinction under a variety of CoT complexities, in order to depict our tailored prompt engineering method’s superiority under different CoT complexities. We propose that the quantity of additional properties added in CoT prompts serves as a viable metric for gauging the complexity of the CoT process. This metric reflects the extent of extra information that is integrated into the CoT reasoning, which in turn influences the complexity and depth of the reasoning required. To operationalize this, we have categorized tasks based on the number of additional properties provided in each question, as shown on Table 5 in Supplementary.

It is, however, worth noticing that the number of additional properties provided (namely, the complexity of CoT) does not necessarily correlates to the difficulty of questions. The aggregated results are shown in Fig. 9:

\begin{figure}[ht]
  \centering
  \includegraphics[width=0.9\linewidth]{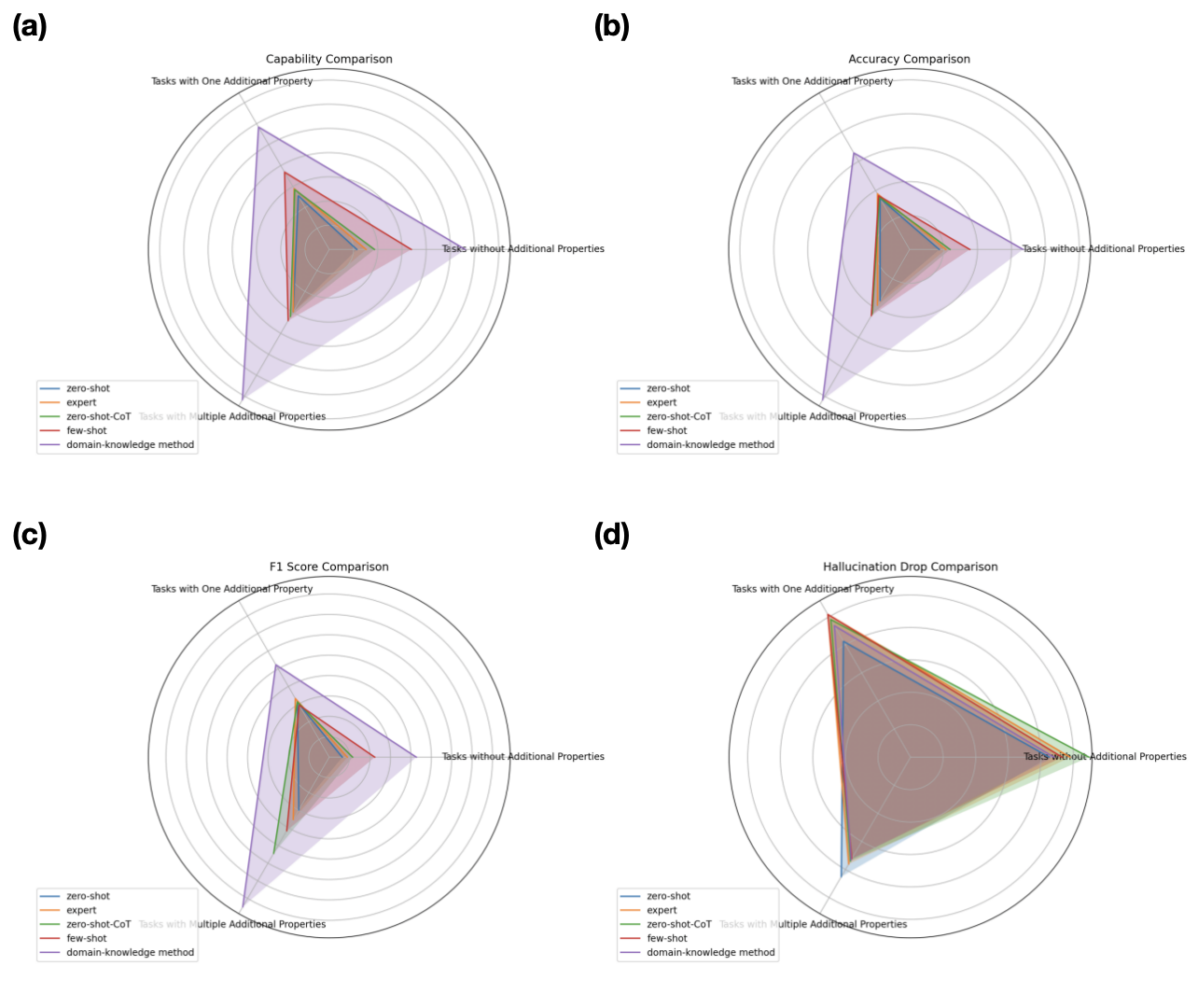}
  \caption{Prompt Engineering Performances by CoT Complexity}
\end{figure}

\textbf{(1) Domain-Knowledge Embedded Prompt Engineering produces greatest performance lift in tasks with most complicated CoT formulation.} In scenarios where LLMs are presented with different amounts of additional information for task execution, the domain-knowledge embedded prompt engineering method emerges as the most effective, outshining others in three key performance metrics: "Capability", "Accuracy", and "F1 Score". Specifically, it excels remarkably in "Tasks with Multiple Additional Properties" (Fig. 9 (a), (b), (c)). This highlights the advantage of domain-knowledge prompts in enhancing LLM performance in tasks that demand a complex CoT formulation. Furthermore, even for simpler zero-shot CoT method, this benefit makes it reverse the lead of few-shot method in "Tasks with Multiple Additional Properties", especially on F1 Score (Fig. 9 (c)). This aligns well with the intuitive understanding of CoT in enhancing inference-related capabilities.\\

\textbf{(2) In-Context Information Could Effectively Reduce Hallucination Level.} Tasks supplied with the most in-context extra information consistently exhibit the lowest levels of hallucination across all prompt engineering methods, as shown in Fig. 9 (d). This trend holds true regardless of whether the prompt engineering method incorporates domain-knowledge features. A notable observation is that many tasks in the "Tasks with Multiple Additional Properties" category are inherently complex and challenging. For example, predicting the drugability of a small molecule often necessitates a thorough and intricate examination under Lipinski's Rule of Five. Similarly, calculating the crystal density of a substance involves complex computations, including the determination of relative molecular mass of a unit cell, the measurement of unit cell volume, and intricate unit transformations. The surprisingly low hallucination levels is indicative of the effectiveness of providing additional in-context information,suggests that enriching LLM prompts with more contextual information and factual details may substantially enhance the robustness and reliability of the generated content.\\

\subsection{Comparison by Material Differences}

In this section, a detailed comparison of prompt engineering accuracy on three types of materials will be portrayed. For clarity, we only focus on our tailored prompting method (namely the domain-knowledge embedded prompting)'s performance on small molecules, enzymes and crystal materials with divergent material traits. The methodology employed to quantify the differences among these materials will be elaborated upon in the following paragraphs.

For small molecules, we propose two indicators—\textbf{molecular weight} and \textbf{elemental composition}—to differentiate the complexity of various molecules. This is predicated on the rationale that more complex molecules typically necessitate a higher level of analytical effort, which could inversely affect accuracy. We aim to investigate whether this assumption aligns with the empirical results obtained from our study.

For enzymes, we also employ two indicators to discern the difficulty in predicting properties among different enzymes: \textbf{enzyme scale}, quantified by the number of amino acids, and the \textbf{current depth of research}, denoted as the number of reviewed publications recorded on Uniprot [53]. It is hypothesized that more complex enzymes, characterized by longer peptide chains and a lesser degree of comprehensive research, necessitate a higher analytical effort, potentially reducing accuracy. We intend to explore whether this hypothesis is consistent with the overall findings of our analysis.

For crystalline materials, we utilize two indicators to gauge complexity: \textbf{formula complexity}, which reflects the complexity of a single unit cell, and \textbf{unit cell symmetry}, denoted by the crystal system to which it belongs. The underlying premise is that more complex crystalline materials demand a more substantial analytical effort, which could, in turn, diminish accuracy. We will investigate whether this premise aligns with the collective results of our study.

\begin{figure}[ht]
  \centering
  \includegraphics[width=0.9\linewidth]{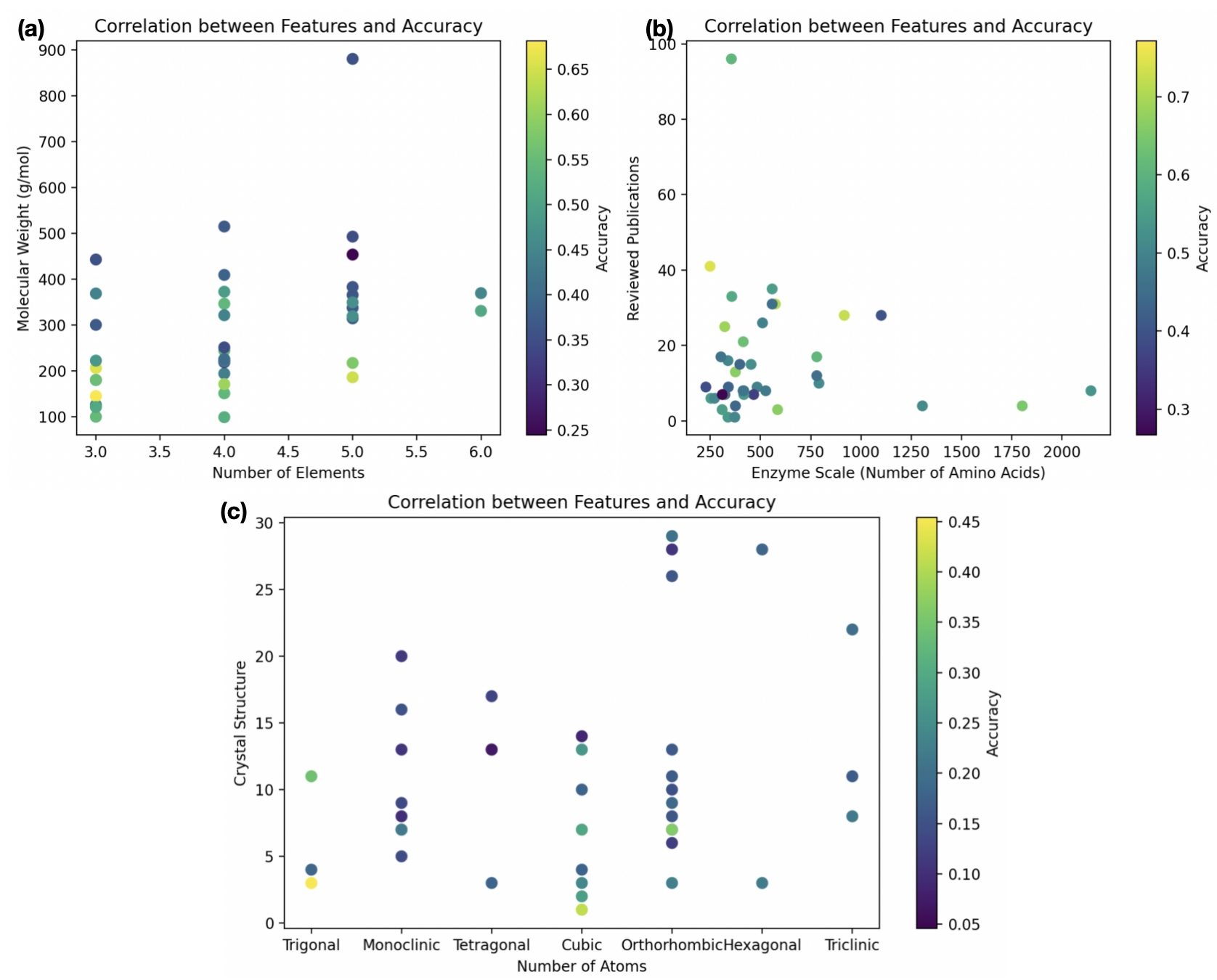}
  \caption{Prompt Engineering Performances on Different Materials}
\end{figure}

\textbf{(1) The prediction accuracy of LLMs deteriorates for larger and more complex organic molecules.} As the molecular weight increases and the elemental composition becomes more diverse, we observe a gradual decline in the LLM's prediction accuracy. Specifically, molecules comprising more than five distinct elements exhibit significantly poorer performance compared to those with fewer components. Moreover, when the molecular weight exceeds 300 g/mol, the overall accuracy for single molecule predictions generally falls below 30\%, as shown in Fig. 10 (a). Furthermore, large organic molecules are less commonly found in literature compared to smaller molecules, exacerbating the difficulty of LLM's information retrieval. \\

\textbf{(2) The accuracy of LLMs in predicting properties of specific enzymes aligns closely with the depth of current research on these enzymes but shows a weak correlation with the enzymes' size.} The number of reviewed publications recorded on Uniprot, which signifies the academic community's past research focus on an enzyme, demonstrates a strong correlation with the LLM's prediction performance. The more thoroughly an enzyme is researched and understood, the higher the accuracy of LLM predictions. Most enzymes with low prediction accuracy concentrate in areas with low number of reviewed publications, as shown in Fig. 10 (b). However, there appears to be no explicit relationship between the size of the enzyme, measured by the number of amino acids, and the accuracy of LLM predictions. This outcome suggests that the predictive ability of LLMs for enzymes primarily relies on information retrieval, specifically from scientific literature reports, rather than on the direct analysis of the enzyme's structure.\\

\textbf{(3) The prediction accuracy of LLMs decreases for crystalline materials with larger, more complex compositions.} As the prediction target's gauge complexity increases, indicated by formula complexity, there is a gradual decline in the LLM's prediction accuracy. Crystals comprising more than four elements perform significantly worse than those with fewer components. Additionally, when the number of formula atoms exceeds 10, the overall accuracy for single crystal predictions generally falls below 15\%, as shown in Fig. 10 (c). Apart from the intrinsic complexity of crystals to bring difficulty in analysis,,since most prediction tasks for crystalline materials in our datasets do not require inference and mainly rely on data retrieval, the rarity of large crystals in the literature compared to more common crystalline materials increases the difficulty of LLM's information retrieval.\\

\textbf{(4) The prediction accuracy of LLMs concerning crystalline materials demonstrates a notable correlation with unit cell symmetry.} Specifically, crystals belonging to the Trigonal, Cubic, or Hexagonal lattice systems are more likely to yield better predictions. The reason for this is twofold: first, these structures are inherently more regular and defined, making them easier subjects for inferential analysis. Secondly, these types of crystal structures are more readily studied and characterized by modern crystallography instruments and techniques, such as X-ray diffraction and electron microscopy, leading to a richer presence in scientific literature. This abundance of data enhances the LLM's ability to retrieve relevant information, thereby improving prediction accuracy for crystals with these symmetries.\\

\textbf{In conclusion, these empirical evidences presented supports the intuitive notion that domain-knowledge embedded prompts enhance the performance of LLMs to different extents.} Firstly, the prompts' inferential capabilities are closely tied to the complexity of the analytical subject matter. Secondly, their proficiency in retrieval is correlated with the depth of contemporary academic research, suggesting that well-crafted prompts can effectively mine the latent knowledge absorbed during the LLM's pre-training phase.

Ultimately, these findings pose future challenges for leveraging LLMs to aid scientific inquiries into complex and novel molecules that are rarely encountered or underrepresented in the academic literature. By addressing these challenges, LLMs could potentially revolutionize the approach to research in the synthesis, analysis, and applications of such molecules, thereby expanding the frontiers of scientific knowledge.

\subsection{Case Studies}

\begin{figure}[ht]
  \centering
  \includegraphics[width=0.9\linewidth]{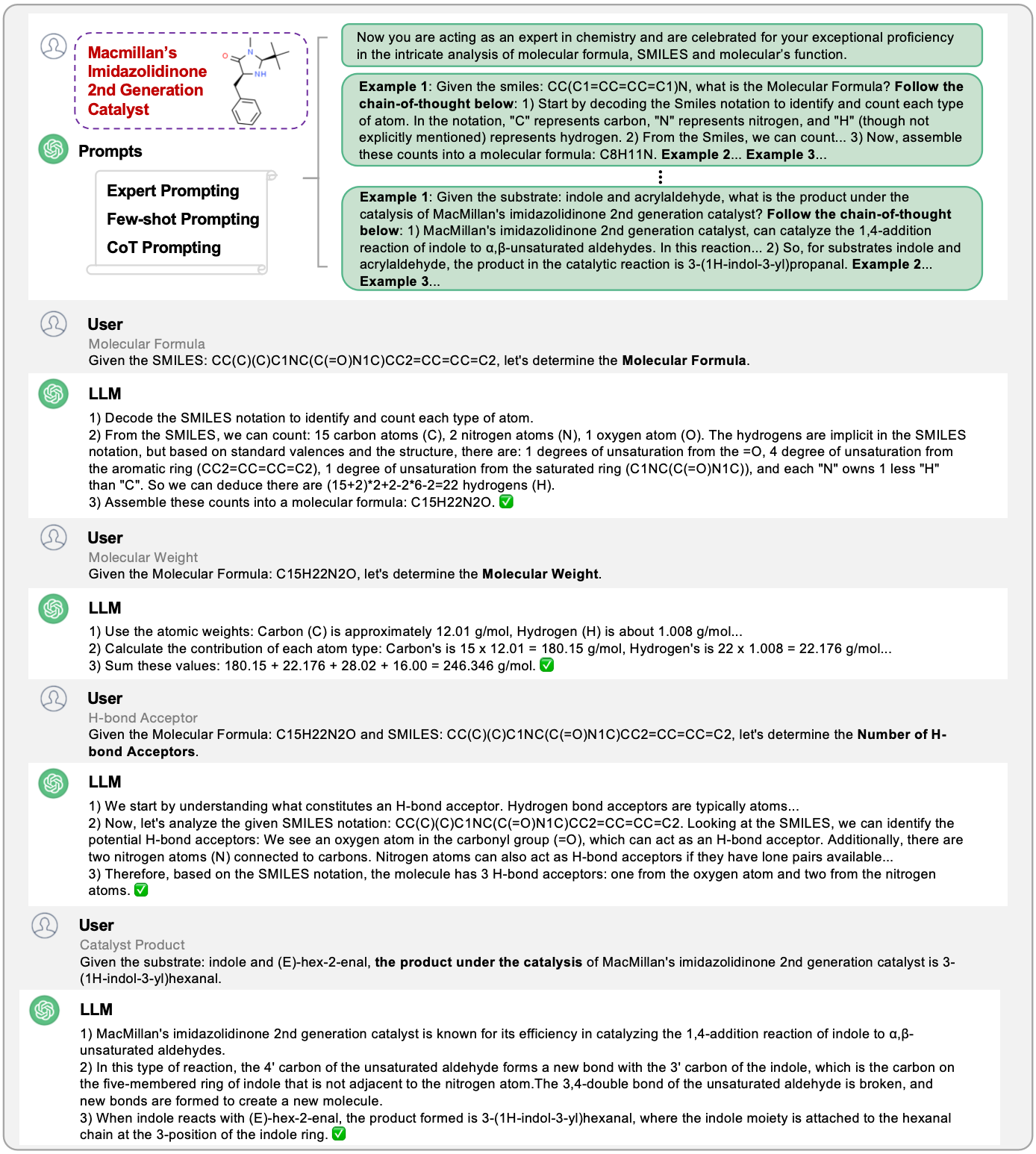}
  \caption{Prompt Engineering Case Study on MacMillan's Imidazolidinone 2nd Generation Catalyst}
\end{figure}

To elucidate the efficacy of the domain-knowledge embedded prompt engineering method in addressing highly domain-specific tasks, we have meticulously designed three case studies. These studies centerpiece the investigation of three materials of profound chemical or biological importance, both in terms of academic research and industrial applications, utilizing our bespoke prompt engineering method that incorporates chemistry-specific domain knowledge. To enhance clarity and conciseness, we illustrate a single expert's prompt engineering workflow, omitting the assembly of contributions from multiple experts, as this singular demonstration already effectively showcases how our prompt engineering method significantly impacts the performance of the LLM.

In the first case study, we direct our attention to the \textbf{MacMillan's imidazolidinone 2nd generation catalyst}, (2S,5S)-(-)-2-tert-Butyl-3-methyl-5-benzyl-4-imidazolidinone. The MacMillan catalyst, a groundbreaking advancement in the field of chemistry, was distinguished by the Nobel Prize in Chemistry in 2021 for its seminal contributions to the development of organocatalysis [59-60]. This innovation has had a transformative impact on both synthetic chemistry and the broader chemical industry, enabling more efficient and environmentally friendly catalytic processes that are pivotal in the synthesis of complex molecules.

The first case study aims to assess the capability of LLMs in assimilating the intricate details of this molecule and in delineating its potential applications. By employing our domain-knowledge embedded prompt engineering method, we seek to uncover how LLMs can be leveraged to provide insights into the reactivity, selectivity, and scope of application of the MacMillan catalyst, thereby enhancing the efficiency and productivity of chemical research in this area.

\begin{figure}[ht]
  \centering
  \includegraphics[width=0.9\linewidth]{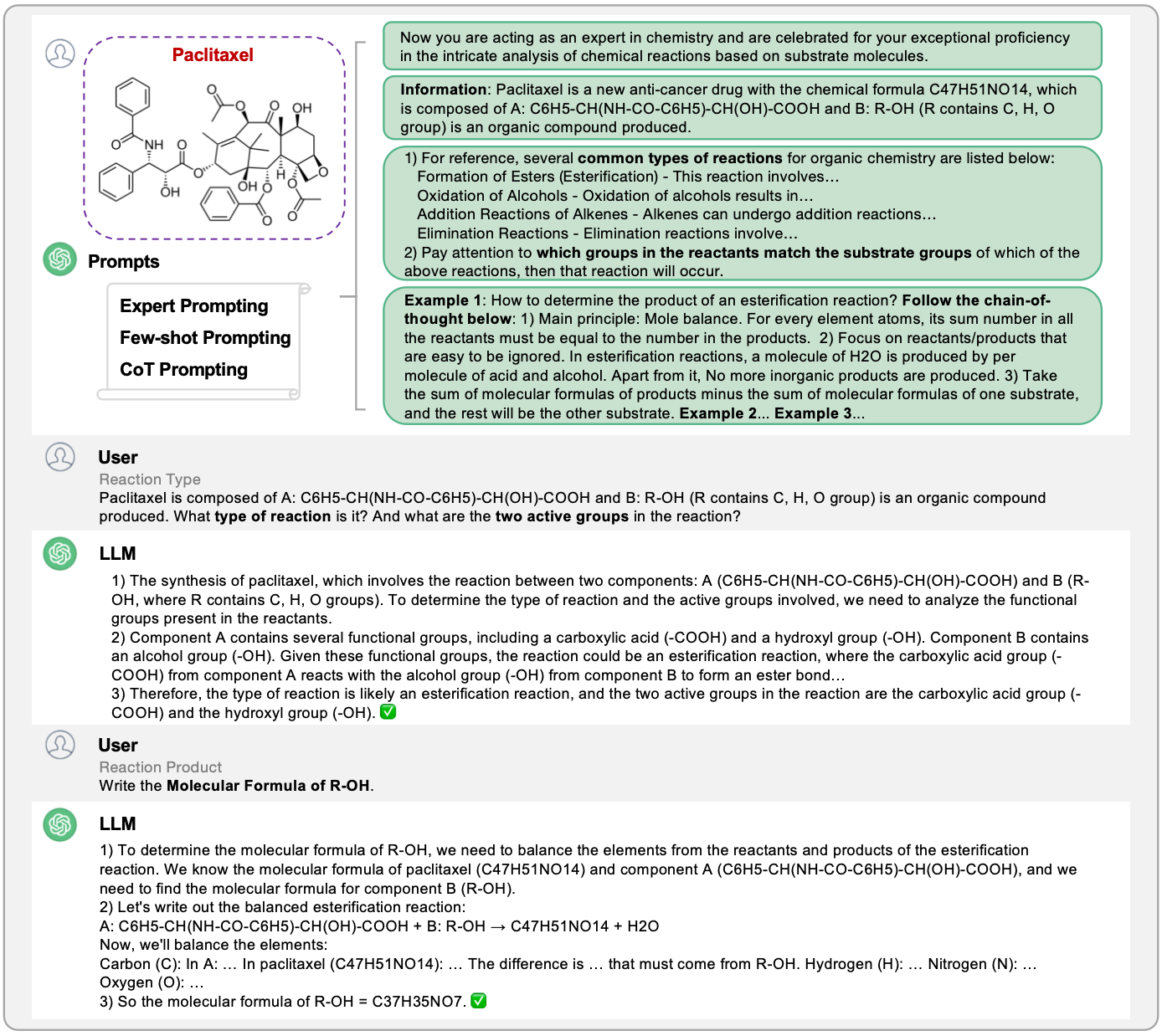}
  \caption{Prompt Engineering Case Study on Paclitaxel}
\end{figure}

As shown in Fig. 11, by utilizing our tailored prompts, the LLM effectively elucidated the fundamental attributes of MacMillan's second-generation imidazolidinone catalyst, demonstrating its proficiency in the analysis of SMILES sequences and elementary arithmetic operations. Additionally, armed with the catalyst's mechanism and illustrative examples, the LLM was able to accurately anticipate the catalytic products from specified substrates, thereby highlighting the model's capacity to inform and potentially guide practical and industrial applications of catalysts. In light of this case study, it is evident that LLMs, embedded with domain-knowledge prompts, have the potential to significantly facilitate the development and optimization of catalysts for chemical reactions, thereby enhancing the efficiency and selectivity of synthetic processes in the field of chemistry. The complete interactive dialogue with LLM could be found in Supplementary S.5.

The next material under examination in our case study is \textbf{paclitaxel} (PTX, C47H51NO14), a compound of profound significance in the field of oncology and a critical component in the treatment of various cancers. Paclitaxel's discovery and subsequent development mark a pivotal moment in the history of cancer therapy, as it introduced a novel mechanism of action that targets microtubules, thereby inhibiting the growth and division of cancer cells. Its efficacy in the treatment of breast, ovarian, and other cancers has established paclitaxel as a cornerstone in the chemotherapy arsenal [61]. The importance of paclitaxel extends beyond its direct clinical applications; it has also served as a template for the development of other taxane derivatives and has been a subject of extensive research in organic synthesis [62]. The complex structure of paclitaxel presents a significant challenge in the synthesis process, leading to the development of various strategies to improve yield, reduce cost, and enhance accessibility to this life-saving compound.

In this prompt engineering case study, we focus on a crucial step in the synthesis of an active intermediate of paclitaxel. Our objective is to assess the ability of LLMs to analyze and provide insights into the pathway of organic synthesis. By utilizing our domain-knowledge embedded prompt engineering method, we aim to demonstrate the potential of LLMs in assisting chemists in the design and optimization of synthetic routes for complex molecules, such as paclitaxel and its derivatives, thereby contributing to the advancement of both chemical research and pharmaceutical development.

\begin{figure}[ht]
  \centering
  \includegraphics[width=0.9\linewidth]{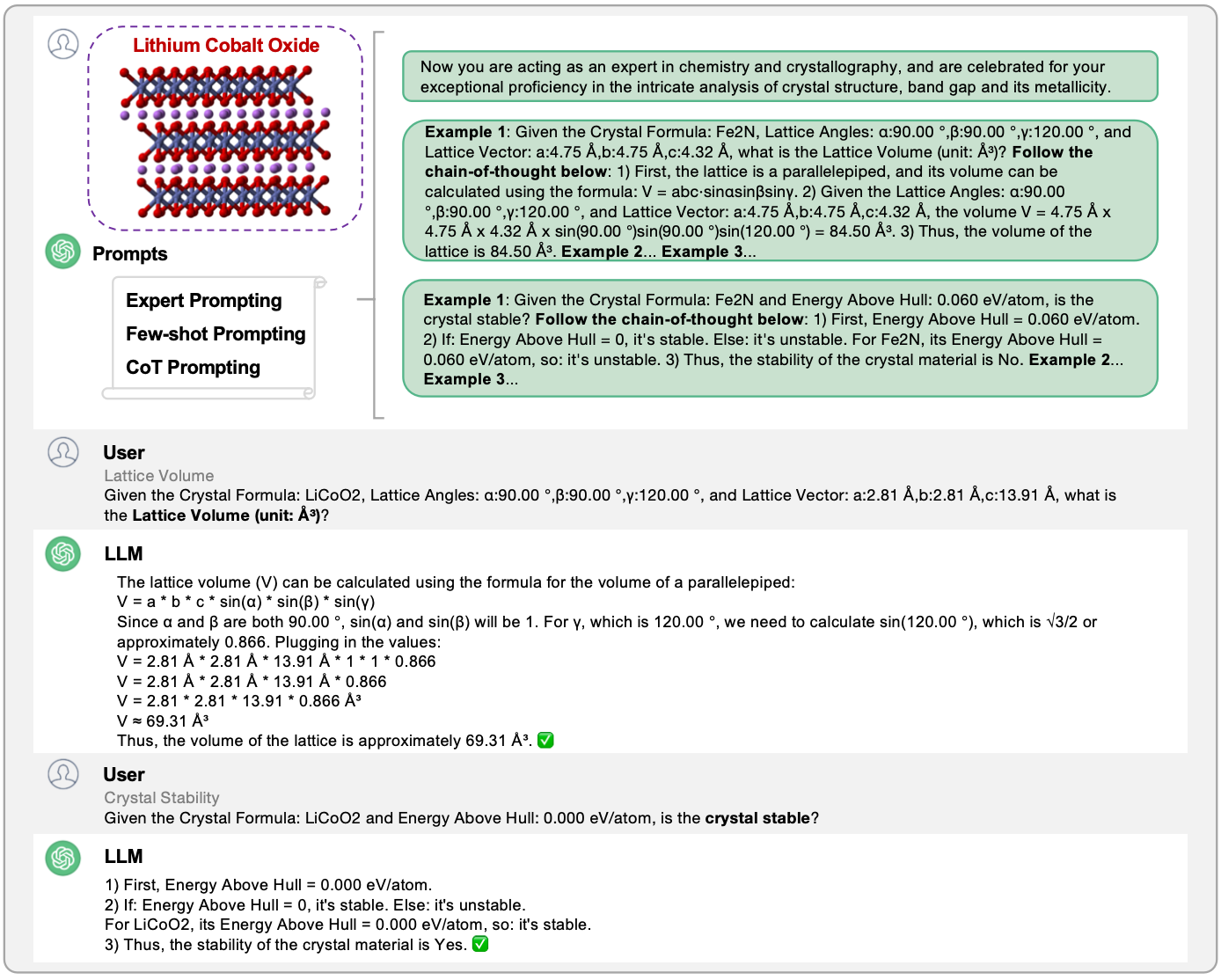}
  \caption{Prompt Engineering Case Study on Lithium Cobalt Oxide}
\end{figure}

As shown in Fig. 12, by utilizing custom-designed prompts, the LLM adeptly dissected a critical step in the synthesis of paclitaxel. It not only identified the reactive groups within the substrates that are capable of engaging in the chemical transformation but also correctly discerned the type of reaction and reconstructed the entire reaction scheme. This accomplishment underscores the LLM's potential in providing guidance for the synthesis of chemical compounds, suggesting that such models could play a pivotal role in streamlining the process of chemical synthesis, offering insights into reaction of complex molecules. This has implications for the advancement of medicinal chemistry and the development of pharmaceuticals, where efficient synthesis routes are of paramount importance. The complete interactive dialogue with LLM could be found in Supplementary S.5.

In the concluding case study, we examine \textbf{lithium cobalt oxide} (LiCoO2), a material of great importance in lithium-ion battery technology. Recognized by the 2018 Nobel Prize in Chemistry, LiCoO2's contribution to energy storage has been transformative, enabling the widespread use of portable electronics and electric vehicles [63]. As a cathode material, LiCoO2 offers high energy density and stability, although research continues to address its lifecycle, cost, and environmental footprint.

In this prompt engineering case study, we delve into the analysis of LiCoO2 crystals and their application advantages. We aim to harness the capabilities of LLMs to provide detailed insights into the crystallographic properties, electrochemical behavior, and optimization strategies for LiCoO2. By employing our domain-knowledge embedded prompt engineering method, we expect to demonstrate the potential of LLMs in aiding researchers in the design and refinement of battery materials, thereby contributing to the progress of energy storage technologies and supporting the global transition towards sustainable energy solutions.

In this case study, the LLM meticulously analyzed the fundamental properties of lithium cobalt oxide (LiCoO2), accurately determining its lattice volume and stability, as shown in Fig. 13. This achievement is of significant importance in the field of crystallography and future development of lithium-ion battery technologies. The complete interactive dialogue with LLM could be found in Supplementary S.5.

\section{Conclusion and Future Directions}

The integration of domain-specific knowledge into prompt engineering has demonstrated its effectiveness in enhancing the performance of LLMs across various tasks in chemistry, materials science, and biology. Our proposed domain-knowledge embedded prompt engineering method outperforms traditional generic prompt engineering strategies on metrics such as capability, accuracy, F1 score, and hallucination drop. The incorporation of domain expertise into prompts not only guides the LLM to synthesize more relevant knowledge but also provides a clear reasoning path for complex tasks. Our case studies further validate the effectiveness of this approach in analyzing intricate materials like the MacMillan catalyst, paclitaxel, and LiCoO2, demonstrating the potential of LLMs to assist experts in molecular design and optimization when equipped with domain-specific prompts. The complete code implementation of our work is listed in Supplementary S.1.

Limitations and potential future directions of our work is also concluded below:

\textbf{Expansion of Domain Coverage}: While our study has focused on chemistry, materials, and biology, the concept of domain-knowledge embedded prompt engineering can be extended to other scientific domains. Future work can explore the development of tailored prompts for fields such as physics, geology, and medicine to unlock the full potential of LLMs in diverse scientific applications.\\

\textbf{Integration of Datasets and Tools}: To further enhance the reasoning capabilities of LLMs, future prompt engineering can integrate external datasets and domain-specific tools. Linking prompts to chemical databases, computational chemistry software, or biological sequence analysis tools could enable the LLM to leverage additional information for more accurate predictions.\\

\textbf{Multi-Modal Prompting}: Incorporating visual information, such as molecular structures or crystal images, into prompts can provide a more intuitive understanding for LLMs. Multi-modal prompting techniques combining textual and visual cues can potentially lead to even stronger performance gains.\\

\textbf{Human-in-the-Loop Refinement}: Iteratively refining prompts with input from domain experts can help to uncover more effective prompting strategies. Human-in-the-loop systems that leverage the complementary strengths of LLMs and human experts have the potential to achieve highly optimized prompts.\\

\textbf{Prompt Engineering Benchmarking}: To ensure comprehensive and fair evaluation of prompting strategies, it is meaningful to establish standardized benchmarks across multiple LLMs, especially the recently released ones. This approach allows researchers to compare the performance of prompt engineering on different LLMs, thereby driving innovation in the field. Creating diverse datasets with a wide range of tasks and molecules will enable robust evaluation and facilitate the development of more effective prompting techniques for various LLMs.\\

In summary, domain-knowledge embedded prompt engineering has shown great promise for unlocking the potential of LLMs in scientific domains. By integrating domain expertise into prompts, LLMs can generate more accurate and contextually relevant responses. As prompt engineering techniques continue to evolve, LLMs have the potential to become powerful allies for scientists, assisting in the exploration and discovery of new materials, molecules, and biological entities.

\section*{References}

{
\small

[1]Wang, S., Fu, J., Liu, Y., Saravanan, R. S., Luo, J., Deng, S., Sham, T.-K., Sun, X., \& Mo, Y. (2023). Design principles for sodium superionic conductors. {\it Nature Communications}, 14(1), 7615.

[2]Dubey, D. K., Thakur, D., Yadav, R. A. K., Ram Nagar, M., Liang, T.-W., Ghosh, S., \& Jou, J.-H. (2021). High-Throughput Virtual Screening of Host Materials and Rational Device Engineering for Highly Efficient Solution-Processed Organic Light-Emitting Diodes. {\it ACS Applied Materials \& Interfaces}, 13(22), 26204–26217.

[3]Medasani, B., Gamst, A., Ding, H., Chen, W., Persson, K. A., Asta, M., Canning, A., \& Haranczyk, M. (2016). Predicting defect behavior in B2 intermetallics by merging ab initio modeling and machine learning. {\it Npj Computational Materials}, 2(1), 1. 

[4]Fetanat, M., Keshtiara, M., Keyikoglu, R., Khataee, A., Daiyan, R., \& Razmjou, A. (2021). Machine learning for design of thin-film nanocomposite membranes. {\it Separation and Purification Technology}, 270, 118383.

[5]Goebel, R., \& Skiborowski, M. (2020). Machine-based learning of predictive models in organic solvent nanofiltration: Pure and mixed solvent flux. {\it Separation and Purification Technology}, 237, 116363.

[6]Guan, J., Huang, T., Liu, W., Feng, F., Japip, S., Li, J., Wu, J., Wang, X., \& Zhang, S. (2022). Design and prediction of metal organic framework-based mixed matrix membranes for CO2 capture via machine learning. {\it Cell Reports Physical Science}, 3(5), 100864.

[7]Senior, A. W., Evans, R., Jumper, J., Kirkpatrick, J., Sifre, L., Green, T., Qin, C., Žídek, A., Nelson, A. W. R., Bridgland, A., Penedones, H., Petersen, S., Simonyan, K., Crossan, S., Kohli, P., Jones, D. T., Silver, D., Kavukcuoglu, K., \& Hassabis, D. (2020). Improved protein structure prediction using potentials from deep learning. {\it Nature}, 577(7792), 706–710.

[8]Jumper, J., Evans, R., Pritzel, A., Green, T., Figurnov, M., Ronneberger, O., Tunyasuvunakool, K., Bates, R., Žídek, A., Potapenko, A., Bridgland, A., Meyer, C., Kohl, S. A. A., Ballard, A. J., Cowie, A., Romera-Paredes, B., Nikolov, S., Jain, R., Adler, J., … Hassabis, D. (2021). Highly accurate protein structure prediction with AlphaFold. {\it Nature}, 596(7873), 583–589.

[9]Lavecchia, A. (2015). Machine-learning approaches in drug discovery: methods and applications. {\it Drug discovery today}, 20(3), 318-331. 

[10]Levy, A. E., Biswas, M., Weber, R., Tarakji, K., Chung, M., Noseworthy, P. A., ... \& Rosenberg, M. A. (2019). Applications of machine learning in decision analysis for dose management for dofetilide. {\it PLoS One}, 14(12), e0227324. 

[11]Schütt, K. T., Gastegger, M., Tkatchenko, A., Müller, K.-R., \& Maurer, R. J. (2019). Unifying machine learning and quantum chemistry with a deep neural network for molecular wavefunctions. {\it Nature Communications}, 10(1), 5024. 

[12]Stöhr, M., Medrano Sandonas, L., \& Tkatchenko, A. (2020). Accurate Many-Body Repulsive Potentials for Density-Functional Tight Binding from Deep Tensor Neural Networks. {\it The Journal of Physical Chemistry Letters}, 11(16), 6835–6843. 

[13]Veríssimo, G. C., dos Santos Junior, V. S., Fernandes, P. O., Ishida, S., Kojima, R., Okuno, Y., ... \& Maltarollo, V. G. (2022). GCN-Based Structure-Activity Relationship and DFT Studies of Staphylococcus aureus FabI Inhibitors. {\it International Journal of Quantitative Structure-Property Relationships (IJQSPR)}, 7(1), 1-16.

[14]Gil, V. H. C., \& Rowley, C. (2024). Graph Neural Networks for Identifying Protein-Reactive Compounds. {\it ChemRxiv}. doi:10.26434/chemrxiv-2023-d0dqp-v2. 

[15]Ziatdinov, M., Dyck, O., Maksov, A., Li, X., Sang, X., Xiao, K., Unocic, R. R., Vasudevan, R., Jesse, S., \& Kalinin, S. v. (2017). Deep Learning of Atomically Resolved Scanning Transmission Electron Microscopy Images: Chemical Identification and Tracking Local Transformations. {\it ACS Nano}, 11(12), 12742–12752. 

[16]Li, J., Telychko, M., Yin, J., Zhu, Y., Li, G., Song, S., Yang, H., Li, J., Wu, J., Lu, J., \& Wang, X. (2021). Machine Vision Automated Chiral Molecule Detection and Classification in Molecular Imaging. {\it Journal of the American Chemical Society}, 143(27), 10177–10188. 

[17]Sanchez-Lengeling, B., \& Aspuru-Guzik, A. (2018). Inverse molecular design using machine learning: Generative models for matter engineering. {\it Science}, 361(6400), 360–365. 

[18]Gebauer, N. W. A., Gastegger, M., Hessmann, S. S. P., Müller, K.-R., \& Schütt, K. T. (2022). Inverse design of 3d molecular structures with conditional generative neural networks. {\it Nature Communications}, 13(1), 973. 

[19]Weiss, T., Mayo Yanes, E., Chakraborty, S., Cosmo, L., Bronstein, A. M., \& Gershoni-Poranne, R. (2023). Guided diffusion for inverse molecular design. {\it Nature Computational Science}, 3(10), 873–882. 

[20]Wong, F., Zheng, E. J., Valeri, J. A., Donghia, N. M., Anahtar, M. N., Omori, S., Li, A., Cubillos-Ruiz, A., Krishnan, A., Jin, W., Manson, A. L., Friedrichs, J., Helbig, R., Hajian, B., Fiejtek, D. K., Wagner, F. F., Soutter, H. H., Earl, A. M., Stokes, J. M., … Collins, J. J. (2024). Discovery of a structural class of antibiotics with explainable deep learning. {\it Nature}, 626(7997), 177–185. 

[21]Watson, J. L., Juergens, D., Bennett, N. R., Trippe, B. L., Yim, J., Eisenach, H. E., Ahern, W., Borst, A. J., Ragotte, R. J., Milles, L. F., Wicky, B. I. M., Hanikel, N., Pellock, S. J., Courbet, A., Sheffler, W., Wang, J., Venkatesh, P., Sappington, I., Torres, S. V., … Baker, D. (2023). De novo design of protein structure and function with RFdiffusion. {\it Nature}, 620(7976), 1089–1100. 

[22]Vázquez Torres, S., Leung, P. J. Y., Venkatesh, P., Lutz, I. D., Hink, F., Huynh, H.-H., Becker, J., Yeh, A. H.-W., Juergens, D., Bennett, N. R., Hoofnagle, A. N., Huang, E., MacCoss, M. J., Expòsit, M., Lee, G. R., Bera, A. K., Kang, A., de La Cruz, J., Levine, P. M., … Baker, D. (2024). De novo design of high-affinity binders of bioactive helical peptides. {\it Nature}, 626(7998), 435–442. 

[23]Burger, B., Maffettone, P. M., Gusev, V. v, Aitchison, C. M., Bai, Y., Wang, X., Li, X., Alston, B. M., Li, B., Clowes, R., Rankin, N., Harris, B., Sprick, R. S., \& Cooper, A. I. (2020). A mobile robotic chemist. {\it Nature}, 583(7815), 237–241.

[24]Jang, W. D., Kim, G. B., Kim, Y., \& Lee, S. Y. (2022). Applications of artificial intelligence to enzyme and pathway design for metabolic engineering. {\it Current Opinion in Biotechnology}, 73, 101-107.

[25]Bannwarth, C., Ehlert, S., \& Grimme, S. (2019). GFN2-xTB—An Accurate and Broadly Parametrized Self-Consistent Tight-Binding Quantum Chemical Method with Multipole Electrostatics and Density-Dependent Dispersion Contributions. {\it Journal of Chemical Theory and Computation}, 15(3), 1652–1671. 

[26]Jha, G., \& Heine, T. (2023). Spin–orbit coupling corrections for the GFN-xTB method. {\it The Journal of Chemical Physics}, 158(4), 044120. 

[27]Zeng, J., Zhang, D., Lu, D., Mo, P., Li, Z., Chen, Y., Rynik, M., Huang, L., Li, Z., Shi, S., Wang, Y., Ye, H., Tuo, P., Yang, J., Ding, Y., Li, Y., Tisi, D., Zeng, Q., Bao, H., … Wang, H. (2023). DeePMD-kit v2: A software package for deep potential models. {\it The Journal of Chemical Physics}, 159(5), 54801. 

[28]Slattery, A., Wen, Z., Tenblad, P., Sanjosé-Orduna, J., Pintossi, D., den Hartog, T., \& Noël, T. (2024). Automated self-optimization, intensification, and scale-up of photocatalysis in flow. {\it Science}, 383(6681), eadj1817. 

[29]Szymanski, N. J., Rendy, B., Fei, Y., Kumar, R. E., He, T., Milsted, D., McDermott, M. J., Gallant, M., Cubuk, E. D., Merchant, A., Kim, H., Jain, A., Bartel, C. J., Persson, K., Zeng, Y., \& Ceder, G. (2023). An autonomous laboratory for the accelerated synthesis of novel materials. {\it Nature}, 624(7990), 86–91. 

[30]Xu, S., Li, J., Cai, P., Liu, X., Liu, B., \& Wang, X. (2021). Self-Improving Photosensitizer Discovery System via Bayesian Search with First-Principle Simulations. {\it Journal of the American Chemical Society}, 143(47), 19769–19777. 

[31]Gao, H., Zhong, S., Zhang, W., Igou, T., Berger, E., Reid, E., Zhao, Y., Lambeth, D., Gan, L., Afolabi, M. A., Tong, Z., Lan, G., \& Chen, Y. (2022). Revolutionizing Membrane Design Using Machine Learning-Bayesian Optimization. {\it Environmental Science \& Technology}, 56(4), 2572–2581. 

[32]Rao, Z., Tung, P.-Y., Xie, R., Wei, Y., Zhang, H., Ferrari, A., Klaver, T. P. C., Körmann, F., Sukumar, P. T., Kwiatkowski da Silva, A., Chen, Y., Li, Z., Ponge, D., Neugebauer, J., Gutfleisch, O., Bauer, S., \& Raabe, D. (2022). Machine learning–enabled high-entropy alloy discovery. {\it Science}, 378(6615), 78–85. 

[33]Bran, A. M., Cox, S., Schilter, O., Baldassari, C., White, A. D., \& Schwaller, P. (2023). ChemCrow: Augmenting large-language models with chemistry tools. {\it ArXiv Preprint}, arXiv:2304.05376. 

[34]Ross, J., Belgodere, B., Chenthamarakshan, V., Padhi, I., Mroueh, Y., \& Das, P. (2022). Large-scale chemical language representations capture molecular structure and properties. {\it Nature Machine Intelligence}, 4(12), 1256–1264. 

[35]Merchant, A., Batzner, S., Schoenholz, S. S., Aykol, M., Cheon, G., \& Cubuk, E. D. (2023). Scaling deep learning for materials discovery. {\it Nature}, 624(7990), 80–85. 

[36]Wei, J., Wang, X., Schuurmans, D., Bosma, M., ichter,  brian, Xia, F., Chi, E., Le, Q. v, \& Zhou, D. (2022). Chain-of-Thought Prompting Elicits Reasoning in Large Language Models. In S. Koyejo, S. Mohamed, A. Agarwal, D. Belgrave, K. Cho, \& A. Oh (Eds.), {\it Advances in Neural Information Processing Systems} (Vol. 35, pp. 24824–24837). Curran Associates, Inc. 

[37]Ahmed, T., \& Devanbu, P. (2023). Few-shot training LLMs for project-specific code-summarization. {\it Proceedings of the 37th IEEE/ACM International Conference on Automated Software Engineering}.

[38]White, J., Fu, Q., Hays, S., Sandborn, M., Olea, C., Gilbert, H., Elnashar, A., Spencer-Smith, J., \& Schmidt, D. C. (2023). A Prompt Pattern Catalog to Enhance Prompt Engineering with ChatGPT. {\it ArXiv Preprint}, arXiv:2302.11382. 

[39]Zhou, Y., Ioan Muresanu, A., Han, Z., Paster, K., Pitis, S., Chan, H., \& Ba, J. (2022). Large Language Models Are Human-Level Prompt Engineers. {\it ArXiv Preprint}, arXiv:2211.01910.

[40]Ekin, S. (2023). Prompt engineering for chatgpt: a quick guide to techniques, tips, and best practices. \url{https://doi.org/10.36227/techrxiv.22683919.v2}

[41]Xie, T., Wan, Y., Huang, W., Yin, Z., Liu, Y., Wang, S., Linghu, Q., Kit, C., Grazian, C., Zhang, W., Razzak, I., \& Hoex, B. (2023). DARWIN Series: Domain Specific Large Language Models for Natural Science. {\it ArXiv Preprint}, arXiv:2308.13565. 

[42]Giray, L. (2023). Prompt Engineering with ChatGPT: A Guide for Academic Writers. {\it Annals of Biomedical Engineering}, 51(12), 2629–2633.

[43]Meskó, B. (2023). Prompt Engineering as an Important Emerging Skill for Medical Professionals: Tutorial. {\it J Med Internet Res}, 25, e50638.

[44]Schreiber, S. L. (2011). Organic synthesis toward small-molecule probes and drugs. {\it Proceedings of the National Academy of Sciences}, 108(17), 6699–6702. 

[45]Kirk, O., Borchert, T. V., \& Fuglsang, C. C. (2002). Industrial enzyme applications. {\it Current Opinion in Biotechnology}, 13(4), 345–351. 

[46]Sharma, A., Gupta, G., Ahmad, T., Mansoor, S., \& Kaur, B. (2021). Enzyme Engineering: Current Trends and Future Perspectives. {\it Food Reviews International}, 37(2), 121–154. 

[47]Surek, T. (2005). Crystal growth and materials research in photovoltaics: progress and challenges. {\it Journal of Crystal Growth}, 275(1), 292–304. 

[48]Zhang, C., Zhang, J., Ma, X., \& Feng, Q. (2021). Semiconductor photovoltaic cells. {\it Springer}. 

[49]Hendrycks, D., Burns, C., Basart, S., Zou, A., Mazeika, M., Song, D., \& Steinhardt, J. (2020). Measuring Massive Multitask Language Understanding. {\it ArXiv E-Preprint}, arXiv:2009.03300.

[50]Srivastava, A., Rastogi, A., Rao, A., Shoeb, A. A. M., Abid, A., Fisch, A., Brown, A. R., Santoro, A., Gupta, A., Garriga-Alonso, A., Kluska, A., Lewkowycz, A., Agarwal, A., Power, A., Ray, A., Warstadt, A., Kocurek, A. W., Safaya, A., Tazarv, A., … Wu, Z. (2022). Beyond the Imitation Game: Quantifying and extrapolating the capabilities of language models. {\it ArXiv E-Preprint}, arXiv:2206.04615.

[51]Cobbe, K., Kosaraju, V., Bavarian, M., Chen, M., Jun, H., Kaiser, L., Plappert, M., Tworek, J., Hilton, J., Nakano, R., Hesse, C., \& Schulman, J. (2021). Training Verifiers to Solve Math Word Problems. {\it ArXiv E-Preprint}, arXiv:2110.14168.

[52]Kim, S., Thiessen, P. A., Bolton, E. E., Chen, J., Fu, G., Gindulyte, A., Han, L., He, J., He, S., Shoemaker, B. A., Wang, J., Yu, B., Zhang, J., \& Bryant, S. H. (2016). PubChem Substance and Compound databases. {\it Nucleic Acids Research}, 44(D1), D1202–D1213. 

[53]Consortium, T. U. (2019). UniProt: a worldwide hub of protein knowledge. {\it Nucleic Acids Research}, 47(D1), D506–D515. 

[54]Jain, A., Ong, S. P., Hautier, G., Chen, W., Richards, W. D., Dacek, S., Cholia, S., Gunter, D., Skinner, D., Ceder, G., \& Persson, K. A. (2013). Commentary: The Materials Project: A materials genome approach to accelerating materials innovation. {\it APL Materials}, 1(1), 011002. 

[55]Brown, T., Mann, B., Ryder, N., Subbiah, M., Kaplan, J. D., Dhariwal, P., Neelakantan, A., Shyam, P., Sastry, G., Askell, A., Agarwal, S., Herbert-Voss, A., Krueger, G., Henighan, T., Child, R., Ramesh, A., Ziegler, D., Wu, J., Winter, C., … Amodei, D. (2020). Language Models are Few-Shot Learners. In H. Larochelle, M. Ranzato, R. Hadsell, M. F. Balcan, \& H. Lin (Eds.), {\it Advances in Neural Information Processing Systems} (Vol. 33, pp. 1877–1901). Curran Associates, Inc.

[56]Zhang, S. J., Florin, S., Lee, A. N., Niknafs, E., Marginean, A., Wang, A., Tyser, K., Chin, Z., Hicke, Y., Singh, N., Udell, M., Kim, Y., Buonassisi, T., Solar-Lezama, A., \& Drori, I. (2023). Exploring the MIT Mathematics and EECS Curriculum Using Large Language Models. {\it ArXiv E-Preprint}, arXiv:2306.08997. 

[57]Chu, Z., Chen, J., Chen, Q., Yu, W., He, T., Wang, H., Peng, W., Liu, M., Qin, B., \& Liu, T. (2023). A Survey of Chain of Thought Reasoning: Advances, Frontiers and Future. {\it ArXiv Preprint}, arXiv:2309.15402.

[58]OpenAI. (2023). OpenAI’s Documentation for Language Models. \url{https://platform.openai.com/docs/models/models}

[59]MacMillan, D. W. C. (2008). The advent and development of organocatalysis. {\it Nature}, 455(7211), 304–308. 

[60]Deepa, \& Singh, S. (2021). Recent Development of Recoverable MacMillan Catalyst in Asymmetric Organic Transformations. {\it Advanced Synthesis \& Catalysis}, 363(3), 629–656. 

[61]Markman, M., \& Mekhail, T. M. (2002). Paclitaxel in cancer therapy. {\it Expert Opinion on Pharmacotherapy}, 3(6), 755–766. 

[62]Mosca, L., Ilari, A., Fazi, F., Assaraf, Y. G., \& Colotti, G. (2021). Taxanes in cancer treatment: Activity, chemoresistance and its overcoming. {\it Drug Resistance Updates}, 54, 100742. 

[63]Wu, Q., Zhang, B., \& Lu, Y. (2022). Progress and perspective of high-voltage lithium cobalt oxide in lithium-ion batteries. {\it Journal of Energy Chemistry}, 74, 283–308.

[64]Goutte, C., \& Gaussier, E. (2005). A Probabilistic Interpretation of Precision, Recall and F-Score, with Implication for Evaluation. In D. E. Losada \& J. M. Fernández-Luna (Eds.), {\it Advances in Information Retrieval} (pp. 345–359). Springer Berlin Heidelberg. 
}

\renewcommand{\thesection}{S}
\renewcommand{\thesubsection}{S.\arabic{subsection}}
\section*{Supplementary Materials}

\subsection{Link to Datasets, Code and Models}

Links to the source code, test datasets, and prompt engineering models are available here: \url{https://github.com/harrylaucngd/prompt-eng-master}. The complete predictive results of different models and full results are also available in the link.

\subsection{Dataset Construction Details}

We collect and curate a comprehensive dataset of \textbf{1280} questions and corresponding solutions for evaluation of LLM’s performances. Our focus is centered on three distinct classes of materials that are prevalently utilized in diverse chemical engineering domains: organic small molecules, enzymes, and crystal materials. These prediction tasks in the three major fields are summarized in Table 2.

The selection of these three types of materials takes into account both their spatial scale distribution and their significance in scientific research. Organic small molecules, enzymes, and crystalline materials represent the three distinct scales of micro-molecules, macro-molecules, and macroscopic materials, respectively. This diversity of scales ensures that our study encapsulates the complexities prevalent in materials science.

\begin{table}
\centering
\caption{Prompt Engineering Prediction Tasks}
\label{tab:question_classification}
\begin{tabular}{p{2.3cm} p{7cm} p{1.5cm} p{1.5cm}}
\toprule
Datasets & Tasks & Number of Molecules & Number of Tasks \\
\midrule
\textbf{Crystal Material} & Space Group Number, Lattice Angle ($\alpha$,$\beta$,$\gamma$), Lattice Vector (a,b,c), Density, Formation Energy, Energy Above Hull, Stability, Band Gap, Direct Gap, Metallic, Total Magnetization, Ordering & 40 & 640 \\
\textbf{Organic Small Molecule} & Molecular Formula, Melting Point, Density, Solubility, Molecular Weight, H-bond Acceptors, H-bond Donors, LogP, Drugability & 40 & 360 \\
\textbf{Enzyme} & Category, Substrate, Product, Active Site, Biological Process, Number of Amino Acids, Ligand & 40 & 280 \\
\textbf{Total} & 32 & 120 & 1280 \\
\bottomrule
\end{tabular}
\end{table}

Furthermore, each of these material categories holds significant relevance in academic research and practical applications. Organic small molecules are commonly utilized in pharmaceutical development and as molecular probes [44]. Enzymes play a critical role in biocatalysis and therapeutic interventions [45-46], while crystalline materials are essential in semiconductor technology and photovoltaic devices [47-48].

We have devised a series of prediction tasks for three types of materials covering a wide range of properties and structural information.

\begin{itemize}

\item \textbf{Organic small molecules}. Derived from PubChem [52], this material class contains \textbf{40} organic small molecules proven to have significant drug properties or potentials. We focus on elucidating the structural and physical-chemical properties that are crucial in determining the functionality and applicability of Organic small molecules, such as \textbf{Molecular Formula}, \textbf{Melting Point}, \textbf{Density}, \textbf{Solubility in Water}, \textbf{Molecular Weight}, \textbf{H-bond donor and acceptor}, \textbf{LogP}, and \textbf{Drugability}.

\item \textbf{Enzymes}. We curate \textbf{40} proteins involved in significant metabolic pathways in vivo from the UniProt database [53], a high-quality and freely accessible database of protein sequence and functional information from the scientific literature and computational analysis. We tested enzymes' efficiency, stability, and suitability for industrial and therapeutic applications. The key performance to be examined include, but are not limited to: \textbf{Category}, \textbf{Number of Amino Acids}, \textbf{Substrate}, \textbf{Product}, \textbf{Active Site}, \textbf{Biological Process} and \textbf{Ligand}.

\item \textbf{Crystal Materials}. We curate \textbf{40} representative crystals from the Materials Project database [54], an open-access database providing information on material properties and crystal structures powered by high-throughout computational techniques. We are primarily concerned with crystal materials' structural and energy properties, which are fundamental in determining their functionality and efficiency in practical applications, such as: \textbf{Space Group Number}, \textbf{Lattice Angle}, \textbf{Lattice Vector}, \textbf{Density}, \textbf{Formation Energy}, \textbf{Energy Above Hull}, \textbf{Stability}, \textbf{Band Gap}, \textbf{Direct Gap}, \textbf{Metallic}, \textbf{Total Magnetization}, \textbf{Ordering}.

\end{itemize}

We categorize these 32 sorts of tasks (16 in crystal materials, 9 in small molecules, and 7 in enzymes) into groups based on output type, reasoning paradigm, and Chain-of-Thought (CoT) complexity denoted with the quantity of additional properties involved in CoT prompts.

\begin{figure}[ht]
  \centering
  \includegraphics[width=0.9\linewidth]{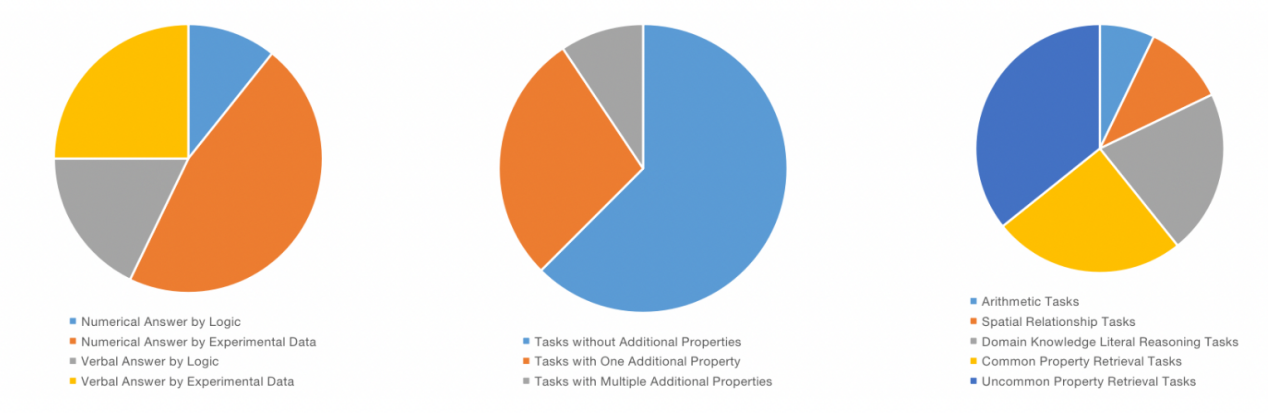}
  \caption{Question Proportion by Output Type, Reasoning Paradigm and CoT Complexity}
\end{figure}

\textbf{Groups Based on Output Type}

\begin{itemize}

\item \textbf{Numerical Answer by Logic}: Tasks where answers are deduced through pure logical reasoning, with the output taking the form of numerical values.

\item \textbf{Numerical Answer by Experimental Data}: Tasks that rely on external experimental data for answers which take the form of numerical values.

\item \textbf{Verbal Answer by Logic}: Tasks where answers are descriptive and are derived from logical reasoning.

\item \textbf{Verbal Answer by Experimental Data}: Tasks where answers are descriptive and based on external experimental data.

\end{itemize}

\begin{table}
\centering
\caption{Question Classification Based on Output Type}
\label{tab:question_classification}
\begin{tabular}{p{4cm} p{7cm} l}
\toprule
Group & Tasks & Total \\
\midrule
\textbf{Numerical Answer by Logic} & Molecular Weight, Number of H-bond Acceptors, Number of H-bond Donors & 120 \\
\textbf{Numerical Answer by Experimental Data} & Lattice Angle, Lattice Vector, Space Group Number, Number of Amino Acids, Melting Point, Density, Solubility, LogP, Crystal Density, Formation Energy, Energy Above Hull, Band Gap, Total Magnetization & 680 \\
\textbf{Verbal Answer by Logic} & Category, Substrate, Product, Biological Process, Molecular Formula & 200 \\
\textbf{Verbal Answer by Experimental Data} & Active Site, Ligand, Drugability, Stability, Direct Gap, Metallic, Ordering & 280 \\
\bottomrule
\end{tabular}
\end{table}

\textbf{Groups Based on Reasoning Paradigm}

\begin{itemize}

\item \textbf{Arithmetic Tasks}: Calculation-based tasks.

\item \textbf{Spatial Relationship Tasks}: Tasks involving understanding of spatial arrangements or geometry.

\item \textbf{Domain Knowledge Literal Reasoning Tasks}: Tasks requiring specific domain knowledge for literal reasoning.

\item \textbf{Common Property Retrieval Tasks}: Tasks involving retrieval of common properties.

\item \textbf{Uncommon Property Retrieval Tasks}: Tasks involving retrieval of less common or obscure information.

\end{itemize}

\begin{table}
\centering
\caption{Question Classification Based on Reasoning Paradigm}
\label{tab:question_classification}
\begin{tabular}{p{4cm} p{7cm} l}
\toprule
Group & Tasks & Total \\
\midrule
\textbf{Arithmetic Tasks} & Molecular Weight, Molecular Formula & 80 \\
\textbf{Spatial Relationship Tasks} & Number of H-bond Acceptors, Number of H-bond Donors, Lattice Angle & 200 \\
\textbf{Domain Knowledge Literal Reasoning Tasks} & Category, Substrate, Product, Biological Process, Stability, Drugability & 240 \\
\textbf{Common Property Retrieval Tasks} & Space Group Number, Number of Amino Acids, Ligand, Ordering, Melting Point, Density, Solubility & 280 \\
\textbf{Uncommon Property Retrieval Tasks} & Lattice Vector, Active Site, Crystal Density, Formation Energy, Energy Above Hull, Band Gap, Total Magnetization, LogP, Direct Gap, Metallic & 480 \\
\bottomrule
\end{tabular}
\end{table}

\textbf{Groups Based on CoT Complexity}

\begin{itemize}

\item \textbf{Tasks without Additional Properties}: No additional property involved in CoT prompts.

\item \textbf{Tasks with One Additional Property}: One additional property involved in CoT prompts.

\item \textbf{Tasks with Multiple Additional Properties}: More than one additional properties involved in CoT prompts.

\end{itemize}

\begin{table}
\centering
\caption{Question Classification Based on CoT complexity}
\label{tab:question_classification}
\begin{tabular}{p{4cm} p{7cm} l}
\toprule
Group & Tasks & Total \\
\midrule
\textbf{Tasks without Additional Properties} & Lattice Angle, Space Group Number, Number of Amino Acids, Melting Point, Density, Solubility, LogP, Formation Energy, Energy Above Hull, Band Gap, Total Magnetization, Category, Substrate, Product, Molecular Formula, Active Site, Ligand, Direct Gap & 800 \\
\textbf{Tasks with One Additional Property} & Molecular Weight, Number of H-bond Acceptors, Number of H-bond Donors, Stability, Lattice Vector, Metallic, Ordering & 360 \\
\textbf{Tasks with Multiple Additional Properties} & Drugability, Crystal Density, Biological Process & 120 \\
\bottomrule
\end{tabular}
\end{table}

\subsection{Answer Evaluation Metrics and Remarks}

The detailed formulation of 4 metrics used in answer evaluation are listed as below:

\textbf{1.Capability}

For some tasks that are complicated or heavily dependent on accessing databases, it's possible that LLM would fail to even complete an answer but instead only announcing its lack of capability in completing the specific tasks. We propose "Capability" index to measure LLM’s ability to provide an answer, regardless of correctness, which we believe will be a critical indicator to evaluate performance of different prompt engineering methods.

To calculate capability, the following formula is applied:

\begin{equation}
C=\frac{\Sigma_{i=1}^n\mathbb{I}_{\{\mathrm{successfully}\,\mathrm{answered}\}}}{n}
\end{equation}

where $\mathbb{I}_{\{\mathrm{successfully}\,\mathrm{answered}\}}$ is a characteristic function representing whether each question is successfully answered, and $n$ is the total number of questions.

\textbf{2.Accuracy}

The "Accuracy" metric measures the average correctness of answers across different tasks. For each question, the final score ranges from 0 to 1, where 0 indicates a failure to provide an answer or an answer that is completely off the mark, and 1 signifies a precise, fully correct answer.

Accuracy is calculated using the formula:

\begin{equation}
A=\frac{\Sigma_{i=1}^nA_i}{n}
\end{equation}

where $A_i$ is the accuracy score for each question, and $n$ is the total number of questions.

\textbf{3.F1 Score}

Specifically for classification and multiple choices tasks, F1 score - a statistical measure that combines precision and recall, is used to provide a balanced view of the LLM’s performance [64].

As F1 Score is typically used to evaluate binary classification performance:

\begin{equation}
\mathrm{precision}=\frac{\mathrm{TP}}{\mathrm{TP}+\mathrm{FP}}
\end{equation}

\begin{equation}
\mathrm{recall}=\frac{\mathrm{TP}}{\mathrm{TP}+\mathrm{FN}}
\end{equation}

\begin{equation}
\mathrm{F1}=2\times\frac{\mathrm{precision}\times\mathrm{recall}}{\mathrm{precision}+\mathrm{recall}}
\end{equation}

for multiple choices questions, F1 score for each choice is evaluated respectively and assembled afterwards:

\begin{equation}
\mathrm{precision}(\mathrm{choice}=\mathrm{A})=\frac{\mathrm{TP}(\mathrm{choice}=\mathrm{A})}{\mathrm{TP}(\mathrm{choice}=\mathrm{A})+\mathrm{FP}(\mathrm{choice}=\mathrm{A})}
\end{equation}

\begin{equation}
\mathrm{recall}(\mathrm{choice}=\mathrm{A})=\frac{\mathrm{TP}(\mathrm{choice}=\mathrm{A})}{\mathrm{TP}(\mathrm{choice}=\mathrm{A})+\mathrm{FN}(\mathrm{choice}=\mathrm{A})}
\end{equation}

\begin{equation}
\mathrm{F1}(\mathrm{choice}=\mathrm{A})=2\times\frac{\mathrm{precision}(\mathrm{choice}=\mathrm{A})\times\mathrm{recall}(\mathrm{choice}=\mathrm{A})}{\mathrm{precision}(\mathrm{choice}=\mathrm{A})+\mathrm{recall}(\mathrm{choice}=\mathrm{A})}
\end{equation}

\begin{equation}
\mathrm{F1}=\frac{1}{5}\Sigma_{\mathrm{choice}={A,B,C,D,E}}\mathrm{F1}(\mathrm{choice})
\end{equation}

\textbf{4.Hallucination Drop}

The concept of "Hallucination Drop" is introduced to quantify the discrepancy between an LLM's ability to answer questions (Capability) and the accuracy of those answers (Accuracy). It serves as an indicator of the LLM's tendency to provide unrealistic or incorrect responses, a phenomenon commonly referred to as "hallucination."

Hallucination Drop $H$ for a specific type of task or material is calculated using the formula:

\begin{equation}
H=1-\frac{A}{C}
\end{equation}

The policy of how numerical tasks transform to multiple-choices questions are listed as follows:

\begin{itemize}

\item "Molecular Weight (unit: g/mol)": [2, 10, "(unit: g/mol)"],
\item "Number of H-bond Acceptors": [1, 2, ""],
\item "Number of H-bond Donors": [1, 2, ""],
\item "a in Lattice Vector [a, b, c] (unit: Å)": [2, 1, "(unit: Å)"],
\item "b in Lattice Vector [a, b, c] (unit: Å)": [2, 1, "(unit: Å)"],
\item "c in Lattice Vector [a, b, c] (unit: Å)": [2, 1, "(unit: Å)"],
\item "Lattice Angle $\alpha$ (among 3 angles as [$\alpha$, $\beta$, $\gamma$])": [0, ["$\alpha$<90", "90<=$\alpha$<100", "100<=$\alpha$<110", "110<=$\alpha$<120", "$\alpha$>120"], ""],
\item "Lattice Angle $\beta$ (among 3 angles as [$\alpha$, $\beta$, $\gamma$])": [0, ["$\beta$<90", "90<=$\beta$<100", "100<=$\beta$<110", "110<=$\beta$<120", "$\beta$>120"], ""],
\item "Lattice Angle $\gamma$ (among 3 angles as [$\alpha$, $\beta$, $\gamma$])": [0, ["$\gamma$<90", "90<=$\gamma$<100", "100<=$\gamma$<110", "110<=$\gamma$<120", "$\gamma$>120"], ""],
\item "Space Group Number": [1, 15, ""],
\item "Number of Amino Acids": [2, 50, ""],
\item "Melting Point (unit: ℃)": [2, 20, "(unit: ℃)"],
\item "Density (unit: g/cm3)": [2, 0.1, "(unit: g/cm3)"],
\item "Solubility (in water, unit: mg/L)": [0, ["<1", "1\~10", "10\~100", "100\~1000", ">1000"], "(in water, unit: mg/L)"],
\item "LogP": [2, 0.5, ""],
\item "Crystal Density (unit: g/cm3)": [2, 0.5, "(unit: g/cm3)"],
\item "Formation Energy (unit: eV/atom)": [2, 0.5, "(unit: eV/atom)"],
\item "Energy Above Hull (unit: eV/atom)": [2, 0.05, "(unit: eV/atom)"],
\item "Band Gap (unit: eV)": [2, 0.5, "(unit: eV)"],
\item "Total Magnetization (unit: µB/f.u.)": [2, 1, "(unit: µB/f.u.)"]

\end{itemize}

Notice that for each key-value pair, the 3-tuple contains complete policy for multiple-choices designs. The first element indicate choices type: fixed (0), single-valued (1) or interval-valued (2). The second element indicate the step size between each choices. And the last element represent the unit (if exists). For tuples with first element being 0, the multiple choices would automatically be the second element in the tuple, while for the rest of the tuples, the multiple-choices generating program randomly choose 4 other values or intervals around ground truth based on the step size provided, generating 5 choices in total. It’s guaranteed that adjacent choices will have adjacent values or ranges corresponding to them.

The specific Accuracy evaluation policy for numerical tasks is that full mark (1) is given if the option is exactly the ground truth and a partial score (0.4) is given if the value or range of the chosen option is adjacent to the ground truth. The detailed rules are listed in Table 6:

\begin{table}[h]
\centering
\caption{Accuracy Grading Policy for Numerical Tasks}
\begin{tabular}{@{}cccccc@{}}
\toprule
\diagbox[dir=SE]{Answer}{Ground Truth} & A & B & C & D & E \\ \midrule
A & 1 & 0.4 & 0 & 0 & 0 \\
B & 0.4 & 1 & 0.4 & 0 & 0 \\
C & 0 & 0.4 & 1 & 0.4 & 0 \\
D & 0 & 0 & 0.4 & 1 & 0.4 \\
E & 0 & 0 & 0 & 0.4 & 1 \\
\bottomrule
\end{tabular}
\end{table}

The scoring policy considering different choices’ adjacency is reasonable because adjacent choices also have adjacent values or ranges, as guaranteed by the policy of numerical tasks transforming to multiple-choices questions introduced prior to this.

\subsection{Detailed Prompt Designs for Performance Tasks and Answer Evaluation}

For all scientific prediction tasks, the corresponding domain-knowledge embedded Chain-of-Thought (CoT) prompts are listed in below:

\textbf{Organic Small Molecule}

\begin{itemize}

\item \textbf{"Molecular Weight (unit: g/mol)"}: "Question: For small molecule, given the Smiles: CC(=O)Nc1ccc(O)cc1, what is the Molecular Weight (unit: g/mol)? Answer: 1) First, decode the Smiles notation to identify the number of each type of atom. In the notation, "C" represents carbon, "O" represents oxygen, "N" represents nitrogen, and "H" (though not explicitly mentioned) represents hydrogen. From the Smiles, we can count: 8 carbon atoms (C), 1 nitrogen atom (N), and 2 oxygen atoms (O). The hydrogens are implicit in the Smiles notation, but based on standard valences, we can deduce there are 9 hydrogens (H). 2) Next, use the atomic weights: Carbon (C) is approximately 12.01 g/mol, Hydrogen (H) is about 1.008 g/mol, Nitrogen (N) is roughly 14.01 g/mol, and Oxygen (O) is around 16.00 g/mol. 3) Calculate the contribution of each atom type: Carbon's is 8 x 12.01 = 96.08 g/mol, Hydrogen's is 9 x 1.008 = 9.072 g/mol, Nitrogen's is 1 x 14.01 = 14.01 g/mol, and Oxygen's is 2 x 16.00 = 32.00 g/mol. 4) Sum these values: 96.08 + 9.072 + 14.01 + 32.00 = 151.162 g/mol. Thus, the molecular weight of the molecule is approximately 151.162 g/mol.",
\item \textbf{"Number of H-bond Acceptors"}: "Question: For small molecule, given the Molecular Formula: C8H9NO2 and Smiles: CC(=O)Nc1ccc(O)cc1, what is the Number of H-bond Acceptors? Answer: 1) we start by understanding what constitutes an H-bond acceptor. Hydrogen bond acceptors are typically atoms with lone pairs that can accept a hydrogen bond. Common H-bond acceptors include oxygen (O) and nitrogen (N) atoms. 2) Now, let's analyze the given Smiles notation: CC(=O)Nc1ccc(O)cc1. In this notation, the symbols represent atoms and their connectivity. Looking at the Smiles, we can identify the potential H-bond acceptors. We see an oxygen atom in the carbonyl group (=O) and another oxygen atom in the phenol group (O). Both of these oxygen atoms can act as H-bond acceptors. 3) Additionally, there's a nitrogen atom (N) connected to a carbon, which can also act as an H-bond acceptor. 4) Therefore, based on the Smiles notation, the molecule has 3 H-bond acceptors: two from the oxygen atoms and one from the nitrogen atom.",
\item \textbf{"Number of H-bond Donors"}: "Question: For small molecule, given the Molecular Formula: C8H9NO2 and Smiles: CC(=O)Nc1ccc(O)cc1, what is the Number of H-bond Donors? Answer: 1) H-bond donors are typically hydrogen atoms that are attached to electronegative atoms like nitrogen (N) or oxygen (O) and can donate a hydrogen bond. 2) Analyzing the given Smiles notation, CC(=O)Nc1ccc(O)cc1, we need to identify such hydrogens. The carbonyl group (=O) does not have a hydrogen attached to the oxygen, so it's not an H-bond donor. However, the phenol group (O) has a hydrogen attached to the oxygen, making it an H-bond donor. 3) The nitrogen (N) in the molecule, being connected to a carbon, likely has a hydrogen attached to it (as the molecular formula suggests there's one more hydrogen than can be accounted for by the carbons and the phenolic oxygen), making it another H-bond donor. 4) Therefore, the molecule has 2 H-bond donors: one from the phenolic oxygen and one from the nitrogen.",
\item \textbf{"Molecular Formula"}: "Question: For small molecule, given the Smiles: CC(=O)Nc1ccc(O)cc1, what is the Molecular Formula? Answer: 1) Start by decoding the Smiles notation to identify and count each type of atom. In the notation, "C" represents carbon, "O" represents oxygen, "N" represents nitrogen, and "H" (though not explicitly mentioned) represents hydrogen. 2) From the Smiles, we can count: 8 carbon atoms (C), 1 nitrogen atom (N), and 2 oxygen atoms (O). The hydrogens are implicit in the Smiles notation, but based on standard valences and the structure, we can deduce there are 9 hydrogens (H). 3) Now, assemble these counts into a molecular formula: C8H9NO2. Thus, the molecular formula of the molecule is C8H9NO2.",
\item \textbf{"Melting Point (unit: ℃)"}: "Question: For small molecule, given the Name: Acetaminophen and Smiles: CC(=O)Nc1ccc(O)cc1, what is the Melting Point (unit: ℃)? Answer: 1) Recognize that Acetaminophen is a common over-the-counter pain reliever and fever reducer, also known as Paracetamol in some regions. The Smiles notation provided confirms the structure of Acetaminophen. 2) Now, to find the melting point, one would typically refer to a trusted chemical database or literature source. 3) Based on existing knowledge up to the last update in January 2022, Acetaminophen has a melting point of approximately 169℃ to 170.5℃. 4) Answer: The melting point of Acetaminophen is approximately 169℃ to 170.5℃.",
\item \textbf{"Density (unit: g/cm3)"}: "Question: For small molecule, given the Name: Acetaminophen and Smiles: CC(=O)Nc1ccc(O)cc1, what is the Density (unit: g/cm3)? Answer: 1) Recognize that Isopropanol, also known as isopropyl alcohol or rubbing alcohol, is a common solvent used in laboratories and households. The Smiles notation CC(C)O confirms its molecular structure. 2) Density is a fundamental physical property that is often documented in various chemical databases and literature because of its extensive use in various applications, including as a solvent and disinfectant. 3) Based on existing knowledge up to the last update in January 2022, the density of Isopropanol at room temperature (25°C) is approximately 0.785 g/cm3. 4) Answer: The density of Isopropanol is approximately 0.785 g/cm3.",
\item \textbf{"Solubility (in water, unit: mg/L)"}: "Question: For small molecule, given the Name: Acetaminophen and Smiles: CC(=O)Nc1ccc(O)cc1, what is the Solubility (in water, unit: mg/L) (A:<1mg/L, B:1\~10mg/L, C:10\~100mg/L, D:100\~1000mg/L, E>1000mg/L)? Answer: 1) First, we need to analyze the SMILES notation, which provides a linear representation of the chemical's molecular structure. By examining the SMILES notation, we can identify key structural features that influence solubility, such as the presence of polar groups, the size and shape of the molecule, and the degree of saturation. Polar groups like hydroxyl (-OH) and amine (-NH2) tend to increase solubility in water, while large hydrophobic regions or high molecular weight often decrease solubility. 2) Next, we consider the chemical's name, which can give additional clues about its structure and functional groups. For instance, the suffix '-ol' in alcohols indicates the presence of a hydroxyl group, which usually enhances solubility. Then, we correlate these structural features with known solubility trends and principles from chemistry, such as 'like dissolves like' and the impact of hydrogen bonding. By integrating this information, we can estimate the solubility range. 3) For example, a small molecule with several polar groups might fall into category C,D or E, whereas a large, non-polar hydrocarbon is likely in category A or B. 4) Answer: Acetaminophen have many strong polar groups like (=O) and (N), so the solubility of Acetaminophen in water at 25°C is >1000mg/L, so the answer is E.",
\item \textbf{"LogP"}: "Question: For small molecule, given the Name: Acetaminophen and Smiles: CC(=O)Nc1ccc(O)cc1, what is the LogP? Answer: 1) Recognize that Acetaminophen, also known as Paracetamol, is a well-known drug. The Smiles notation confirms its structure. 2) LogP, or the partition coefficient between octanol and water, is a measure of a compound's lipophilicity and plays a crucial role in drug design because it can influence a drug's absorption, distribution, metabolism, and excretion. A higher LogP indicates greater lipophilicity. 3) To find the LogP, one would typically consult a trusted chemical database, use computational tools, or refer to specific literature sources that provide this physicochemical property. 4) Based on existing knowledge up to the last update in January 2022, Acetaminophen has a LogP value of approximately 0.46. 5) Answer: The LogP of Acetaminophen is approximately 0.46.",
\item \textbf{"Drugability (Yes or No)"}: "Question: For small molecule, given the Molecular Weight (unit: g/mol): 151.162 g/mol, Number of H-bond Acceptors: 3, Number of H-bond Donors: 2, Solubility (in water, unit: mg/L): 14,000 mg/L and LogP: 0.46, what is the Drugability (Yes or No)? Answer: 1) Drugability is a complex concept that evaluates the likelihood of a molecule to become an orally active drug in humans. 2) The Lipinski's Rule of Five is a commonly used set of guidelines to predict drugability. According to this rule, a molecule is likely to be drugable if it meets the following criteria: Molecular Weight is less than 500 g/mol, Number of H-bond Acceptors is no more than 10, Number of H-bond Donors is no more than 5, LogP is less than 5, and the molecule is not excessively insoluble. 3) Comparing the given properties to these criteria: The Molecular Weight of 151.162 g/mol is well below 500 g/mol, the Number of H-bond Acceptors is 3, which is below 10, the Number of H-bond Donors is 2, which is below 5, the LogP of 0.46 is well below 5, and a Solubility of 14,000 mg/L indicates the molecule is not excessively insoluble. 4) Given that the molecule meets all the criteria, it is likely to be drugable. So answer is Yes."

\end{itemize}

\textbf{Enzyme}

\begin{itemize}

\item \textbf{"Category (Among Oxidoreductases, Transferases, Hydrolases, Lyases, Isomerases, Ligases, and Translocases)"}: "Question: For enzyme, given the Enzyme: D-Glucose-6P dehydrogenase and EC (Enzyme Commission number): 1.1.1.363, what is the Category? Answer: 1) Focus on the first number in EC. If =1, then the enzyme belongs to Oxidoreductases; =2, Transferases; =3, Hydrolases; =4, Lyases; =5, Isomerases; =6, Ligases; =7, Translocases. 2) For D-Glucose-6P dehydrogenase, its EC=1.1.363, the first number is 1. So it's an Oxidoreductase. 3) Answer: So Category = Oxidoreductases.",
\item \textbf{"Number of Amino Acids"}: "Question: For enzyme, given the Enzyme: D-Glucose-6P dehydrogenase and EC (Enzyme Commission number): 1.1.1.363, what is the Number of Amino Acids? Answer: 1) **Context Analysis** The EC number is a numerical classification scheme for enzymes, based on the chemical reactions they catalyze. D-Glucose-6P dehydrogenase is an enzyme that catalyzes the oxidation of glucose-6-phosphate, primarily in the pentose phosphate pathway. 2) **Specify Information** The number of amino acids in an enzyme can be found by looking at its amino acid sequence, which is determined by the gene encoding the enzyme. For D-Glucose-6P dehydrogenase, according to the the last update in January 2022, the number of amino acid is 791. 3) Answer: So Number of Amino Acid = 791.",
\item \textbf{"Substrate"}: "Question: For enzyme, given the Enzyme: D-Glucose-6P dehydrogenase and EC (Enzyme Commission number): 1.1.1.363, what is the Substrate? Answer: 1) **Context Analysis** D-Glucose-6P dehydrogenase is an enzyme that catalyzes the oxidation of glucose-6-phosphate (first number in EC=1, so it's an oxidoreductase), primarily in the pentose phosphate pathway. 2) The name and EC always provides so much to inference. EC-Numbers —> 1 —> Oxidoreductases —> 1.1—> Acting on the CH-OH group of donors —> 1.1.1 — With NAD or NADP aw acceptor. So naturally its substrate is D-Glucose-6P, also known as D-glucose 6-phosphate. 3) Answer: So Substrate = D-glucose 6-phosphate.",
\item \textbf{"Product"}: "Question: For enzyme, given the Enzyme: D-Glucose-6P dehydrogenase and EC (Enzyme Commission number): 1.1.1.363, what is the Product? Answer: 1) **Context Analysis** D-Glucose-6P dehydrogenase is an enzyme that catalyzes the oxidation of glucose-6-phosphate (first number in EC=1, so it's an oxidoreductase), primarily in the pentose phosphate pathway. 2) The name and EC always provides so much to inference. EC-Numbers —> 1 —> Oxidoreductases —> 1.1—> Acting on the CH-OH group of donors —> 1.1.1 — With NAD or NADP aw acceptor. So naturally its product is the dehydrogenase product of D-Glucose-6P on CH-OH group of donors, which is 6-phospho-D-glucono-1,5-lactone. 3) Answer: So Product = 6-phospho-D-glucono-1,5-lactone.",
\item \textbf{"Active Site"}: "Question: For enzyme, given the Enzyme: D-Glucose-6P dehydrogenase and EC (Enzyme Commission number): 1.1.1.363, what is the Active Site? Answer: 1) **Context Analysis** The EC number is a numerical classification scheme for enzymes, based on the chemical reactions they catalyze. D-Glucose-6P dehydrogenase is an enzyme that catalyzes the oxidation of glucose-6-phosphate, primarily in the pentose phosphate pathway. 2) **Specify Information** Information about the active site of an enzyme is typically obtained from its three-dimensional structure. Once the structure is available, the active site can be identified based on the location where substrates are known to bind. This often involves looking at the enzyme in complex with its substrate or an inhibitor. For D-Glucose-6P dehydrogenase, according to the the last update in January 2022, the active site is H-267. 3) Answer: So Active Site = H-267.",
\item \textbf{"Biological Process"}: "Question: For enzyme, given the Enzyme: D-Glucose-6P dehydrogenase and EC (Enzyme Commission number): 1.1.1.363, what is the Biological Process? Answer: 1) **Context Analysis** D-Glucose-6P dehydrogenase is an enzyme that catalyzes the oxidation of glucose-6-phosphate (first number in EC=1, so it's an oxidoreductase), primarily in the pentose phosphate pathway. 2) The name and EC always provides so much to inference. EC-Numbers —> 1 —> Oxidoreductases —> 1.1—> Acting on the CH-OH group of donors —> 1.1.1 — With NAD or NADP aw acceptor. So naturally its substrate is D-Glucose-6P, and its product is the dehydrogenase product of D-Glucose-6P on CH-OH group of donors, which is 6-phospho-D-glucono-1,5-lactone. The dehydrogenation of these two compounds are frequently seen in Pentose phosphate pathway. 3) **Completeness** Pentose pathway belongs to a part of Carbohydrate metabolism in a more general view. 4) Answer: So Biological Process = Carbohydrate metabolism/Pentose phosphate pathway.",
\item \textbf{"Ligand"}: "Question: For enzyme, given the Enzyme: D-Glucose-6P dehydrogenase and EC (Enzyme Commission number): 1.1.1.363, what is the Ligand? Answer: 1) **Context Analysis** D-Glucose-6P dehydrogenase is an enzyme that catalyzes the oxidation of glucose-6-phosphate (first number in EC=1, so it's an oxidoreductase), primarily in the pentose phosphate pathway. 2) The name and EC always provides so much to inference. EC-Numbers —> 1 —> Oxidoreductases —> 1.1—> Acting on the CH-OH group of donors —> 1.1.1 — With NAD or NADP aw acceptor. So naturally its ligand must be NAD/NADP. 3) Answer: So Ligand = NAD/NADP."

\end{itemize}

\textbf{Crystal Material}

\begin{itemize}

\item \textbf{"a in Lattice Vector [a, b, c] (unit: Å)"}: "Question: For crystal material, given the MP-id mp-248, Formula Fe2N and Lattice Angle $\alpha$ = 90°, $\beta$ = 90°, and $\gamma$ = 120°, what is a in Lattice Vector [a, b, c] (unit: Å)? Answer: 1)**Lattice Vector Analysis** The lattice vectors a, b, and c describe the periodicity of the crystal lattice in three dimensions. The lattice angles ($\alpha$, $\beta$, $\gamma$) describe the angles between these vectors. In a crystal system, these parameters define the unit cell's shape and size. 2)**Determining Lattice Vector** The scale of lattice vector is usually around 3\~10 Å. It's very encouraged to give a hypothesis between that range. Here in compound Fe2N, according to existing knowledge up to the last update in January 2022, the Lattice Vector is a:4.75 Å,b:4.75 Å,c:4.32 Å, so a in Lattice Vector [a, b, c] (unit: Å) is 4.75. 3)**Conclusion** so a in Lattice Vector [a, b, c] (unit: Å) = 4.75.",
\item \textbf{"b in Lattice Vector [a, b, c] (unit: Å)"}: "Question: For crystal material, given the MP-id mp-248, Formula Fe2N and Lattice Angle $\alpha$ = 90°, $\beta$ = 90°, and $\gamma$ = 120°, what is b in Lattice Vector [a, b, c] (unit: Å)? Answer: 1)**Lattice Vector Analysis** The lattice vectors a, b, and c describe the periodicity of the crystal lattice in three dimensions. The lattice angles ($\alpha$, $\beta$, $\gamma$) describe the angles between these vectors. In a crystal system, these parameters define the unit cell's shape and size. 2)**Determining Lattice Vector** The scale of lattice vector is usually around 3\~10 Å. It's very encouraged to give a hypothesis between that range. Here in compound Fe2N, according to existing knowledge up to the last update in January 2022, the Lattice Vector is a:4.75 Å,b:4.75 Å,c:4.32 Å, so b in Lattice Vector [a, b, c] (unit: Å) is 4.75. 3)**Conclusion** so b in Lattice Vector [a, b, c] (unit: Å) = 4.75.",
\item \textbf{"c in Lattice Vector [a, b, c] (unit: Å)"}: "Question: For crystal material, given the MP-id mp-248, Formula Fe2N and Lattice Angle $\alpha$ = 90°, $\beta$ = 90°, and $\gamma$ = 120°, what is c in Lattice Vector [a, b, c] (unit: Å)? Answer: 1)**Lattice Vector Analysis** The lattice vectors a, b, and c describe the periodicity of the crystal lattice in three dimensions. The lattice angles ($\alpha$, $\beta$, $\gamma$) describe the angles between these vectors. In a crystal system, these parameters define the unit cell's shape and size. 2)**Determining Lattice Vector** The scale of lattice vector is usually around 3\~10 Å. It's very encouraged to give a hypothesis between that range. Here in compound Fe2N, according to existing knowledge up to the last update in January 2022, the Lattice Vector is a:4.75 Å,b:4.75 Å,c:4.32 Å, so c in Lattice Vector [a, b, c] (unit: Å) is 4.32. 3)**Conclusion** so c in Lattice Vector [a, b, c] (unit: Å) = 4.32.",
\item \textbf{"Lattice Angle $\alpha$ (among 3 angles as [$\alpha$, $\beta$, $\gamma$])"}: "Question: For crystal material, given the MP-id: mp-248 and Formula: Fe2N, what is the Lattice Angle $\alpha$ (among 3 angles as [$\alpha$, $\beta$, $\gamma$])? Answer: 1)**Question Analysis** Determining the lattice angles for Fe2N involves a theoretical analysis of its bonding, electronic structure, and likely packing style, followed by an educated guess of its crystal system and corresponding lattice angles. 2)**Chemical Formula Analysis** Fe (Iron) is a transition metal known for its metallic bonding characteristics, while N (Nitrogen) typically forms covalent or ionic bonds. In Fe2N, we expect a mix of metallic and covalent bonding. 3)**Hypothesizing Crystal System** In compounds like Fe2N, the metallic nature of iron suggests a dense packing arrangement, so given the metallic characteristics and potential covalent interactions, a reasonable guess for the crystal system might be hexagonal, trigonal or tetragonal, as these systems are common for compounds with mixed bonding types. Let's hypothesize that Fe2N crystallizes in a trigonal system. 4)**Determining Lattice Angles** In a trigonal crystal system, the lattice angles are typically $\alpha$ = 90°, $\beta$ = 90°, and $\gamma$ = 120°. This is because the hexagonal system is characterized by two equal axes at right angles to each other, and a third axis (of a different length) at a 120° angle to one of the other axes. 5)**Conclusion** So the Lattice Angle $\alpha$ (among 3 angles as [$\alpha$, $\beta$, $\gamma$]) = 90°.",
\item \textbf{"Lattice Angle $\beta$ (among 3 angles as [$\alpha$, $\beta$, $\gamma$])"}: "Question: For crystal material, given the MP-id: mp-248 and Formula: Fe2N, what is the Lattice Angle $\beta$ (among 3 angles as [$\alpha$, $\beta$, $\gamma$])? Answer: 1)**Question Analysis** Determining the lattice angles for Fe2N involves a theoretical analysis of its bonding, electronic structure, and likely packing style, followed by an educated guess of its crystal system and corresponding lattice angles. 2)**Chemical Formula Analysis** Fe (Iron) is a transition metal known for its metallic bonding characteristics, while N (Nitrogen) typically forms covalent or ionic bonds. In Fe2N, we expect a mix of metallic and covalent bonding. 3)**Hypothesizing Crystal System** In compounds like Fe2N, the metallic nature of iron suggests a dense packing arrangement, so given the metallic characteristics and potential covalent interactions, a reasonable guess for the crystal system might be hexagonal, trigonal or tetragonal, as these systems are common for compounds with mixed bonding types. Let's hypothesize that Fe2N crystallizes in a trigonal system. 4)**Determining Lattice Angles** In a trigonal crystal system, the lattice angles are typically $\alpha$ = 90°, $\beta$ = 90°, and $\gamma$ = 120°. This is because the hexagonal system is characterized by two equal axes at right angles to each other, and a third axis (of a different length) at a 120° angle to one of the other axes. 5)**Conclusion** So the Lattice Angle $\beta$ (among 3 angles as [$\alpha$, $\beta$, $\gamma$]) = 90°.",
\item \textbf{"Lattice Angle $\gamma$ (among 3 angles as [$\alpha$, $\beta$, $\gamma$])"}: "Question: For crystal material, given the MP-id: mp-248 and Formula: Fe2N, what is the Lattice Angle $\gamma$ (among 3 angles as [$\alpha$, $\beta$, $\gamma$])? Answer: 1)**Question Analysis** Determining the lattice angles for Fe2N involves a theoretical analysis of its bonding, electronic structure, and likely packing style, followed by an educated guess of its crystal system and corresponding lattice angles. 2)**Chemical Formula Analysis** Fe (Iron) is a transition metal known for its metallic bonding characteristics, while N (Nitrogen) typically forms covalent or ionic bonds. In Fe2N, we expect a mix of metallic and covalent bonding. 3)**Hypothesizing Crystal System** In compounds like Fe2N, the metallic nature of iron suggests a dense packing arrangement, so given the metallic characteristics and potential covalent interactions, a reasonable guess for the crystal system might be hexagonal, trigonal or tetragonal, as these systems are common for compounds with mixed bonding types. Let's hypothesize that Fe2N crystallizes in a trigonal system. 4)**Determining Lattice Angles** In a trigonal crystal system, the lattice angles are typically $\alpha$ = 90°, $\beta$ = 90°, and $\gamma$ = 120°. This is because the hexagonal system is characterized by two equal axes at right angles to each other, and a third axis (of a different length) at a 120° angle to one of the other axes. 5)**Conclusion** So the Lattice Angle $\gamma$ (among 3 angles as [$\alpha$, $\beta$, $\gamma$]) = 120°.",
\item \textbf{"Space Group Number"}: "Question: For crystal material, given the MP-id mp-248 and Formula Fe2N, what is the Space Group Number? Answer: 1)**Space Group Number Analysis** The space group of a crystal structure describes the symmetry of the structure. It is a combination of translational and point symmetries and is a critical piece of information for understanding the crystal structure. Space groups are numbered from 1 to 230 in the International Tables for Crystallography. 2)**Hypothesizing Crystal System** In compounds like Fe2N, the metallic nature of iron suggests a dense packing arrangement, so given the metallic characteristics and potential covalent interactions, a reasonable guess for the crystal system might be hexagonal, trigonal or tetragonal, as these systems are common for compounds with mixed bonding types. Let's hypothesize that Fe2N crystallizes in a trigonal system. 3)**Determining Space Group Number** In a trigonal crystal system, the Space Group Number is in 142-167. For Fe2N, the Space Group Number is 162. 4)**Conclusion** So the Space Group Number = 162.",
\item \textbf{"Crystal Density (unit: g/cm3)"}: "Question: For crystal material, given the Formula and Lattice vector, what is the Density (unit: g/cm3)? Answer: 1)**Determine the formula weight** alculate the formula weight (molar mass) of the crystal material by summing the atomic weights of all the atoms in its formula unit. 2)**Calculate the Volume of the Unit Cell** Use the lattice vectors to determine the volume of the unit cell. For a unit cell with lattice vectors a, b, and c, the volume V is given by V = abc. 3)**Calculate the Number of Formula Units per Unit Cell** Determine how many formula units are in each unit cell. 4)**Calculate Density** The density $\rho$ can be calculated using the formula: $\rho$ = Formula Weight*Number of Formula Units per Unit Cell / (V*N\_A), where N\_A is Avogadro's number (approximately 6.02*10\^23 mol\^(-1). Ensure that the units are consistent. For instance, if the formula weight is in grams per mole and the volume in cubic centimeters, the density will be in grams per cubic centimeter. 5)**Example** Suppose you have a crystal with a simple cubic structure (which has one formula unit per unit cell), a lattice parameter a = 5 Å, and it's composed of a hypothetical material XYZ with a formula weight of 150 g/mol. The volume of the unit cell is V=a\^3=(0.5*10\^(-9)m)\^3. The density will be calculated as $\rho$ = 150g/mol*1/(V*6.02*10\^23mol\^(-1))=1.99g/cm3.",
\item \textbf{"Formation Energy (unit: eV/atom)"}: "Question: For crystal material, given the MP-id: mp-248 and Formula: Fe2N, what is the Formation Energy (unit: eV/atom)? Answer: 1)**Formation Energy Analysis** In materials science, Formation Energy is a measure of the energy change when a material is formed from its constituent elements in their standard states. It's a fundamental property that indicates the stability of the material; a more negative formation energy typically suggests a more stable compound. 2)**Determining Formation Energy** Formation Energy for a crystal material is usually under 0 eV/atom if it's stable. The absolute value is usually around -2\~2 eV/atom. For pure metal crystal materials, the value will be 0. It's very encouraged to give a hypothesis between that range. Here, based on existing knowledge up to the last update in January 2022, the Formation Energy of Fe2N is -0.149 eV/atom. 3)**Conclusion** So the Formation Energy (unit: eV/atom) = -0.149.",
\item \textbf{"Energy Above Hull (unit: eV/atom)"}: "Question: For crystal material, given the MP-id: mp-248 and Formula: Fe2N, what is the Energy Above Hull (unit: eV/atom)? Answer: 1)**Energy Above Hull Analysis** In materials science, the "Energy Above Hull" is a metric used to assess the stability of a crystal structure. It represents how much energy per atom a given compound is above the most stable phase (the "hull") of its respective phase diagram. 2)**Determining Energy Above Hull** Energy Above Hull for a crystal material is always around 0\~1.50 eV/atom. The more likely this material is stable, the lower the Energy Above Hull is. It's very encouraged to give a hypothesis between that range. Here, based on existing knowledge up to the last update in January 2022, the Energy Above Hull of Fe2N is 0.060 eV/atom. 3)**Conclusion** So the Energy Above Hull (unit: eV/atom) = 0.060.",
\item \textbf{"Stability (Yes or No)"}: "Question: For crystal material, given the MP-id mp-248, Formula Fe2N and Energy Above Hull: 0.060 eV/atom, what is the Stability (Yes or No) ? Answer: 1)**Determining Stability** As long as Energy Above Hull = 0, it's stable. Otherwise it's unstable. For Fe2N, its Energy Above Hull is 0.060 eV/atom, so it's unstable. 2)**Conclusion** So the Stability (Yes or No) = No.",
\item \textbf{"Band Gap (unit: eV)"}: "Question: For crystal material, given the MP-id mp-248 and Formula Fe2N, what is the Band Gap (unit: eV)? Answer: 1)**Band Gap Analysis** In materials science, the band gap is a fundamental property that describes the energy difference between the top of the valence band and the bottom of the conduction band in a material. It determines the material's electrical conductivity, with a larger band gap typically indicating a better insulator. 2)**Determining Formation Energy** Band Gap for a good conductor is usually zero or negligible while for insulator it's usually high for about 0.5~2.5 eV. It's very encouraged to give a hypothesis between that range. Here, based on existing knowledge up to the last update in January 2022, the Band Gap for this metallic conductor Fe2N is 0.00 eV. 3)**Conclusion** So the Band Gap (unit: eV) = 0.",
\item \textbf{"Direct Gap (Yes or No)"}: "Question: For crystal material, given the MP-id mp-248 and Formula Fe2N, what is the Direct Gap (Yes or No)? Answer: 1)**Direct Gap Analysis** A material is said to have a direct band gap if the minimum of the conduction band and the maximum of the valence band occur at the same momentum (k-point) in the Brillouin zone. This contrasts with an indirect band gap, where these points occur at different momenta. 2)**Determining Direct Gap** Fe2N is an iron nitride compound. The electronic properties, including the nature of the band gap, of such compounds are determined by their electronic structure, which in turn is influenced by factors like the arrangement of atoms and the nature of chemical bonding. Here, based on existing knowledge up to the last update in January 2022, the Direct Gap for Fe2N is No. 3)**Conclusion** So the Direct Gap (Yes or No) = No.",
\item \textbf{"Metallic (Yes or No)"}: "Question: For crystal material, given the MP-id: mp-248, Formula: Fe2N and Band Gap: 0.00 eV, what is the Metallic nature (Yes or No) ? Answer: 1)**Band Gap Analysis** A material with a very small band gap (close to 0 eV) or no band gap is typically metallic, meaning it conducts electricity well. Metals have overlapping valence and conduction bands or a partially filled conduction band, which allows for free movement of electrons, leading to high electrical conductivity. Conversely, a material with a significant band gap is generally a semiconductor or an insulator, not a metal. The larger the band gap, the lesser the electrical conductivity, with insulators having the largest band gaps. 2)**Determining Metallic nature** If a material is known to be metallic (Metallic: Yes), you can infer that its band gap is effectively zero or negligible. If a material has a significant band gap (say, >2 eV), you can infer it's not metallic (Metallic: No). Because the band gap of Fe2N is 0.00eV, it's know to be metallic. 3)**Conclusion** So the Metallic (Yes or No) = Yes.",
\item \textbf{"Total Magnetization (unit: µB/f.u.)"}: "Question: For crystal material, given the MP-id mp-248 and Formula Fe2N, what is the Total Magnetization (unit: µB/f.u.)? Answer: 1)**Total Magnetization Analysis** In materials science, Total Magnetization refers to the magnetic moment per formula unit (f.u.), often measured in Bohr magnetons (µB). This property is crucial for understanding the magnetic properties of a material, which can be influenced by factors like the arrangement of atoms and the electronic structure. 2)**Determining Total Magnetization** Total Magnetization for Ferromagnetic or Ferrimagnetic crystal is usually around 3 µB/f.u. while for Antiferromagnetic or Non-magnetic crystal it's usually 0. It's very encouraged to give a hypothesis between that range. Here, based on existing knowledge up to the last update in January 2022, the Total Magnetization for this Ferromagnetic Fe2N is 3.00 µB/f.u.. 3)**Conclusion** So the Total Magnetization (unit: µB/f.u.) = 3.00.",
\item \textbf{"Ordering (Ferromagnetic/Ferrimagnetic/Antiferromagnetic/Non-magnetic)"}: "Question: For crystal material, given the MP-id mp-248, Formula Fe2N and Total Magnetization: 3.00 µB/f.u, what is the Ordering (Ferromagnetic/Ferrimagnetic/Antiferromagnetic/Non-magnetic)? Answer: 1)**Total Magnetization Analysis** This refers to the net magnetic moment of a material, expressed in units of Bohr magnetons per formula unit. It represents the sum of all magnetic moments in the material, taking into account both the direction and magnitude of each magnetic moment. 2)**Magnetic Ordering Analysis** In ferromagnetic materials, the magnetic moments of the atoms align parallel to each other, resulting in a strong net magnetic moment. A material with significant total magnetization is likely ferromagnetic. Ferrimagnetic materials have magnetic moments that are aligned in opposite directions but are unequal in magnitude, resulting in a net magnetic moment. Therefore, a non-zero total magnetization could indicate ferrimagnetic ordering. Antiferromagnetic materials have magnetic moments that are aligned in opposite directions and are equal in magnitude, resulting in no net magnetization. A material with zero or very low total magnetization might be antiferromagnetic. Non-magnetic materials do not have a net magnetic moment, indicating zero or negligible total magnetization. 3)**Determining Ordering** If the total magnetization is significantly high like > 0.50 µB/f.u, the material is likely ferromagnetic or ferrimagnetic. If the total magnetization is zero or very low, the material could be antiferromagnetic or non-magnetic. For Fe2N, Total Magnetization: 3.00 µB/f.u is significantly high, so it's Ferromagnetic. 4)**Conclusion** So the Ordering (Ferromagnetic/Ferrimagnetic/Antiferromagnetic/Non-magnetic) = Ferromagnetic."

\end{itemize}

Prompts for answer evaluation uniformly use the standard grading example for Molecular Formula prediction:

\begin{itemize}

\item Here're some fake examples of rating: [For small molecule Molecular Formula, one person gave an answer N/A. The ground truth is CH3OH. LLM: 0]. Here no information is ever given by examinee, so the score is 0. [For small molecule Molecular Formula, one person gave an answer C. The ground truth is CH3OH. LLM: 1]. Here the examinee gave a meaningful answer (though seems ridiculous), so the score is 1. [For small molecule Molecular Formula, one person gave an answer CHO. The ground truth is CH3OH. LLM: 3]. Here the examinee gave a meaningful answer and pointed out all the elements (though the number is wrong), so the score is 3. [For small molecule Molecular Formula, one person gave an answer C(CCCCO)CCCCBr. The ground truth is C(CCCCCO)CCCCCBr. LLM: 4]. Here the examinee gave a meaningful answer very close to ground truth, so so the score is 4. [For small molecule Molecular Formula, one person gave an answer OCC (or C(O)C). The ground truth is CCO. LLM: 5]. Here the examinee gave an answer intrinsically the same to ground truth, so the score is 5."

\end{itemize}

\subsection{Detailed Interactive Case Study Dialogues using Domain-Knowledge Embedded Prompt Engineering Method}

\begin{figure}[ht]
  \centering
  \includegraphics[width=0.9\linewidth]{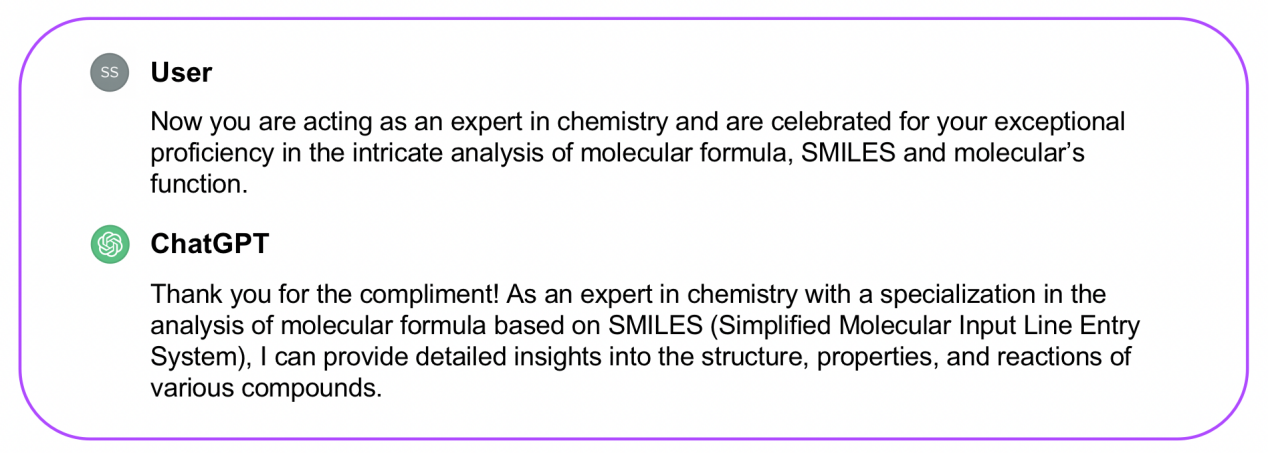}
\end{figure}

\begin{figure}[ht]
  \centering
  \includegraphics[width=0.9\linewidth]{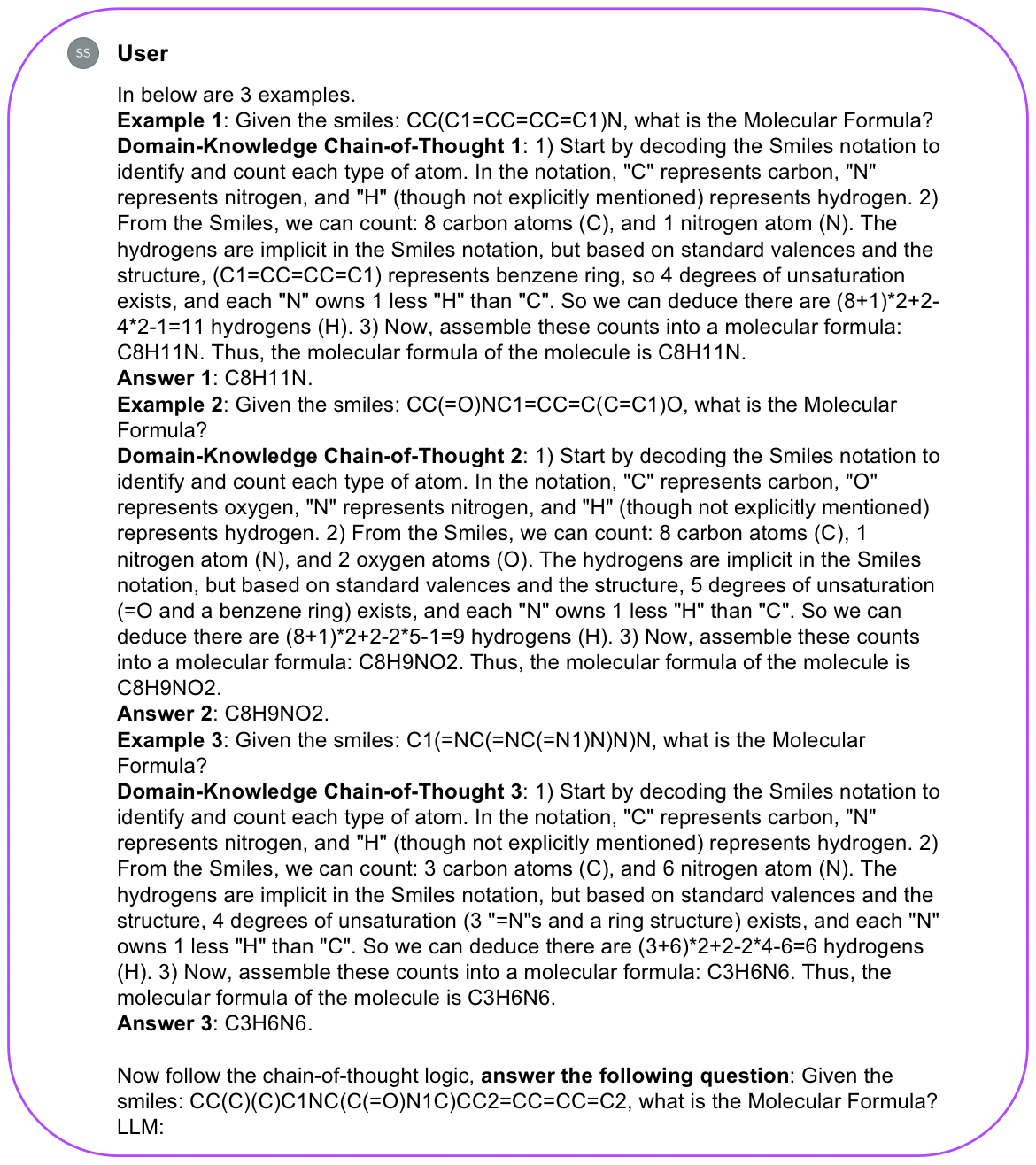}
\end{figure}

\begin{figure}[ht]
  \centering
  \includegraphics[width=0.9\linewidth]{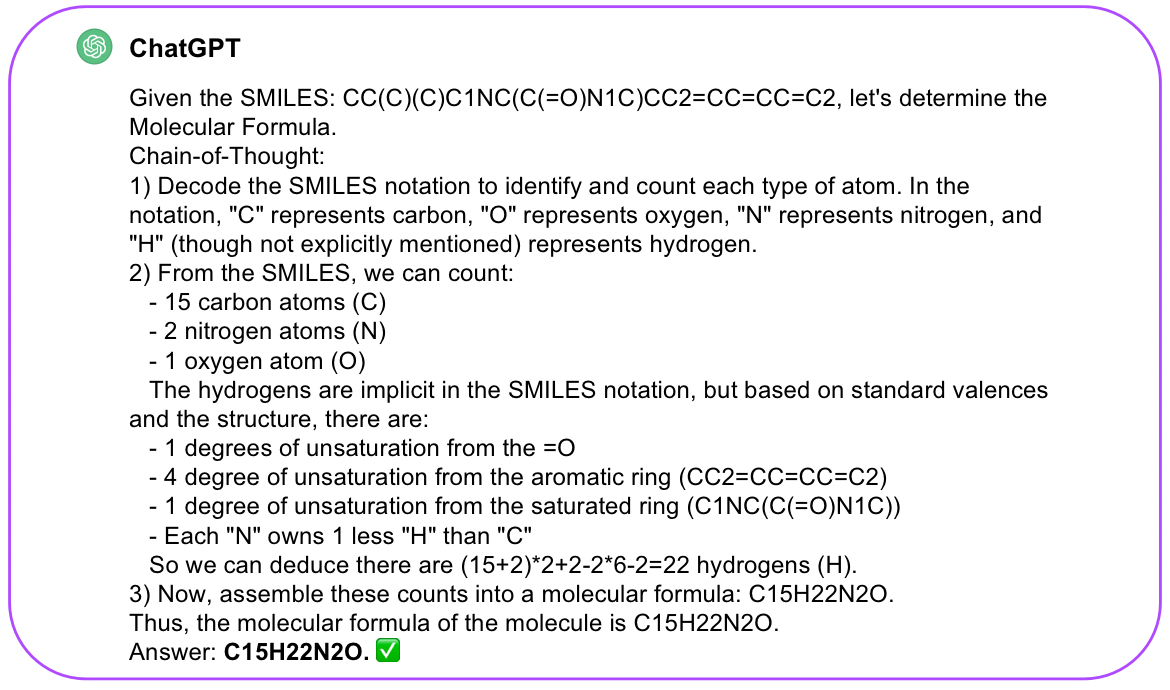}
\end{figure}

\begin{figure}[ht]
  \centering
  \includegraphics[width=0.9\linewidth]{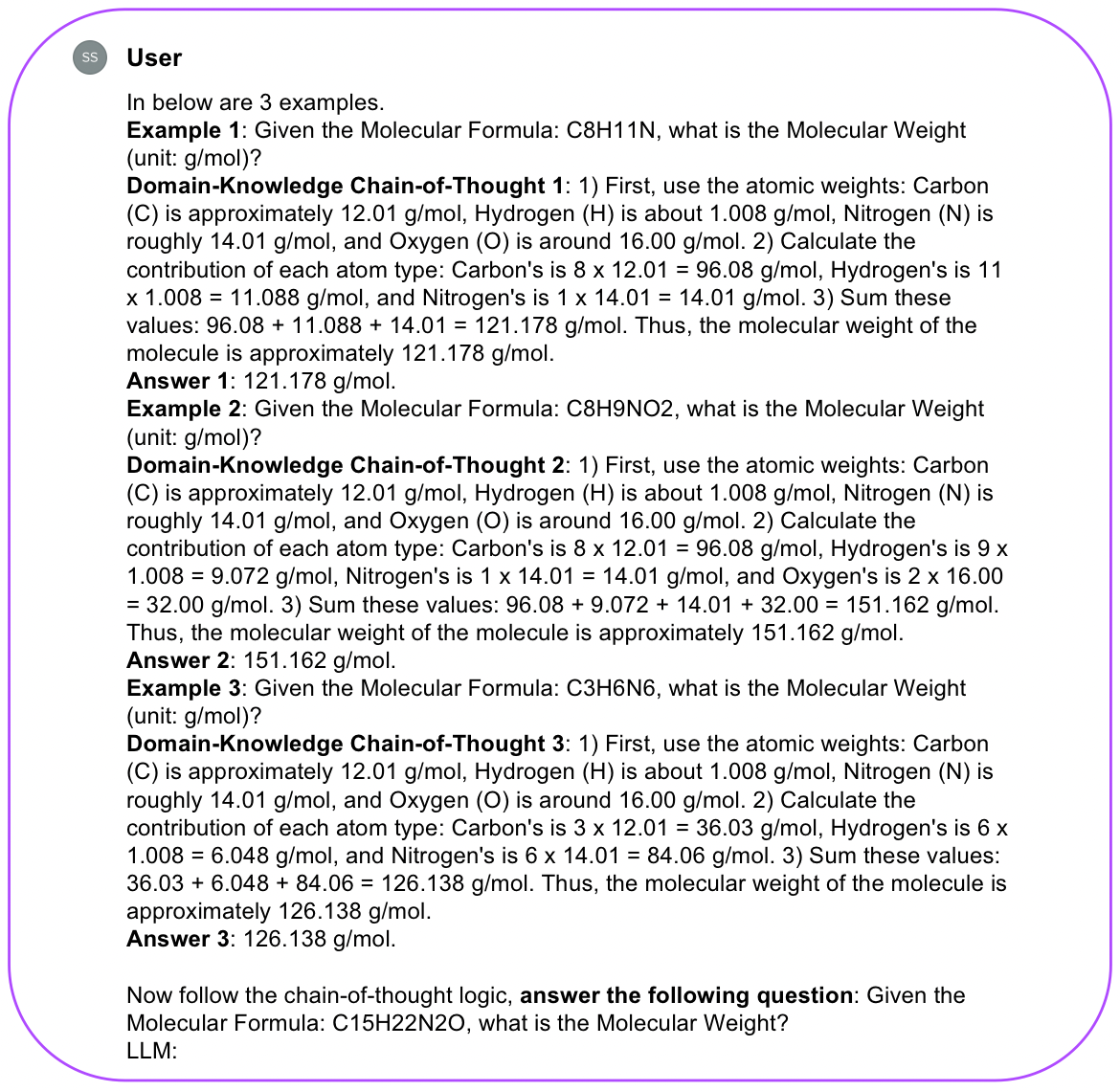}
\end{figure}

\begin{figure}[ht]
  \centering
  \includegraphics[width=0.9\linewidth]{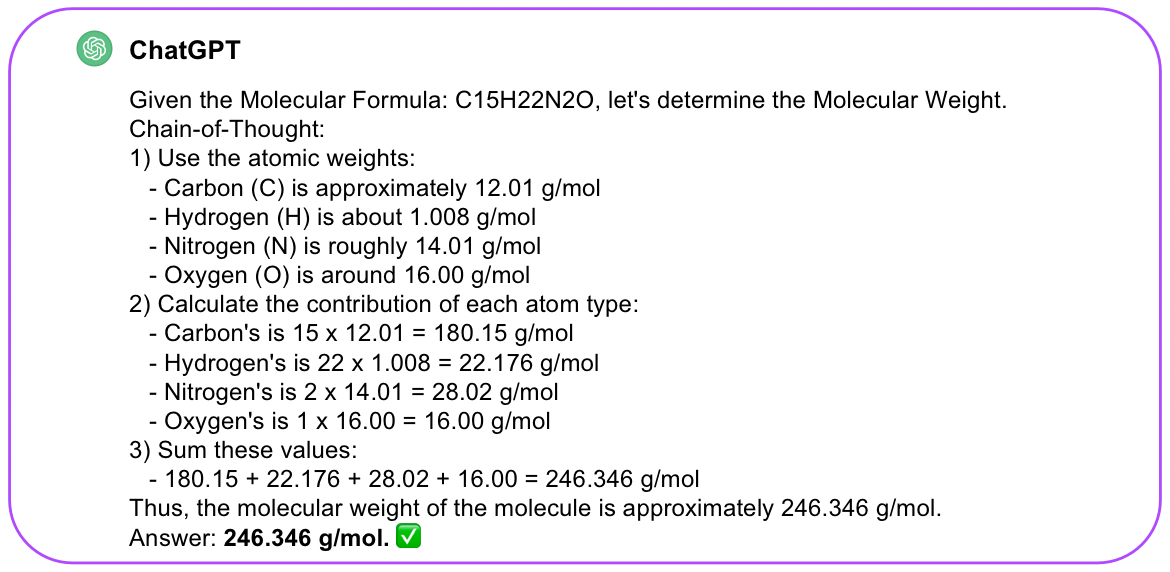}
\end{figure}

\begin{figure}[ht]
  \centering
  \includegraphics[width=0.9\linewidth]{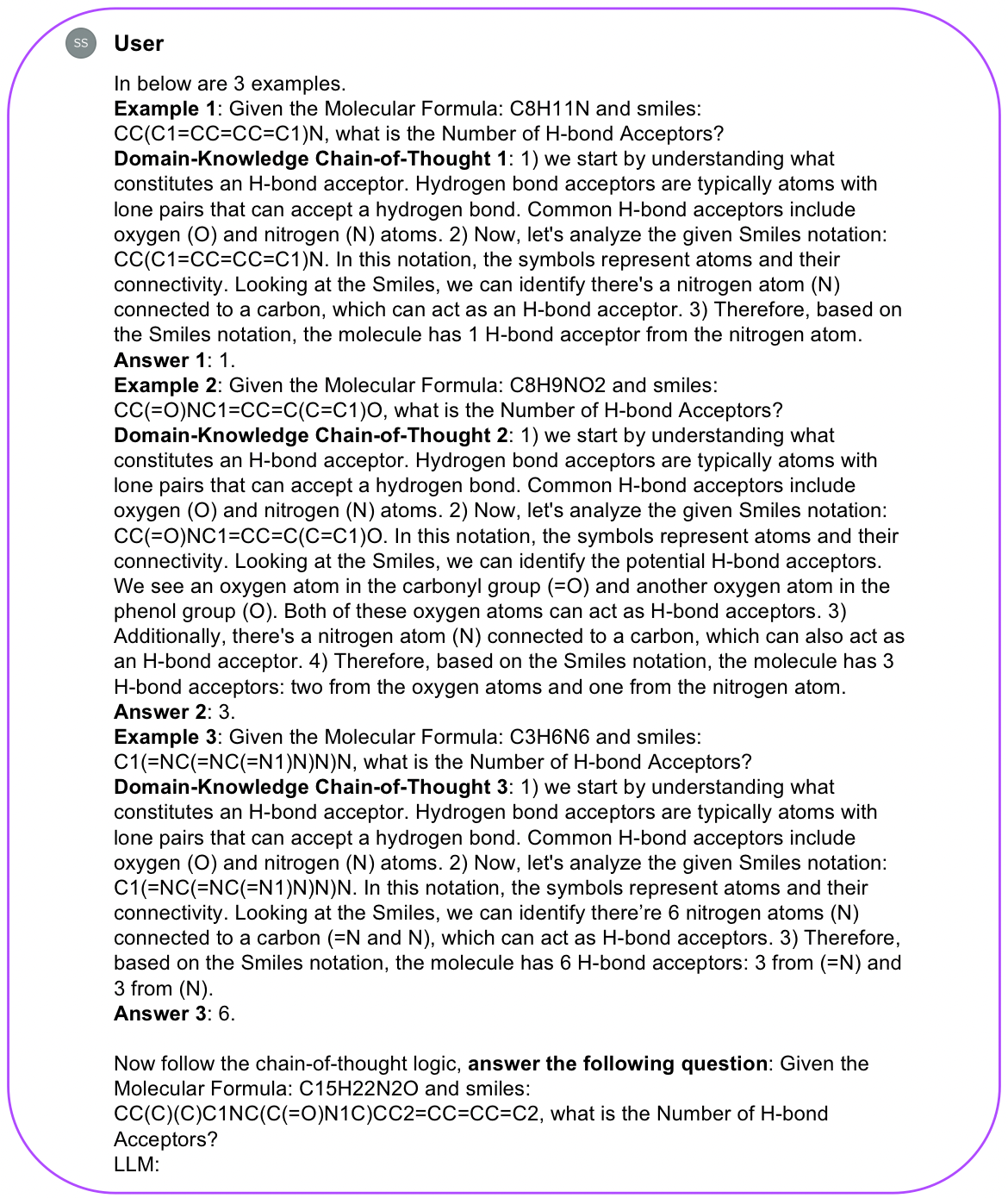}
\end{figure}

\begin{figure}[ht]
  \centering
  \includegraphics[width=0.9\linewidth]{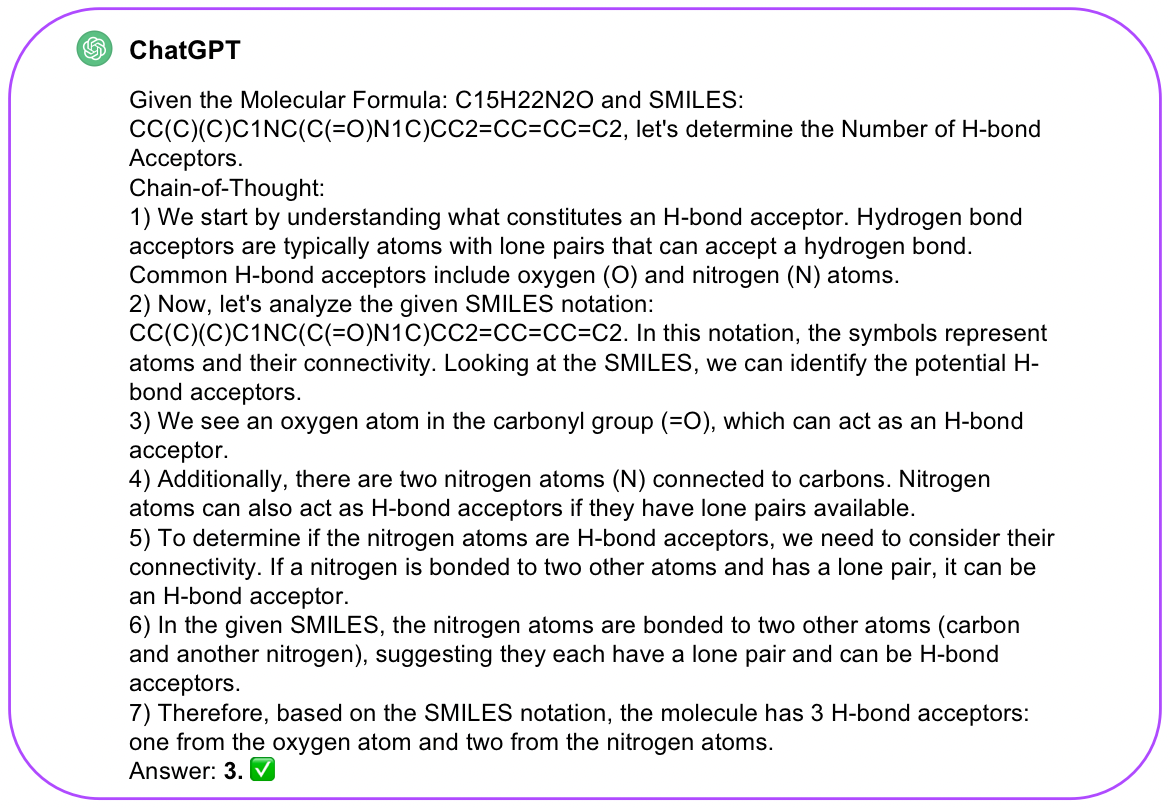}
\end{figure}

\begin{figure}[ht]
  \centering
  \includegraphics[width=0.9\linewidth]{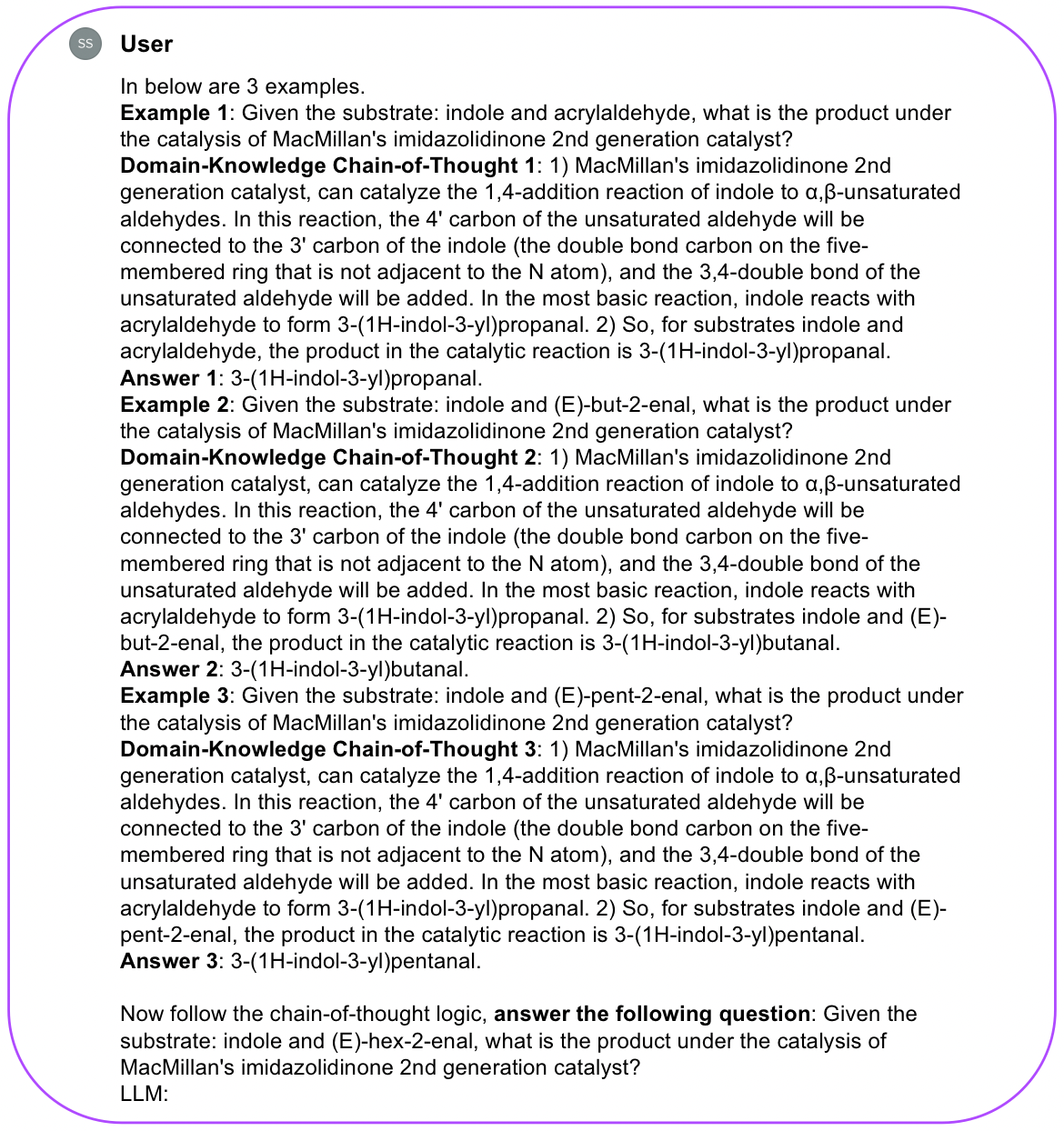}
\end{figure}

\begin{figure}[ht]
  \centering
  \includegraphics[width=0.9\linewidth]{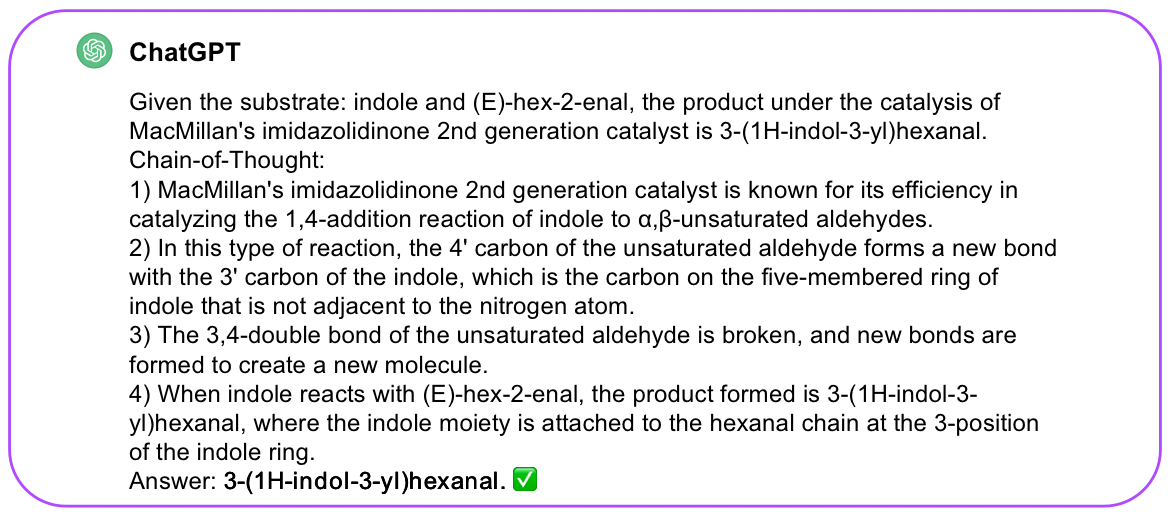}
  \caption{Detailed Prompt Engineering Interactive Dialogue on MacMillan's Imidazolidinone 2nd Generation Catalyst}
\end{figure}

\begin{figure}[ht]
  \centering
  \includegraphics[width=0.9\linewidth]{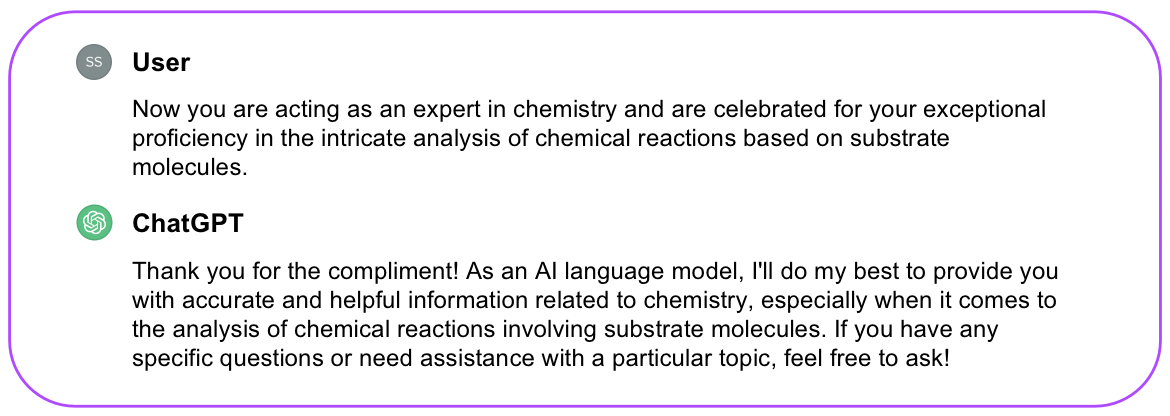}
\end{figure}

\begin{figure}[ht]
  \centering
  \includegraphics[width=0.9\linewidth]{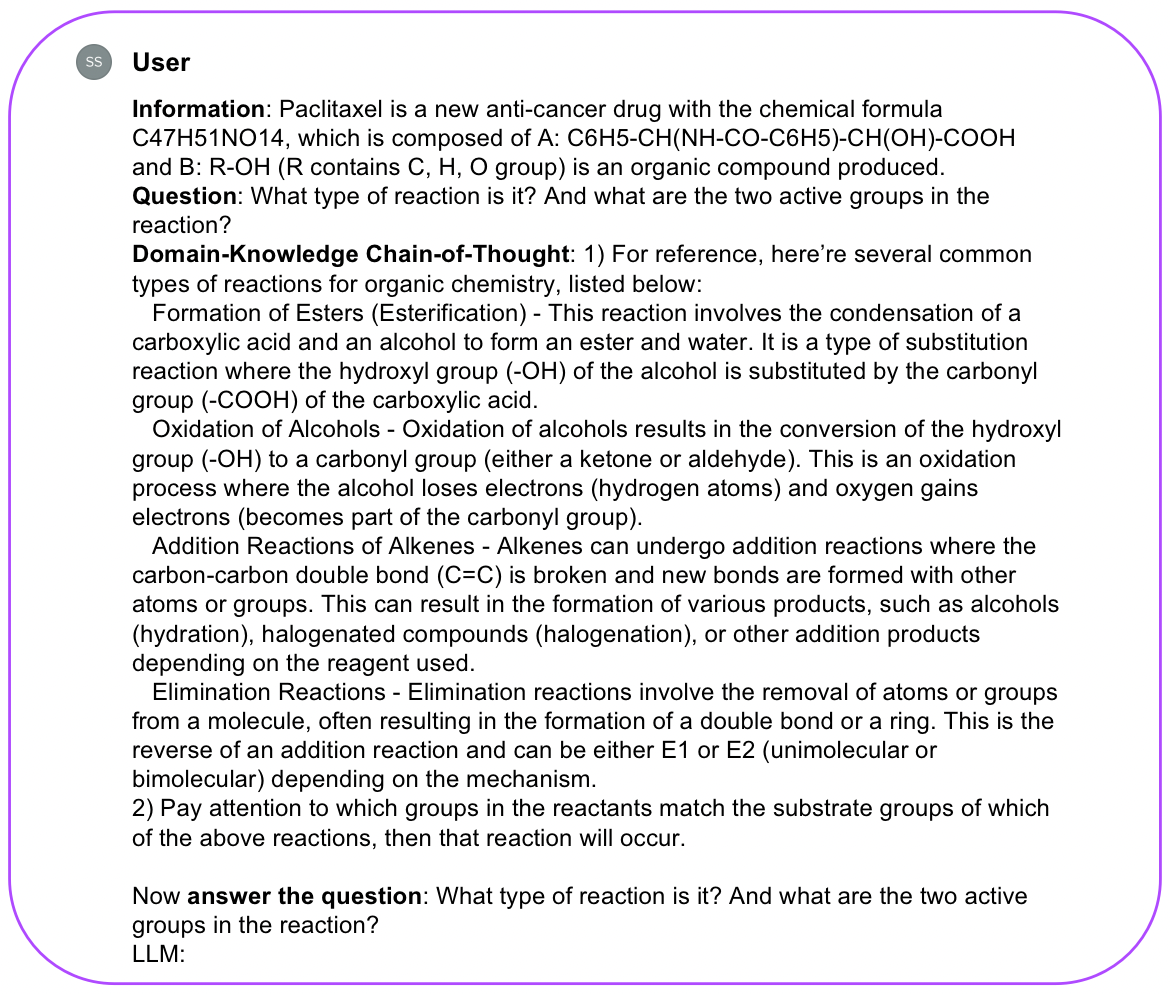}
\end{figure}

\begin{figure}[ht]
  \centering
  \includegraphics[width=0.9\linewidth]{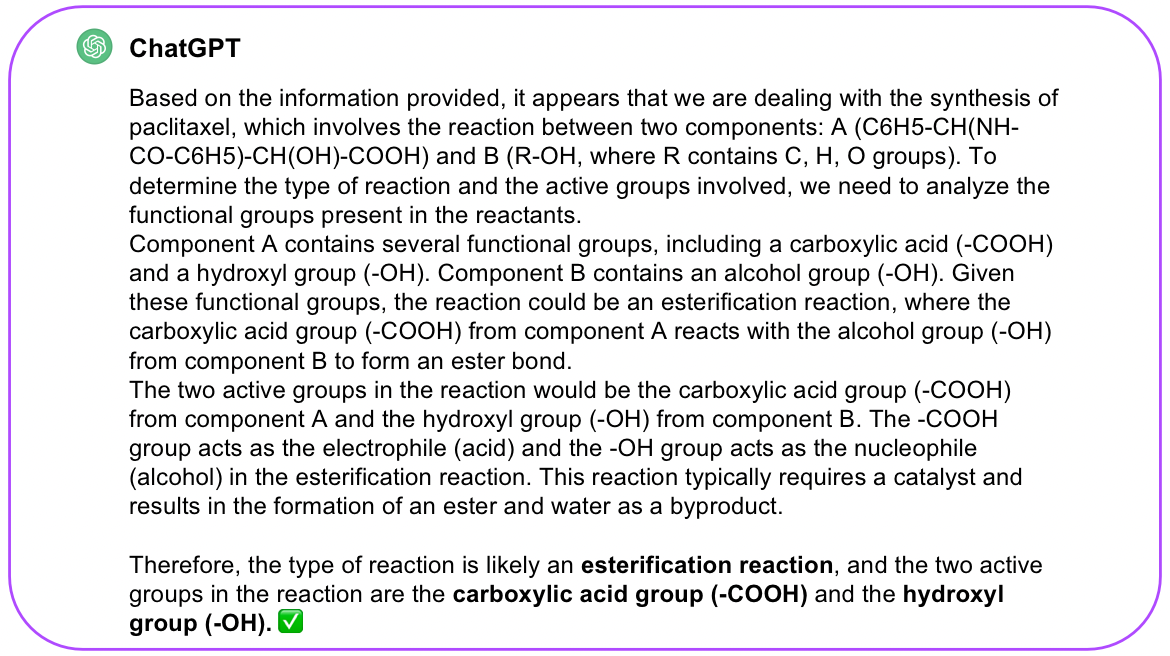}
\end{figure}

\begin{figure}[ht]
  \centering
  \includegraphics[width=0.9\linewidth]{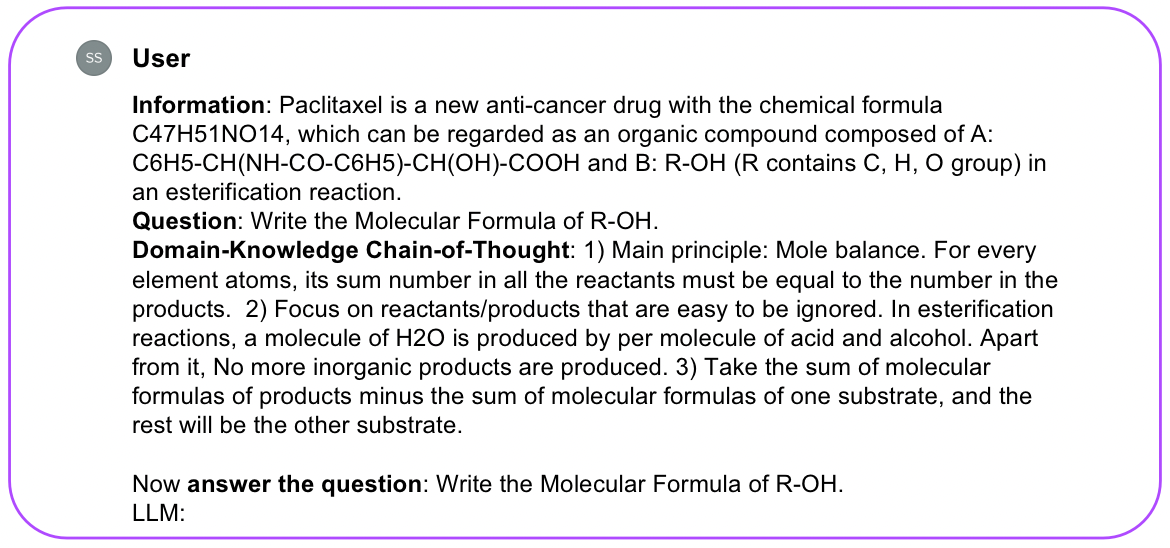}
\end{figure}

\begin{figure}[ht]
  \centering
  \includegraphics[width=0.9\linewidth]{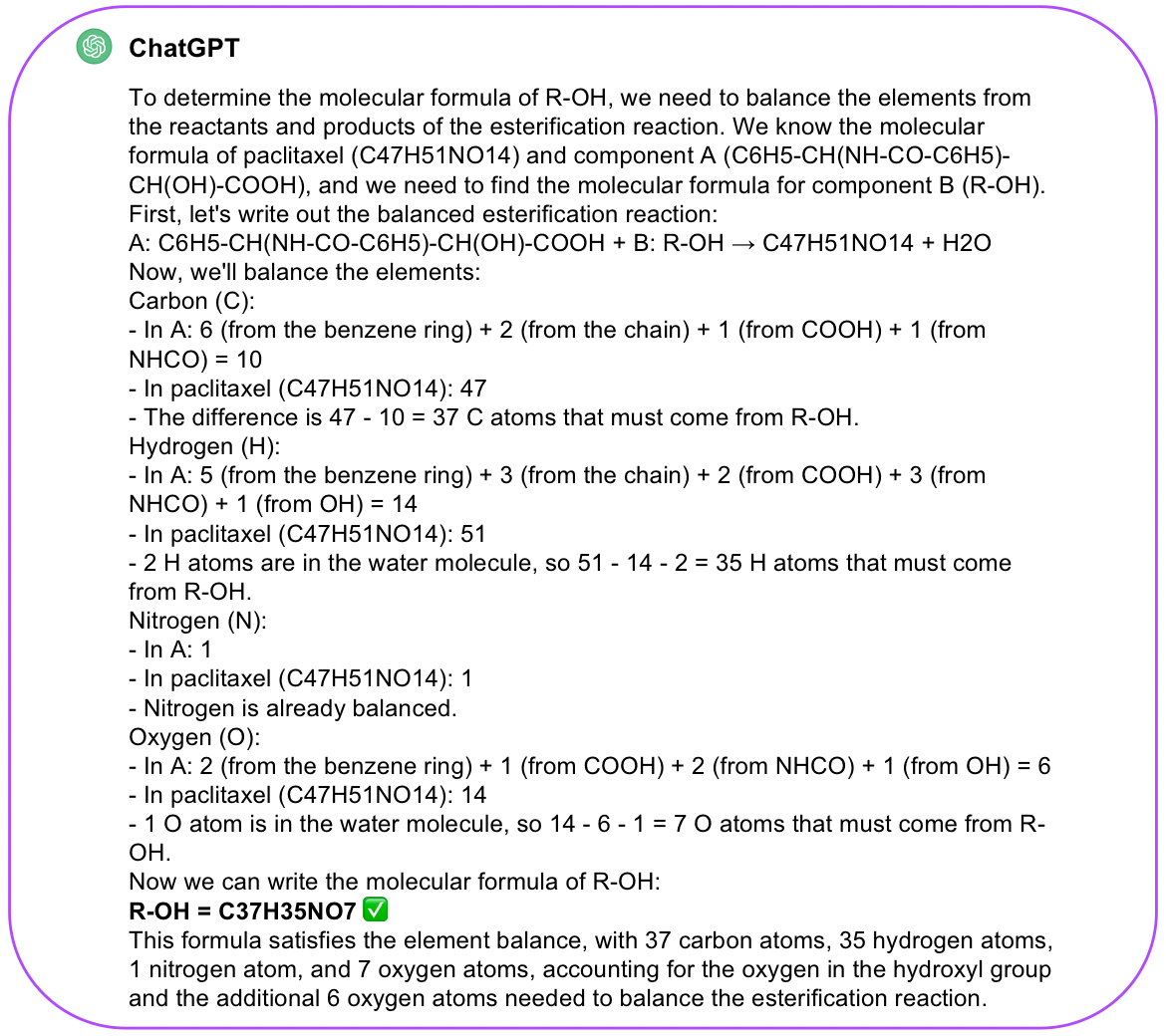}
  \caption{Detailed Prompt Engineering Interactive Dialogue on Paclitaxel}
\end{figure}

\begin{figure}[ht]
  \centering
  \includegraphics[width=0.9\linewidth]{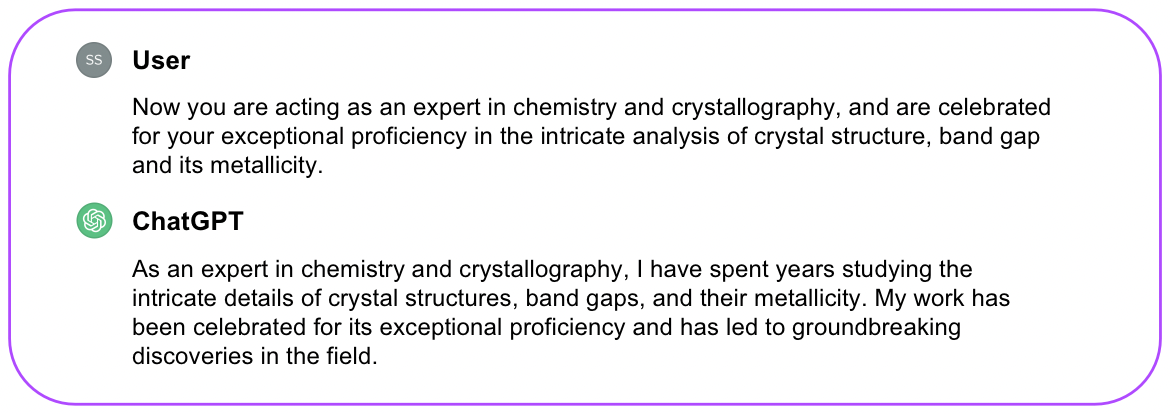}
\end{figure}

\begin{figure}[ht]
  \centering
  \includegraphics[width=0.9\linewidth]{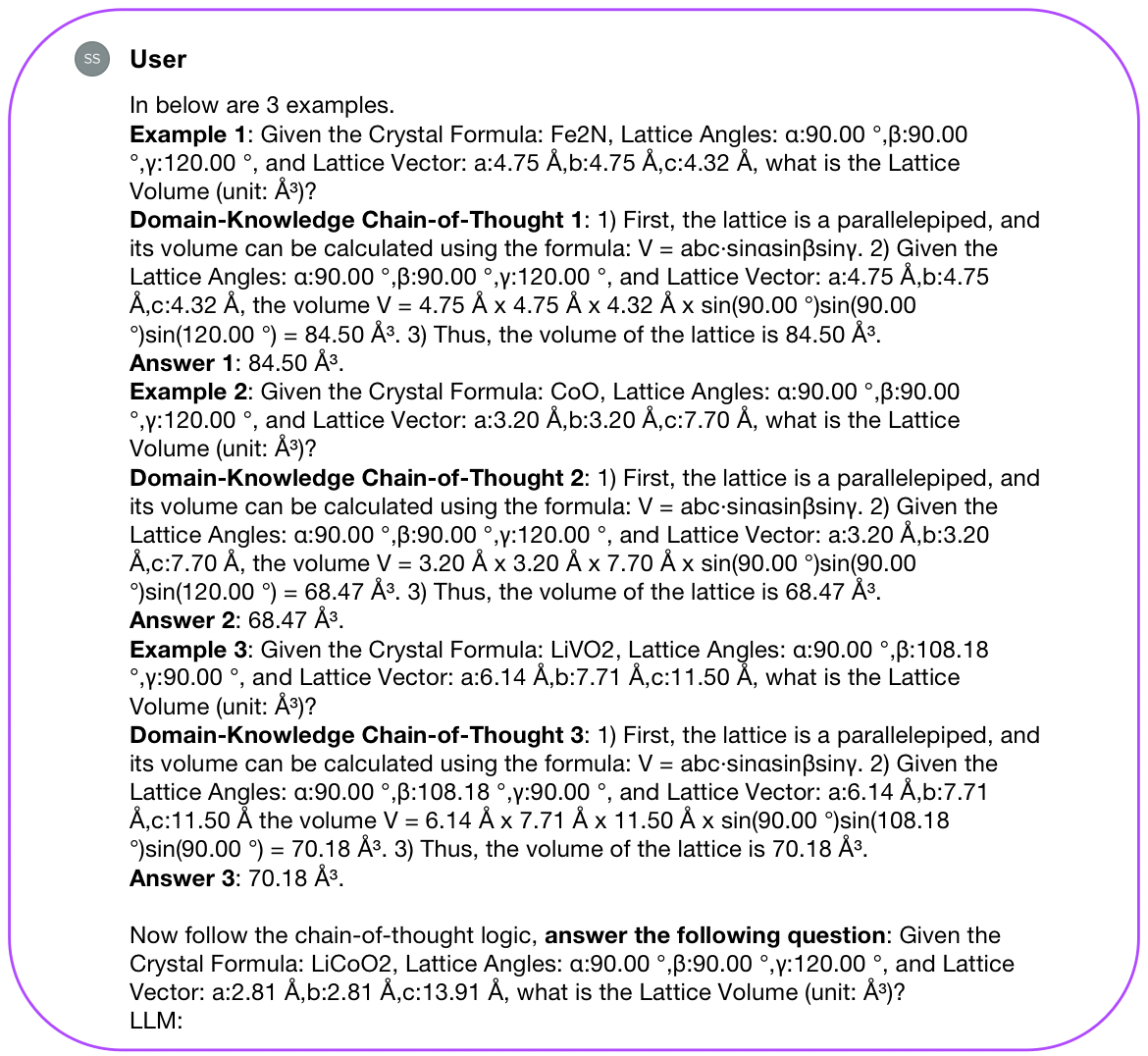}
\end{figure}

\begin{figure}[ht]
  \centering
  \includegraphics[width=0.9\linewidth]{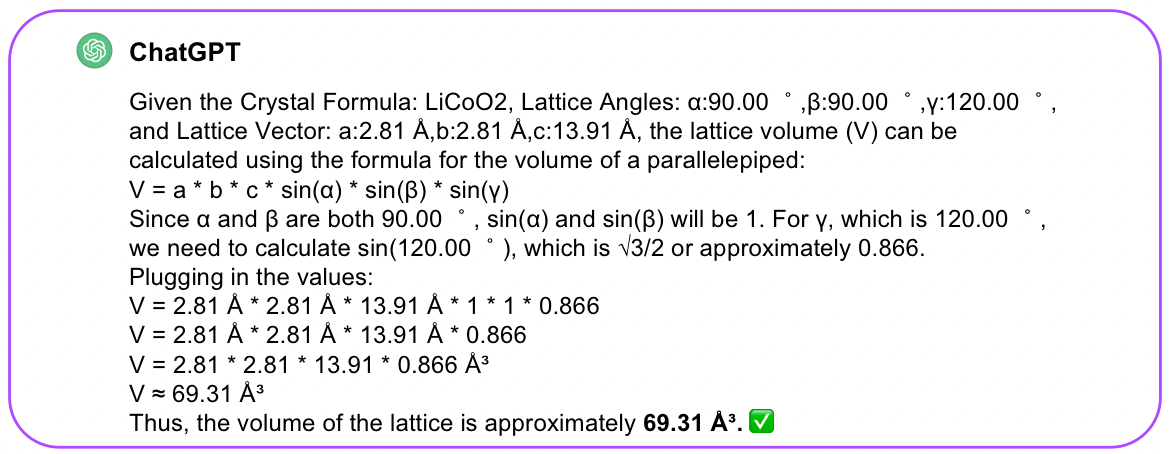}
\end{figure}

\begin{figure}[ht]
  \centering
  \includegraphics[width=0.9\linewidth]{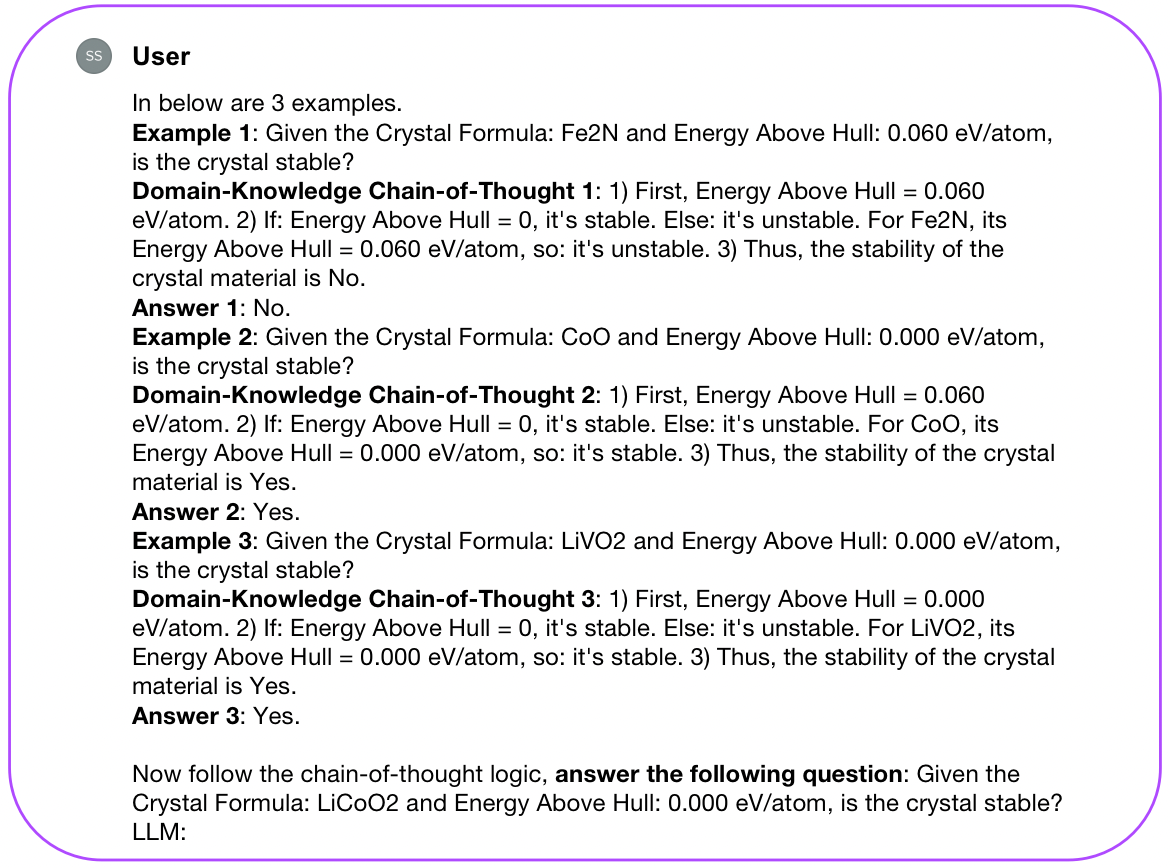}
\end{figure}

\begin{figure}[ht]
  \centering
  \includegraphics[width=0.9\linewidth]{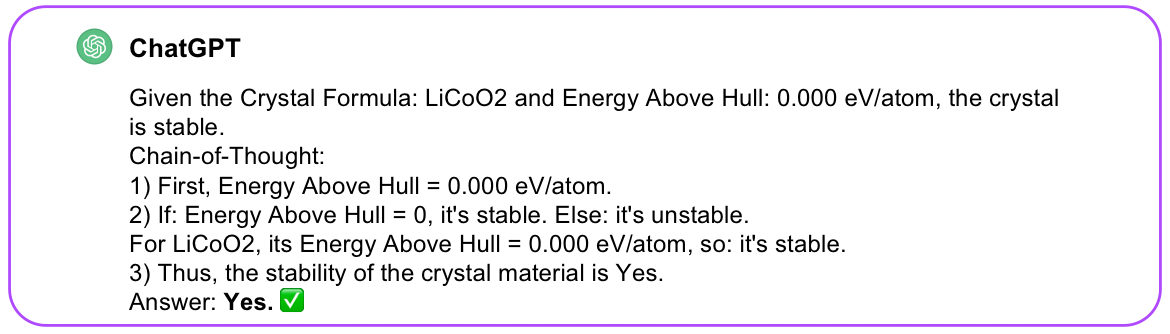}
  \caption{Detailed Prompt Engineering Interactive Dialogue on Lithium Cobalt Oxide}
\end{figure}

%%%%%%%%%%%%%%%%%%%%%%%%%%%%%%%%%%%%%%%%%%%%%%%%%%%%%%%%%%%%

\end{document}